\documentclass[mnsc,nonblindrev]{informs3}

\usepackage{etoolbox}

\AtBeginEnvironment{table}{\centering}
\AtBeginEnvironment{figure}{\centering}

\OneAndAHalfSpacedXI

\usepackage{natbib}
 \bibpunct[, ]{(}{)}{,}{a}{}{,}%
 \def\bibfont{\small}%

\usepackage[pdfencoding=auto]{hyperref}
\usepackage{bookmark}

\usepackage{multirow}
\usepackage{array}
\usepackage{booktabs}
\usepackage{amsmath}

\usepackage{thmtools, thm-restate}
 \usepackage[ruled]{algorithm2e}
\usepackage{algorithmic}
\usepackage{enumitem}
\usepackage{bbm}

\usepackage{comment}
\usepackage{setspace}
\usepackage{tikz}
\usepackage{pgfplots}
\usepgfplotslibrary{patchplots}
\usetikzlibrary{patterns, positioning, arrows}
\pgfplotsset{compat=1.15}
\usepackage{tikz-3dplot}
\usetikzlibrary{3d, calc}

\pgfmathdeclarefunction{noisydecay}{1}{%
    \pgfmathparse{(1/sqrt(#1)) + (0.1 * (rand - 0.5))}%
}
\TheoremsNumberedThrough  

\EquationsNumberedThrough

\numberwithin{equation}{section}
\numberwithin{remark}{section}
\numberwithin{lemma}{section}
\numberwithin{theorem}{section}
\numberwithin{corollary}{section}
\numberwithin{assumption}{section}
\numberwithin{proposition}{section}

\makeatletter
\long\def\@maketablecaption#1#2{
  \EGT\TableCaptionFontStyle
  \noindent
    {\TableNameFontStyle #1.}\enskip #2\HD{0}{6}\endgraf%
}

\long\def\@makefigurecaption#1#2{
  \EGT\FigureCaptionFontStyle
  \noindent
    {\FigureNameFontStyle #1.}\enskip #2\HD{0}{9}\endgraf
}
\makeatother

\begin{document}

\RUNAUTHOR{Aznag, Cummings, and Elmachtoub}
 \RUNTITLE{An active learning framework for multi-group mean estimation}
\TITLE{An active learning framework for multi-group mean estimation}
\ARTICLEAUTHORS{
\AUTHOR{Abdellah Aznag\qquad\qquad Rachel Cummings \qquad\qquad Adam N. Elmachtoub}
\AFF{Department of Industrial Engineering and Operations Research \& Data Science Institute, Columbia University}
}
\ABSTRACT{ We study a fundamental learning problem over multiple groups with unknown data distributions, where an analyst would like to learn the mean of each group. Moreover, we want to ensure that this data is collected in a relatively fair manner such that the noise of the estimate of each group is reasonable. In particular, we focus on settings where data are collected dynamically, which is important in adaptive experimentation for online platforms or adaptive clinical trials for healthcare. In our model, we employ an active learning framework to sequentially collect samples with bandit feedback, observing a sample in each period from the chosen group. After observing a sample, the analyst updates their estimate of the mean and variance of that group and chooses the next group accordingly. The analyst's objective is to dynamically collect samples to minimize the collective noise of the estimators, measured by the norm of the vector of variances of the mean estimators. 

We propose an algorithm, Variance-UCB, that sequentially selects groups according to an upper confidence bound on the variance estimate. We provide a general theoretical framework for providing efficient bounds on learning from any underlying distribution where the variances can be estimated reasonably. This framework yields upper bounds on regret that improve significantly upon all existing bounds, as well as a collection of new results for different objectives and distributions than those previously studied.}

\maketitle

\section{Introduction}\label{section:introduction}

Obtaining accurate estimates from limited labeled data is a fundamental challenge. The main framework in classical estimation theory starts from readily available samples and builds an estimator that satisfies desirable guarantees such as unbiasedness and accuracy. This classical framework is not concerned with the data collection process itself, i.e., how the samples were chosen. In practice, if done arbitrarily, the collected samples can be non-representative of the population as a whole, which can lead to bias and discrimination. To tackle this issue, active learning strategies where a decision-maker dynamically selects (or collects) samples have emerged as promising solutions. In this work, we address the fundamental challenge of dynamically collecting data from different groups in a population, in a way that balances overall estimation accuracy while controlling the noise of our estimators across all groups.

For instance, adaptive experimentation for online platforms and adaptive clinical trials for healthcare are settings where the sampling strategy changes as more is learned about the parameters of interest. These strategies view data as a limited resource, and the key decision is to how to dynamically collect data across different subgroups of the population. In particular, when learning a parameter in each subgroup, the analyst wants to ensure that the variances of the estimators are reasonable across all subgroups. Important examples include understanding the effect of a pharmaceutical drug, revenue effect of a promotion strategy, or a neighborhood-level quantity in a census study.

To address this challenge, we consider a general active learning framework for dynamic data collection when faced with different subgroups in the population that each have different distributions of data. We consider a population partitioned into $G$ disjoint \textit{groups}, where group $g \in [G]$ has data coming from a distribution $\mathcal{D}_g$  with \textit{unknown} mean $\mu_g$ and \textit{unknown} variance $\sigma_g^2$, and an analyst would like to learn the mean of each group. All the distributions $\mathcal{D}_g$ belong to a known set $\mathcal{H}$, which we call the hypothesis class. At each time period, the analyst selects one group and observes a sample of data from that group. After each observation, the analyst uses their history of observed samples to optimize their choice of the next group. At the end of a fixed time horizon $T$, the analyst holds $n_{g,T}$ samples from each group $g$, and computes the sample mean $\hat{\mu}_{g,T}$.

The variance of $\hat{\mu}_{g,T}$ is $\frac{\sigma_g^2}{n_{g,T}}$.  The objective of the analyst is to return a vector of mean estimates with the smallest $p$-norm of the variance vector of the mean estimates: 
\begin{equation*}
    R_p(\boldsymbol{n}_{T};\boldsymbol{\sigma}) := \left\|\left\{\frac{\sigma_g^2}{n_{g,T}}\right\}_{g \in [G]}\right\|_p.
\end{equation*}
The norm of the variance vector reflects how uncertainty in the estimates is spread across the different groups. For different choices of $p$-norms, denoted $\|\cdot\|_p$, we capture different aspects of multi-group estimation performance \citep{kloft2011lp,liu2022weighted}, such as the overall spread of the estimates ($p = 2$, Euclidean norm) or the worst case deviation from the true mean ($p = +\infty$, infinity norm). This choice of objective provides a more nuanced understanding of how the diverse underlying population is represented in the final estimation of the group means. The analyst can choose $p$ in a way that captures relevant notions of fairness in their data collection process or application domain. 

As an idealized upper bound, we compare against the smallest achievable performance in the full information setting where the variance vector for each group's data, $\boldsymbol{\sigma}^2$, is known:

\begin{equation*}
    R^*_{p}(\boldsymbol{\sigma}) := \quad \min_{\boldsymbol n \in \mathbb{R}_+^G} R_p(\boldsymbol n; \boldsymbol{\sigma}) \quad s.t. \quad \sum_{g \in [G]}n_g = T.
\end{equation*}

Since $\boldsymbol{\sigma}^2$ is unknown, our performance metric is \textit{normalized regret}. Formally, for a sampling policy $\boldsymbol{\pi}$, the normalized regret is
\begin{equation*}
    \text{Regret}_{p, T}(\boldsymbol{\pi},\boldsymbol{\mathcal{D}}|\mathcal{H}) = \frac{\mathbb{E}_{\boldsymbol{n}_T \sim (\boldsymbol{\pi}, \boldsymbol{\mathcal{D}})}\left[R_p(\boldsymbol{n}_T;\boldsymbol{\sigma})\right] - R^*_p(\boldsymbol{\sigma})}{R^*_p(\boldsymbol{\sigma})}.
\end{equation*}
This is also referred to as the \textit{optimality gap} or the \textit{multiplicative regret} in the literature.

\subsection{Our contributions}

The main results of this paper can be summarized as follows:
\begin{itemize}
    \item  \textbf{A general UCB procedure.} We propose a natural algorithm, Variance-UCB, that selects groups according to an upper bound on the estimate of the variances. Our algorithm can leverage any external procedure for generating upper confidence bounds on variance estimators and thus works in a wide range of settings, which we explain in Section \ref{section:algorithm}. 

      \item \textbf{A width-based unified regret analysis.} We develop a unified regret upper bound in Section \ref{subsection:width}. Starting from any instance $\boldsymbol{\mathcal{D}}$ and any hypothesis class $\mathcal{H}$, we show that the main driver for the regret is a quantity $\Bar{w}_p$ capturing the \textit{decision error}, and is a function of the widths of the upper confidence bounds on the standard deviations of $\boldsymbol{\mathcal{D}}$. Intuitively, $\bar{w}_p$ is the effective cost of mis-estimating $\boldsymbol{\sigma}$ with respect to the downstream performance $R_p$.  We show in Theorems \ref{theorem:upperboundinfinite} and \ref{theorem:upperboundfinite} that the final regret upper bound is a simple function of $\bar{w}_p$ for various types of $p$-norms, where the two theorems differentiate the infinite and finite norm settings since there is extra smoothness to be leveraged by finite $p$-norms. 

    The major benefit of Theorems \ref{theorem:upperboundinfinite} and \ref{theorem:upperboundfinite} is that both the distributions $\boldsymbol{\mathcal{D}}$ and the hypothesis class $\mathcal{H}$ are chosen arbitrarily.  These benefits allow us to improve and generalize prior results significantly.

    \item \textbf{Improved regret bounds for infinite norm.} For the case where $p=\infty$ and the underlying distributions are Gaussian and sub-Gaussian, we improve significantly upon existing literature \citep{activelearning, carpentier2011upper} (see Table \ref{tab:comparison} for details).  In particular, a consequence of our work is solving the long standing open question in \citet{carpentier2011upper} about the best possible bound in the case of sub-Gaussian feedback and infinite norm (see Corollary \ref{corollary:subgaussian}). A more detailed discussion on how these bounds are improved is deferred to the related work in Section \ref{section:literaturereview}.

\begin{table}[h]
    {\small
    \setlength{\tabcolsep}{7pt}
    \begin{tabular}{c c c c}
        Nature of feedback & Reference &Additional information& Leading regret term\\
        \addlinespace
        \hline
        \addlinespace
        Bounded & \cite{activelearning} & support $[a, b]$&$\frac{C(b-a)^2\|\boldsymbol{\sigma}\|_2G^2\log^2 GT}{\sigma_{\min}^2\sqrt{T}}$, $C \gg 1$\\
        \addlinespace
        \hline
        \addlinespace
        Sub-Gaussian & \cite{carpentier2011upper}&  Upper bound $s \geq \|\boldsymbol{\sigma}\|_\infty$&$7.6 \times 10^5\frac{s^2}{\sigma^2_{\min}}\frac{ G^2 \log^2 T}{\sqrt{T}}$\\
        \addlinespace
        & Our paper (Theorem \ref{theorem:subgaussian}) & Upper bound $\hat{\boldsymbol{c}} \geq \boldsymbol{\sigma}$&$4\sqrt{3}\frac{\|\hat{\boldsymbol{c}}\|_2}{\|\boldsymbol{\sigma}\|_2}\sqrt{\frac{G\log T}{T}}$ \\
        \addlinespace
        \hline
        \addlinespace
        Gaussian & \cite{carpentier2011upper} & Upper bound $\hat{\Sigma} \geq \|\boldsymbol{\sigma}\|^2_2$& $1.05 \times 10^5\frac{\hat{\Sigma}}{\|\boldsymbol{\sigma}\|^2_2}\frac{G^2\log^2 T}{\sqrt{T}}$\\
        \addlinespace
        & Our paper (Theorem \ref{theorem:gaussianfeedback})& None&$2\sqrt{3}\frac{\|\boldsymbol{\sigma}\|_1}{\|\boldsymbol{\sigma}\|_2}\sqrt{\frac{\log T}{T}} $ \\
        \addlinespace
        \hline
        \addlinespace
        Exponential & Previous best bound & No previous bound &-\\
        \addlinespace
        & Our paper (Theorem \ref{theorem:exponential})& None&$2\sqrt{3}\sqrt{\frac{\log T}{T}} \frac{\|\boldsymbol{\sigma}\|_1}{\|\boldsymbol{\sigma}\|_2}$
    \end{tabular}
    }
    \caption{Summary of asymptotic regret bounds for infinite norm with common distributions. {\normalfont Regret bounds' leading term for Variance-UCB in the case of the infinite norm compared against previously best known regret bounds for $G$ groups, time horizon $T$, and standard deviation vector $\boldsymbol{\sigma}$.}}
    \label{tab:comparison}
\end{table}

\item \textbf{New bounds for finite $p$-norms.} Unlike the infinite norm case, the finite \(p\)-norm setting has remained unstudied prior to this work. We prove a set of regret bounds any finite $p$ (see Table \ref{tab:comparisonfinite}). These regret bounds are novel in the sense that they provide the first performance guarantees in the finite \(p\)-norm regime.

 \begin{table}[h]
    \centering
    {\small
    \begin{tabular}{c c c c}
        Nature of feedback &Additional information& Leading regret term &Reference\\
        \addlinespace
        \hline
        \addlinespace
        Sub-Gaussian & Upper bound $\|\hat{\boldsymbol{c}}\|_2 \geq \|\boldsymbol{\sigma}\|_2$&$85p\left(\sum_{g \in [G]} \frac{\hat{c}_g^2}{\sigma_g^2}\right)\frac{\log T}{T}$ &Theorem \ref{theorem:subgaussian}\\
        \addlinespace
        \hline
        \addlinespace
        Gaussian & None&$\frac{43pG\log T}{T}$ &Theorem \ref{theorem:gaussianfeedback}\\
        \addlinespace
        \hline
        \addlinespace
        Exponential & None&$\frac{43pG\log T}{T}$ &Theorem \ref{theorem:exponential}
        \vspace{0.2cm}
        \end{tabular}
    }
    \caption{Summary of asymptotic regret bounds for finite \(p\)-norms with common distributions. {\normalfont Regret bounds' leading term for Variance-UCB in the case of $p < +\infty$ for $G$ groups, time horizon $T$, and standard deviation vector $\boldsymbol{\sigma}$.}}
    \label{tab:comparisonfinite}
\end{table}

\item \textbf{Strong performance with small sample size.} While our bounds essentially focus on the leading term in the regret, we also obtain remarkably small leading constants. Thus, our results practically outperform previous bounds as the resulting sample complexity is orders of magnitude better than the current state of the art  (see Table \ref{table:samplesg}).
   
\item \textbf{Regret bounds for exponential tails.} 
Theorems \ref{theorem:upperboundinfinite} and \ref{theorem:upperboundfinite} present the first-ever regret bounds for this problem that are applicable to exponential distributions, significantly broadening the scope beyond the traditionally assumed sub-Gaussian framework. This extension is particularly noteworthy, as existing results have universally relied on assumptions of sub-Gaussian tails to derive meaningful regret bounds. Our approach fundamentally shifts the analytical focus from tail behavior to the intrinsic complexity of estimating distributions within the hypothesis class $\mathcal{H}$. Tables \ref{tab:comparison} and \ref{tab:comparisonfinite} illustrate empirically that our algorithm attains remarkably low regret even for exponential distributions.

 This contribution highlights a deeper structural insight: it is the complexity of hypothesis class, rather than tail constraints, that dictates learning difficulty and performance.

\item \textbf{Novel analysis of coupled learning-decision dynamics.} The proofs of Theorems \ref{theorem:upperboundinfinite} and \ref{theorem:upperboundfinite} (see Section \ref{section:proof}) rely on expressing the learning dynamics in a novel way. At any time step $t \in [T]$, two dynamics simultaneously happen: a learning dynamic, and a decision dynamic. First, the decision made at each step consists of generating a new independent sample, which improves information on $\boldsymbol{\sigma}$ -- this is the learning dynamic. Second, information on $\boldsymbol{\sigma}$ will induce information on the optimal target number of samples $\boldsymbol{n}^*_t$ -- this is the decision dynamic. This exchange between the parameter space (represented by possible values of variances), and the decision space (represented by the number of collected samples) defines a dynamic system that describes the learning aspects of the problem (see Proposition \ref{prop:improvedrates}). We then obtain an upper bound on the deviations, and thus regret bounds, by studying the fixed points of the dynamic system.

\end{itemize}
\subsection{Related work}\label{section:literaturereview}

Decision-makers across industries have increasingly adopted large-scale, principled experimental design methodologies to assess the impact of strategic interventions. Notably, these methodologies are not limited to firms but extend to a wide range of domains, including clinical trials \citep{malani2024representation}, government intervention \citep{riboni2023policy}, and electoral polling \citep{gelman2023forecasts,nkouaga2024exploring}. However, these approaches come with inherent constraints, such as experimental risk and cost per trial—both of which scale with the number of experiments, often reaching millions or even billions per day \citep{xie2016improving}. A central challenge in this field is thus how to efficiently allocate experimentation resources to maximize learning while mitigating these constraints. To tackle this issue, adaptive experimental design methods, where the analyst sequentially determines which observations to collect based on interim results, have emerged as promising solutions \citep{bhandari2024dose,reshidi2024sequential,simchi2024multiarmed}. These approaches frame data collection as an experimental resource allocation problem, in which the key decision is how to dynamically assign experimental units across different strata of the population. The rationale behind these methods is that by allowing for adaptive treatment assignment, the experimenter can mitigate biases and iteratively refine the sampling strategy. A central challenge in this framework is how to adaptively allocate experimental resources across different subgroups in a way that maximizes statistical efficiency while ensuring equitable representation. 

Beyond experiment design, our motivation stems from growing attention to data collection methods \citep{sarkies2015data, NBERw26296,de2008choosing, grove1992observational}. We focus on the problem of mean estimation and dynamically collecting data to achieve this goal. While there is a substantial body of literature on data acquisition \citep{doi:10.1287/mnsc.2019.3424, liu2019data}, specifically in the presence of privacy concerns and associated costs \citep{10.1145/1993574.1993605, 10.1145/2229012.2229073,10.1145/2554797.2554835, pmlr-v40-Cai15,CFMT22,CEG+23}, our approach differs as we do not consider the costs of sharing data. Instead, we concentrate on how the data collection process itself should be conducted.

Considering data acquisition from the perspective of active learning is a natural approach \citep{BALCAN200978, balcan2010true, liu2017iterative}, and in that sense our work is also related to active learning \citep{cohn1996active, 9351818,kuwata2009real,9426901}. In \cite{10.1145/2764468.2764519}, the authors address the optimal data acquisition problem under the assumption of additive objective functions. They formulate the problem as an online learning problem and leverage well-understood tools from online convex optimization \citep{NIPS2017_dc960c46, MAL-024, contextual}. However, their ideas do not apply to our setting due to the non-additive nature of our regret over time.  

Our work is naturally related to multi-armed bandits problems \citep{auer2002finite, slivkins2019introduction}, in the sense that each group can be seen as an arm, and choosing a group from which to sample at each time step corresponds to choosing which arm to pull. However, the performance criterion for multi-armed bandits is measured by the difference between the mean of the chosen arm and the best arm \citep{agrawal2014bandits, cella2020meta,soare2014best}. In our framework, the means of the chosen arms do not impact the performance. It is their variances that matter in the optimal solution, as we measure performance by considering the $p$-norm of the variance of the estimator, which can be non-convex. This is fundamentally different from the reward maximization paradigms in the classic bandits framework, where a learner's goal is to identify the single best arm, and typical ``good'' policies (e.g., Thompson sampling, UCB) end up sampling from only one arm as $T\to \infty$. Because of this, and to the best of our knowledge, usual bandits algorithms and proof techniques \citep{lattimore} do not apply. Instead, we propose Variance-UCB, an algorithm that leverages existing bounds on the variance adjusted to the chosen norm.

Our work is closest to \cite{activelearning} and \cite{carpentier2011upper}, where the authors consider the same problem with $p = +\infty$. In \cite{activelearning}, the authors propose GAFS-MAX, and derive a $\Tilde{O}\left(\frac{(a-b)^2\|\boldsymbol{\sigma}\|_2G^2\log^2 GT}{\sigma_{\min}^2\sqrt{T}}\right)$ regret bound  for the special case where the distributions $\boldsymbol{\mathcal{D}}$ have bounded support with known bounds on $[a, b]$. They conjectured that the $\sigma^{-1}_{\min}$ dependency can be improved. \cite{carpentier2011upper} provide two algorithms, CH-AS and B-AS, and yield a regret upper bound for the much larger class of sub-Gaussian distribution, given a known upper bound on the variances $s^2$, and derive a $\Tilde{O}\left(\frac{s^2 G^2 \log^2 T}{\sigma^2_{\min}\sqrt{T}}\right)$ regret bound. The authors leave the following open question: what is the optimal dependence in $\sigma_{\min}^{-1}$ for the sub-Gaussian class? We solve this open problem by providing an improved regret bound with no dependence on $\sigma_{\min}^{-1}$. \cite{carpentier2011upper} also give an algorithm for the special case of Gaussian distributions, assuming a known upper bound on the variances, $\hat{\Sigma} \geq \|\boldsymbol{\sigma}\|_2^2$, and derive a $\Tilde{O}\left(\frac{G^2}{\sqrt{T}}\frac{\hat{\Sigma}}{\|\boldsymbol{\sigma}\|^2_2}\right)$ regret bound. Our analysis also improves these guarantees for the Gaussian case to an instance-dependent regret bound of $\tilde{O}\left(\frac{1}{\sqrt{T}}\cdot\frac{\|\boldsymbol{\sigma}\|_1}{\|\boldsymbol{\sigma}\|_2}\right)$ that requires no upper bound knowledge on $\boldsymbol{\sigma}$, which also implies a worst-case regret bound of $\Tilde{O}\left(\sqrt{\frac{G}{T}}\right)$.

\subsection{Organization of the paper and notation}
 In Section \ref{section:prelims}, we formally frame the multi-group estimation problem. In Section \ref{section:algorithm}, we present our algorithm, Variance-UCB. In Section \ref{subsection:width}, we formally introduce the \textit{decision error}, and present our main regret upper bounds in Theorems \ref{theorem:upperboundinfinite} and \ref{theorem:upperboundfinite}. In Section \ref{section:applications}, we show how our theorems can be applied to various hypothesis classes of distributions. In Section \ref{section:proof}, we provide the main steps behind deriving Theorems \ref{theorem:upperboundinfinite} and \ref{theorem:upperboundfinite}, and we conclude in Section \ref{section:conclusion}

 For a function $f$, a fixed point $x$ is an element satisfying $f(x) = x$. Similarly, a postfixed point $x$ is an element satisfying $x \leq f(x)$. Vectors are denoted in \textbf{bold} (e.g., $\boldsymbol{x}, \boldsymbol{\sigma}, \ldots)$, and scalars are in standard font (e.g., $x_1, \sigma_1, \ldots$). $\boldsymbol{0}$ represents the null vector (where the vector space is understood from context), and $\boldsymbol{1}$ the vector with $1$'s in every coordinate. For ease of notation, vector coordinate-wise operations are denoted in the same way as their scalar analog (e.g., $\boldsymbol{x} \boldsymbol{y}$ denotes the vector $(x_1 y_1,\ldots,x_n y_n)$, and $\frac{\boldsymbol{x}}{\boldsymbol{y}}$ denotes the vector $\left(\frac{x_1}{y_1}, \ldots, \frac{x_n}{y_n}\right)$). For two vectors $\boldsymbol{a}, \boldsymbol{b} \in \mathbb{R}^n$, we have $\boldsymbol{a} \geq \boldsymbol{b}$ if $a_k \geq b_k$ for all $k = 1, \ldots, n$. Similar notation is used for $>, \leq, <$. We use the convention that $\frac{1}{+\infty} = 0$, and $\frac{1}{0^+} = +\infty$.  For event $A$, its complement is denoted $A^{\sf c}$ and we have $\mathbb{P}(A) + \mathbb{P}(A^{\sf c}) = 1$. We use standard Big-$O$ notation, and consider the asymptotic regime where $T \to +\infty$. 

 \section{Problem formulation}\label{section:prelims}

We consider a population partitioned into $G \geq 2$ disjoint groups. Each individual in the population holds a real-valued data point. Data from each group $g \in [G]$ are distributed according to an \textit{unknown} distribution $\mathcal{D}_g$ with unknown mean $\mu_g$ and unknown variance $\sigma_g^2 > 0$. We refer to $\boldsymbol{\mathcal{D}} = \mathcal{D}_1 \times \ldots \times \mathcal{D}_G$ as an instance, and they will represent the distributions of a multi-armed bandit. Moreover, we assume that each distribution $\mathcal{D}_g$ belong to a \textit{known} hypothesis class $\mathcal{H}$.

The analyst wishes to compute an unbiased estimate of the population mean for each group within a finite time horizon $T$. At each time $t \in [T]$, the analyst selects a group $X_t \in [G]$ from which to collect a new \textit{independent} sample $Y_t \sim \mathcal{D}_{X_t}$. The analyst will design a data collection policy $\boldsymbol{\pi} = \{\pi_t\}_{t \in [T]}$, where each $\pi_t$ maps the observed history to the group $X_t$ (possibly randomly) selected at time $t$. In that sense, this is an active learning setting with bandit feedback. Formally, the set of feasible policies is defined as,
\begin{equation*}
    \Pi = \left\{\boldsymbol{\pi} = \{\pi_t\}_{t \in [T]}\; | \;  \pi_t \in G^{t-1} \times \mathbb{R}^{t-1} \to \Delta_G, \; \forall t \in [T] \right\},
\end{equation*}
where $\Delta_G$ is the set of distributions supported on $[G]$. Let $n_{g,T}$ denote the number of collected samples from group $g$ at the final time $T$, and let $\hat{\mu}_{g,T}$ be the sample mean estimator of $\mu_g$ given the $n_{g,T}$ collected samples. Once all data have been collected at the end of the time horizon $T$, the analyst will compute the sample mean of each group:
\begin{equation*}
    \hat{\mu}_{g,T} = \frac{1}{\sum_{t = 1}^T \mathbbm{1}_{X_t = g}}\sum_{t = 1}^T \mathbbm{1}_{X_t = g}Y_t .
\end{equation*}

Given a vector of number of samples $\boldsymbol{n}_T$, the variance of $\hat{\mu}_{g,T}$ is $\frac{\sigma_g^2}{n_{g,T}}$. The $p$-norm of the vector of variances $\frac{\boldsymbol{\sigma}^2}{\boldsymbol{n}_T} = \left\{\frac{\sigma_g^2}{n_{g,T}}\right\}_{g \in [G]}$ is denoted by 
\begin{equation*}
R_{p}(\boldsymbol{n}_T;\boldsymbol{\sigma}) := \left\| \frac{\boldsymbol{\sigma}^2}{\boldsymbol{n}_T} \right\|_p,
\end{equation*}
for any $p \in [1, +\infty]$. We assume the analyst wishes to minimize $\mathbb{E}_{\boldsymbol{\pi},\boldsymbol{\mathcal{D}}}[R_p(\boldsymbol{n}_T;\boldsymbol{\sigma})]$, where the randomness stems from the interaction of the policy $\boldsymbol{\pi}$ and the instance $\boldsymbol{\mathcal{D}}$. The norm of the variance vector aggregates the accuracy of the estimation across groups.

When choosing a policy, the analyst does not have access to the true standard deviation vector $\boldsymbol{\sigma} = (\sigma_1, \ldots, \sigma_G)$, which is needed to compute the value $R_{p}(\boldsymbol{n}_T)$. Therefore the analyst must learn $\boldsymbol\sigma$ through their decisions and data collection. A reasonable benchmark would be the performance of the best possible policy under complete information,  where $\boldsymbol{\sigma}$ is known. That is,
\begin{equation*}
\min_{\boldsymbol n \in \mathbb{N}^G} R_{p}(\boldsymbol n;\boldsymbol{\sigma}) \quad s.t. \quad \sum_{g \in [G]}n_g = T.
\end{equation*}

However, the optimization program above can be difficult to solve and analyze due to the integer constraints. Instead of using its solution as a benchmark, we use the solution to its continuous relaxation (which is a lower bound), which we denote as:
\begin{equation}\label{linearprogram}
R^*_{p}(\boldsymbol{\sigma}) = \quad \min_{\boldsymbol n \in \mathbb{R}_+^G} R_p(\boldsymbol n; \boldsymbol{\sigma}) \quad s.t. \quad \sum_{g \in [G]}n_g = T.
\end{equation}

Let $\boldsymbol{n}^*_T$ denote the minimizer of the optimization program in Equation \eqref{linearprogram}. (We will show later in Lemma \ref{lemma:completeinformation} that $\boldsymbol{n}^*_T$ exists and is unique.)
Thus, $R_p(\boldsymbol{n}^*_T;\boldsymbol{\sigma}) = R^*_p(\boldsymbol{\sigma})$. We define the normalized regret of a policy $\boldsymbol{\mathcal{\pi}}$ on distributions $\boldsymbol{\mathcal{D}}$ with norm parameter $p$ and time horizon $T$ as:
\begin{equation}\label{equation:definitionregret}
    \text{Regret}_{p,T}(\boldsymbol{\pi}, \boldsymbol{\mathcal{D}}) := \frac{\mathbb{E}_{\boldsymbol{n}_T \sim (\boldsymbol{\pi}, \boldsymbol{\mathcal{D}})}\left[R_p(\boldsymbol{n}_T;\boldsymbol{\sigma})\right] - R^*_p(\boldsymbol{\sigma})}{R^*_p(\boldsymbol{\sigma})}.
\end{equation}
While \textit{normalized} regret is different than usual regret $\mathbb{E}_{\boldsymbol{n}_T \sim (\boldsymbol{\pi}, \boldsymbol{\mathcal{D}})}\left[R_p(\boldsymbol{n}_T;\boldsymbol{\sigma})\right] - R^*_p(\boldsymbol{\sigma})$, one can be recovered from the other by multiplying by a factor $R^*_p(\boldsymbol{\sigma})$ independent of the policy $\boldsymbol{\pi}$, for which we have a closed form, as given in Lemma \ref{lemma:completeinformation}. Note that in $\text{Regret}_{p,T}(\boldsymbol{\pi}, \boldsymbol{\mathcal{D}})$, the randomness lies in the interaction between the policy $\boldsymbol{\pi}$ and the instance $\boldsymbol{\mathcal{D}}$. Given $(\boldsymbol{\pi}, \boldsymbol{\mathcal{D}})$,  the only relevant random variable for calculating regret is $\boldsymbol{n}_T$, the vector of number of samples collected per group at the end of the time horizon $T$.

\begin{restatable}[Complete information optimal policy]{lemma}{completeinformation}\label{lemma:completeinformation}
        For any norm parameter $p \in [1, +\infty]$,
    \begin{equation*}
    R^*_p(\boldsymbol{\sigma}) = \frac{1}{T}\left(\sum_{g \in [G]} \sigma_g^{\frac{2p}{p+1}}\right)^{1 + \frac{1}{p}}.
    \end{equation*}
    Additionly, $R_p(\boldsymbol{n};\boldsymbol{\sigma}) = R^*_p(\boldsymbol{\sigma})$ uniquely at $\boldsymbol{n}^*_T := T  \left(\frac{\sigma_g^{\frac{2p}{p+1}}}{\sum_{h \in [G]}\sigma_h^{\frac{2p}{p+1}}}\right)_{g \in [G]}$.
\end{restatable}
Lemma \ref{lemma:completeinformation} shows that if the data distributions $\boldsymbol{\mathcal{D}}$ are fully known, then the optimal sampling policy has a simple closed form. Each coordinate $n^*_{g,T}$, corresponding to the number of times group $g$ is selected under the optimal complete information policy, is proportional to $\sigma^{\frac{2p}{p+1}}_g$, which is an increasing function of $\sigma_g$. This is to be expected, since groups with higher variance will require more samples to account for the additional uncertainty. As a special case, when $p = 1$, we recover \textit{Neyman's allocation rule} \citep{neyman1992two}, which is the allocation that minimizes the total variance across groups. 

\section{Variance-UCB algorithm}\label{section:algorithm}

In this section, we present our main algorithm for solving our active learning problem, Variance-UCB. As an input, Variance-UCB will utilize an \textit{exogenous} UCB procedure to generate confidence bounds on variances  that are accurate with 
high probability. In this way, Variance-UCB separates the data acquisition task (handled by Variance-UCB) from the estimation task (handled by the UCB subroutine).

Informally, a UCB-procedure is an exogenous black-box that generates upper confidence bounds on the standard deviations $\sigma_g^2$ using the observed history and known hypothesis class $\mathcal{H}$.  
\begin{definition}[UCB-procedure]\label{definition:UCB}
    A double-indexed sequence of functions $\boldsymbol{{\sf UCB}} = ({\sf UCB}_{g,t})_{g \in [G], t \geq 0}$ is a UCB-procedure if for each $g \in [G]$ and $t \geq 0$, the function ${\sf UCB}_{g,t}$ maps the $n_{g,t}$ collected samples and the hypothesis set $\mathcal{H}$ to a real number.
\end{definition}
We do not a priori constrain the behavior or performance of $\boldsymbol{{\sf UCB}}$ because as we will show in Section \ref{subsection:width}, the performance of Variance-UCB depends directly on the accuracy of estimation of $\boldsymbol{{\sf UCB}}$. This has the added benefit that if improved estimation techniques are developed in the literature, these methods can be immediately used in Variance-UCB to improve the performance of our algorithm. In practice, the implementation of UCB procedures typically leverages established concentration inequalities or incorporates specific domain expertise. 
The literature on machine learning has developed a number of UCB-style algorithms with good performance guarantees that can be leveraged; we illustrate specific concentration-bound-based UCB procedures through various examples in Section \ref{section:applications}.

\subsection{The algorithm}\label{section:algorithmsub}
Given a UCB-procedure $\boldsymbol{{\sf UCB}}$, at each time $t \geq 1$, Variance-UCB (Algorithm \ref{algorithmucb}) selects the group with the highest ratio $\frac{\left({\sf UCB}_{g,t}\right)^{\frac{2p}{p+1}}}{n_{g,t}}$. That is,
\begin{equation}\label{equation:selectionrule}
    X_{t + 1} = \text{argmax}_{g \in [G]} \frac{\left({\sf UCB}_{g,t}\right)^{\frac{2p}{p+1}}}{n_{g,t}},
\end{equation}
where ties in the argmax can be broken arbitrarily. Upon observing a new sample $Y_{t+1} \sim \mathcal{D}_{X_{t+1}}$, the algorithm moves to time step $t+1$ and updates the UCB estimator to ${\sf UCB}_{t+1}$.
\begin{algorithm}[htb]\label{algorithmucb}
\caption{Variance-UCB $(p, T, \boldsymbol{{\sf UCB}})$}
\textbf{Input:} norm parameter $p$, time horizon $T$, a UCB-procedure $\boldsymbol{{\sf UCB}}$.

\begin{algorithmic}
\STATE Initialize $n_{g,0} = 0$, $\forall g \in [G]$.

\For{$t = 0, \ldots, T-1$}{
    \STATE Calculate ${\sf UCB}_{g,t}$ \quad $\forall g \in [G]$.
    \STATE Select group $X_{t+1} = \text{argmax}_{g} \frac{\left({\sf UCB}_{g,t}\right)^{\frac{2p}{p+1}}}{n_{g,t}}$.
    \STATE Observe feedback $Y_{t+1} \sim \mathcal{D}_{X_{t+1}}$.
    \STATE Update the number of samples for group $X_t$: $n_{X_{t},t+1} = n_{X_{t},t} + 1$,
    \STATE Update the mean estimate for group $X_t$: $     \hat{\mu}_{X_t,t+1} = \frac{n_{X_{t}, t+1}}{1 + n_{X_{t}, t+1}}\hat{\mu}_{X_t, t+1} + \frac{1}{1 + n_{X_{t}, t+1}}X_{t}$.
}
\end{algorithmic}

\textbf{Output:} $\hat{\mu}_{g,T}, \quad  \forall g \in [G]$
\end{algorithm}

To understand the intuition behind Variance-UCB, assume the algorithm has access to a perfect UCB-procedure that returns exactly $\boldsymbol{\sigma}^2$. It is easy to see that in this case, Variance-UCB chooses groups to equalize the ratios $\frac{\sigma_g^{\frac{2p}{p+1}}}{n_{g,t}}$, which (up to a constant error in $\boldsymbol{n}$) is the same as sampling group $g$ proportionally to $\sigma_g^{\frac{2p}{p+1}}$. This corresponds to the complete information optimal policy stated in Lemma \ref{lemma:completeinformation}. Variance-UCB mimics the behavior of this optimal policy using estimates of $\boldsymbol{\sigma}^2$ from $\boldsymbol{{\sf UCB}}$.

\section{Regret analysis for Variance-UCB}\label{subsection:width}

In this section we present a general regret guarantee for Variance‑UCB -- one that holds for any problem parameters, any distributions $\boldsymbol{\mathcal{D}}$, and any hypothesis class $\mathcal{H}$. To do so, we first establish two technical tools in Section \ref{subsection:decisionerror}: \emph{admissible widths}, which capture how confidence estimates tighten as more data are collected, and \emph{decision errors}, which translate those confidence bounds into performance of the estimator as measured by $R_p(\boldsymbol{n}, \boldsymbol{\sigma})$. With these concepts in hand, we then state our two main theorems in Section \ref{subsection:bounds} -- Theorem \ref{theorem:upperboundinfinite} for the infinite norm and Theorem \ref{theorem:upperboundfinite} for finite $p$ norms -- and show how each term in the regret bound reflects a trade‑off between estimation accuracy and allocation mismatch.

\subsection{Admissible width and decision error}\label{subsection:decisionerror}

The performance of Variance-UCB depends critically on the quality of the upper confidence bound procedure $\boldsymbol{{\sf UCB}}$. Specifically, it hinges on the gap between the estimated upper bound $\boldsymbol{{\sf UCB}}$ and the true standard deviation $\boldsymbol{\sigma}$. Any regret bound should naturally depend on the distribution of the (data-dependent) random variable $\boldsymbol{{\sf UCB}} - \boldsymbol{\sigma}$, as it captures the estimation error in the learning process. The goal of this section is to develop minimal yet expressive tools to characterize how such uncertainty translates into regret. To that end, we introduce the notion of \emph{admissible widths} -- deterministic functions that quantify tolerance to uncertainty in $\boldsymbol{\sigma}$ arising from $\boldsymbol{{\sf UCB}}$:

\begin{definition}[Admissible width]\label{definition:width} For each $g \in [G]$, let $\textsf{w}_g$ be a function from $\mathbb{R}_+$ to $\mathbb{R}_+$. We say that $\boldsymbol{{\sf w}} = \{\textsf{w}_g\}_{g \in [G]}$ is an \textit{admissible width} if each $\textsf{w}_g$ is deterministic, decreasing, differentiable, convex, and goes to $0$ at $+\infty$.
\end{definition}

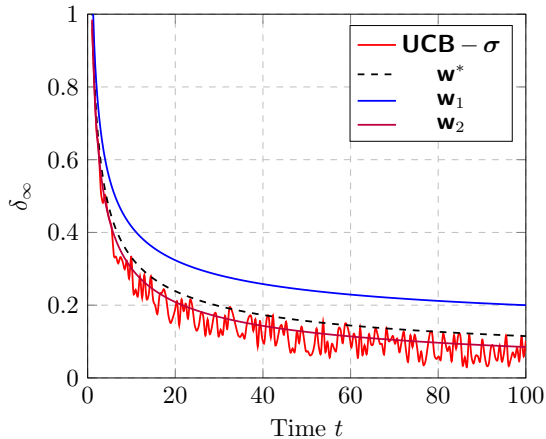
\begin{figure}[h]
    \centering
    \scalebox{0.85}{
        \begin{tikzpicture}
    \begin{axis}[
            xlabel={Time $t$},
            ylabel={$\delta_\infty$},
            xmin=0, xmax=100,
            ymin=0, ymax=1,
            domain=1:100,
            samples=200,
            legend pos=north east,
            every axis plot/.append style={thick},
            ymajorgrids=true,
            xmajorgrids=true,
            grid style=dashed,
        ]

        \addplot+[
            no markers,
            smooth,
            color=red,
        ] (x, {1/sqrt(x) - 0.03 + 0.05*rand});
        \addlegendentry{$\boldsymbol{{\sf UCB}} - \boldsymbol{\sigma}$}

        \addplot+[
            no markers,
            smooth,
            color=black,
            dashed,
        ] (x, {1/sqrt(x) + 0.015});
        \addlegendentry{$\boldsymbol{{\sf w}}^*$}

        \addplot+[
            no markers,
            smooth,
            color=blue,
        ] (x, {1/sqrt(x) + 0.1});
        \addlegendentry{$\boldsymbol{{\sf w}}_1$}

        \addplot+[
            no markers,
            smooth,
            color=purple,
        ] (x, {1/sqrt(x) - 0.015});
        \addlegendentry{$\boldsymbol{{\sf w}}_2$}

    \end{axis}
\end{tikzpicture}
    }
    \caption{\textnormal{Illustration of admissible widths for bounding the random gap $\boldsymbol{{\sf UCB}} - \boldsymbol{\sigma}$ (red). The larger function  $\boldsymbol{{\sf w}}_1$ (blue) yields valid but loose bounds; the smaller (tighter) function $\boldsymbol{{\sf w}}_2$ (purple) underestimates uncertainty and may violate validity. The ideal width $\boldsymbol{{\sf w}}^*$ (black, dashed) provides a convexified, data-agnostic upper bound that closely tracks the average behavior of the random gap.}}
    \label{figure:algorithm}
\end{figure}

The admissible width $\boldsymbol{{\sf w}}$ quantifies how uncertainty about the true parameter $\boldsymbol{\sigma}$ decreases with sample size. Intuitively, if the admissible width is chosen to be too small, it becomes overly restrictive and thus uninformative (since it does not realistically reflect the variability in $\boldsymbol{{\sf UCB}}$). Conversely, admissible width that is too large becomes vacuous, providing little insight or practical utility. This is illustrated in Figure \ref{figure:algorithm}

The assumptions on $\boldsymbol{{\sf w}}$ are natural. The decreasing assumption captures the idea that more independent samples should improve the confidence bounds, and the convexity assumption captures the idea that the marginal value of samples decreases over time. The convergence to $0$ at infinity captures the idea that infinite samples provide a perfect estimation. The differentiability assumption is made to simplify the proofs, and all the results still hold under minimal regularity conditions. In practice, $ \boldsymbol{{\sf w}}$ might come from a readily available concentration inequality given some knowledge about the data distributions (see Section \ref{section:applications} for examples).

A central challenge in this setting is that regret is not additive in the number of suboptimal decisions; instead, it is determined by the curvature of the performance function $R_p$, which governs how deviations from the optimal allocation affect overall performance. The goal is therefore to define an error measure that is both data-dependent and aligned with the structure of $R_p(\boldsymbol{n}, \boldsymbol{\sigma})$, enabling precise control of regret through admissible approximations of uncertainty. To address this challenge, we introduce the notion of \textit{decision error}, denoted $\bar{w}_p$, which captures how uncertainty in estimation translates into suboptimal data collection.

The decision error $\bar{w}_p$ is chosen so that it quantifies precisely the effect of estimation error -- modeled through an admissible width $\boldsymbol{{\sf w}}$ -- on the performance metric $R_p(\boldsymbol{n}, \boldsymbol{\sigma})$. In contrast to classical regret formulations, performance loss in this setting does not accumulate additively with the number of suboptimal decisions. Instead, it depends on how estimation error alters the allocation relative to the optimal one. 

To formalize this, we consider the performance incurred under the optimal allocation $\boldsymbol{n}^*_T$. For each group $g$, collecting $n^*_{g,T}$ samples leads to a width of at most ${\sf w}_g(n^*_{g,T} - 1)$, representing the deviation between the estimated and true standard deviation. Since the optimal policy should sample each group proportionally to $\sigma_g^{\frac{2p}{p+1}}$ (see Lemma \ref{lemma:completeinformation}), the corresponding \emph{relative increase per group $g$} in regret should be captured by the term 
$$\frac{\left(\sf UCB_{g,t}\right)^{\frac{2p}{p+1}} - \sigma_g^{\frac{2p}{p+1}}}{\sigma_g^{\frac{2p}{p+1}}} \leq \left(1 + {\sf w}_g(n^*_{g,T} - 1)\right)^{\frac{2p}{p+1}}- 1.$$
The overall decision error $\bar{w}_p$ then aggregates these group-level contributions in two ways: first, via a weighted average across groups (denoted $w_{1,p}$), where each term is scaled by its optimal sampling frequency $\frac{n^*_{g,T}}{T}$, and second, through a second-moment aggregation (denoted $w_{2,p}$), which accounts for the dispersion of these contributions. The combined measure $\bar{w}_p$ thus captures both the typical and the worst-case deviation effects, providing a principled, instance-dependent complexity metric for any given admissible width. This is formally stated in the definition below.

\begin{definition}[decision error]\label{definition:uncertainty}
    For any pair $(\boldsymbol{\sigma}, \boldsymbol{{\sf w}})$ and norm parameter $p \geq 1$, the \textit{decision error} induced by the pair $(\boldsymbol{\sigma}, \boldsymbol{\sf w})$ is defined as
    \begin{equation*}
        \bar{w}_p(\boldsymbol{\sigma},  \boldsymbol{{\sf w}}) := \sqrt{(w_{1,p}(\boldsymbol{\sigma},  \boldsymbol{{\sf w}}))^2 + \mathbbm{1}_{p < +\infty}  (w_{2,p}(\boldsymbol{\sigma},  \boldsymbol{{\sf w}}))^2},
    \end{equation*}
    where $w_{1,p}$ and $w_{2,p}$ are the following expressions
    we introduce the following quantities 
    \begin{align*}
   {w}_{1,p}(\boldsymbol{\sigma},  \boldsymbol{{\sf w}}) &:= \sum_{g \in [G]} \frac{n^*_{g,T}}{T}\left[\left(1 + {\sf w}_g(n^*_{g,T} - 1)\right)^{\frac{2p}{p+1}}- 1\right], \\
   {w}_{2,p}(\boldsymbol{\sigma},  \boldsymbol{{\sf w}}) &:= \sqrt{\sum_{g \in [G]} \frac{n^*_{g,T}}{T}\left(\left(1 + {\sf w}_g(n^*_{g,T} - 1)\right)^{\frac{2p}{p+1}}- 1\right)^2}.
\end{align*}
\end{definition}
When $\boldsymbol{\sigma}$ and $\boldsymbol{{\sf w}}$ are clear from context, we will suppress dependence on them and simply write $\bar{w}_p$.

As a final remark, we highlight a technical distinction between the finite- and infinite-norm settings. The definition of decision error $\bar{w}_p$ includes the second-order term $w_{2,p}$ only when $p < \infty$. This is because for finite norms, the function $R_p(\boldsymbol{n}, \boldsymbol{\sigma})$ exhibits local smoothness around the optimal allocation $\boldsymbol{n}^*_{T}$, which causes deviations in the input to result in second-order (i.e., quadratic) changes in the objective. In contrast, for $p = \infty$, the performance measure lacks this smoothness, and regret varies linearly with the largest group-level deviation. The inclusion of $w_{2,p}$ in the finite-norm case therefore reflects the need to account for curvature in the performance landscape.

\subsection{Main Regret Bounds}\label{subsection:bounds}

We are now ready to state our main results, Theorems \ref{theorem:upperboundinfinite} and \ref{theorem:upperboundfinite}. We start with the case of $p = +\infty$ in Theorem \ref{theorem:upperboundinfinite}. The proofs of both results are presented in Section \ref{section:proof}.
\begin{restatable}[Regret bounds for the infinite norm]{theorem}{upperboundinfinite}\label{theorem:upperboundinfinite}
   For any instance $\boldsymbol{\mathcal{D}} \in \mathcal{H}$ with variance vector $\boldsymbol{\sigma}^2$, any  UCB-procedure $\boldsymbol{{\sf UCB}}$, and any admissible width $\boldsymbol{{\sf w}}$, Algorithm \ref{algorithmucb} Variance-UCB($+\infty,T,\boldsymbol{{\sf UCB}}$) satisfies
    \begin{equation*}
        \textnormal{Regret}_{\infty,T}(\textnormal{V-UCB}, \boldsymbol{\mathcal{D}}|\mathcal{H}) \leq \mathbb{P}\left[\left(\boldsymbol{0} \leq \boldsymbol{{\sf UCB}} - \boldsymbol{\sigma} \leq \boldsymbol{\sigma}\boldsymbol{{\sf w}}\right)^{\sf c}\right] T + (1 + o(1))\left(\bar{w}_\infty(\boldsymbol{\sigma}, \boldsymbol{{\sf w}}) + \frac{G}{T}\right).
    \end{equation*}
\end{restatable}

Theorem \ref{theorem:upperboundinfinite} holds for any choice of UCB-procedure and admissible width, and most importantly holds for arbitrary distributions $\boldsymbol{\mathcal{D}}$ and hypothesis class $\mathcal{H}$. As such, the bound is fully generic but not directly interpretable in the standard sense, since it lacks explicit dependence on underlying distributional parameters. Concrete instantiations for specific distribution families are presented in Section \ref{section:applications}.

The regret bound consists of two terms. The first term, $\mathbb{P}\left[\left(\boldsymbol{0} \leq \boldsymbol{{\sf UCB}} - \boldsymbol{\sigma} \leq \boldsymbol{\sigma}\boldsymbol{{\sf w}}\right)^{\sf c}\right] T$, captures the contribution from realizations where the admissible width $\boldsymbol{{\sf w}}$ fails to upper bound the random gap $\boldsymbol{{\sf UCB}} - \boldsymbol{\sigma}$. Although the algorithm itself remains well-defined and produces decisions at every round, this term reflects the cost incurred when its decision-making is based on inaccurate confidence estimates. It is therefore controlled by ensuring that $\boldsymbol{{\sf w}}$ bounds the gap with high probability.

The second term includes two components. The rounding term $\frac{G}{T}$ arises from the relaxation used to define the optimal allocation $\boldsymbol{n}^*$, which may not be integer-valued, resulting in at most $1/T$ suboptimality per group. The dominant term is $\bar{w}_\infty(\boldsymbol{\sigma}, \boldsymbol{{\sf w}})$, which captures the decision error introduced by acting under uncertainty bounded by $\boldsymbol{{\sf w}}$. As shown earlier, this quantity aggregates the regret contributions from all groups and reflects how the admissible width interacts with the problem instance. In the high-probability regime where the first term vanishes, $\bar{w}_\infty$ becomes the primary determinant of performance.

Since Theorem \ref{theorem:upperboundinfinite} holds for all $\boldsymbol{{\sf w}}$, then the regret can be improved with an optimal choice of $\boldsymbol{{\sf w}}$, and the bound can be written as:
    \begin{equation*}
        \textnormal{Regret}_{\infty,T}(\textnormal{V-UCB}, \boldsymbol{\mathcal{D}}|\mathcal{H}) \leq \inf_{\boldsymbol{{\sf w}}}\left\{\underbrace{\mathbb{P}\left[\left(\boldsymbol{0} \leq \boldsymbol{{\sf UCB}} - \boldsymbol{\sigma} \leq \boldsymbol{\sigma}\boldsymbol{{\sf w}}\right)^{\sf c}\right] T}_{\text{Information error term}} + \underbrace{(1 + o(1))\left(\bar{w}_\infty(\boldsymbol{\sigma}, \boldsymbol{{\sf w}}) + \frac{G}{T}\right)}_{\text{Decision error term}}\right\}.
    \end{equation*}
    
    This expression highlights the fundamental trade-off controlled by the choice of $\boldsymbol{{\sf w}}$. When ${\boldsymbol{{\sf w}}}$ is too small, the induced decision error term will also be small, but the failure probability $\mathbb{P}\left[\left(\boldsymbol{0} \leq \boldsymbol{{\sf UCB}} - \boldsymbol{\sigma} \leq \boldsymbol{\sigma}\boldsymbol{{\sf w}}\right)^{\sf c}\right]$ will be large, yielding a large information error term. On the other hand, if ${\boldsymbol{{\sf w}}}$ is chosen to be too large, the failure probability $\mathbb{P}\left[\left(\boldsymbol{0} \leq \boldsymbol{{\sf UCB}} - \boldsymbol{\sigma} \leq \boldsymbol{\sigma}\boldsymbol{{\sf w}}\right)^{\sf c}\right]$ will be small, but the induced decision cost $\bar{w}_{\infty}$ will increase, resulting in a high decision error term. The fundamental trade-off of the problem is to balance between the informativeness of the upper confidence bounds (represented by $\mathbb{P}\left[\left(\boldsymbol{0} \leq \boldsymbol{{\sf UCB}} - \boldsymbol{\sigma} \leq \boldsymbol{\sigma}\boldsymbol{{\sf w}}\right)^{\sf c}\right]$) and the decision error (represented by $\bar{w}(\boldsymbol{\sigma}, \boldsymbol{{\sf w}})$). 

    In practice, UCB-procedures with established high-probability guarantees (e.g., via concentration inequalities) should be used to instantiate $\boldsymbol{{\sf UCB}}$, enabling the design of an admissible width $\boldsymbol{{\sf w}}$ such that the failure term $\mathbb{P}\left[\left(\boldsymbol{0} \leq \boldsymbol{{\sf UCB}} - \boldsymbol{\sigma} \leq \boldsymbol{\sigma}\boldsymbol{{\sf w}}\right)^{\sf c}\right] T$ remains controlled.

    Existing bounds in the literature typically provide high-probability guarantees on this type of deviation, often decaying at a polynomial or sub-exponential rate in $T$. In such cases, achieving a small failure probability only requires a modest (e.g., logarithmic) increase in the width, and the resulting contribution to regret becomes asymptotically negligible in $T$ -- that is, it does not dominate the leading-order behavior. As a result, the term $\bar{w}_\infty + \frac{G}{T}$ captures the leading-order behavior of regret and serves as the principal driver of performance in the asymptotic regime.
    Under these conditions (which are also satisfied in the examples in Section~\ref{section:applications}), Theorem~\ref{theorem:upperboundinfinite} reduces to the simplified bound:
    
\begin{equation}\label{equation:regretinftyasymptotic}
    \textnormal{Regret}_{\infty,T}(\textnormal{V-UCB}, \boldsymbol{\mathcal{D}}|\mathcal{H}) \leq \left(1 + o(1)\right)\left(\bar{w}_\infty\left(\boldsymbol{{\sf w}},\boldsymbol{\mathcal{D}}\right)+ \frac{G}{T}\right).
\end{equation}

Next, we can state our second main result Theorem \ref{theorem:upperboundfinite}, which extends the regret analysis to include finite $p$-norms for $p\in(1,\infty)$.

\begin{restatable}[Regret bound for finite $p$-norms]{theorem}{upperboundfinite}\label{theorem:upperboundfinite}
For any instance $\boldsymbol{\mathcal{D}} \in \mathcal{H}^G$, any  UCB-procedure $\boldsymbol{{\sf UCB}}$, and any admissible width ${\sf w}$, the policy Variance-UCB instantiated at $\boldsymbol{{\sf UCB}}$ satisfies
    \begin{equation*}
        \textnormal{Regret}_{p,T}(\textnormal{V-UCB}, \boldsymbol{\mathcal{D}}|\mathcal{H}) \leq \mathbb{P}\left(\left(\boldsymbol{0} \leq \boldsymbol{{\sf UCB}} - \boldsymbol{\sigma} \leq \boldsymbol{\sigma}{\boldsymbol{\sf w}}\right)^{\sf c}\right)T + \frac{p+1}{2}(1 + o(1))\left(\bar{w}^2_p + \frac{G}{T}\right).
    \end{equation*}
\end{restatable}

Theorem~\ref{theorem:upperboundfinite} provides a regret bound analogous to the infinite-norm case, with an important distinction arising from the smoothness of the objective $R_p(\boldsymbol{n}_T; \boldsymbol{\sigma})$ for finite $p$. Specifically, in the finite-$p$ setting, $R_p$ is locally quadratic around the optimal allocation $\boldsymbol{n}^*_T$. Consequently, deviations from optimality induce changes in regret that scale quadratically with the decision error, captured in the the new $\bar{w}_p^2$ term . This additional smoothness yields more precise local approximations, but also introduces a multiplicative factor of $(p+1)/2$, stemming from the second-order Taylor expansion of the objective. As before, standard high-probability arguments ensure the failure-probability term remains negligible, leaving the dominant regret contributions explicitly characterized by these second-order and rounding effects:

\begin{equation}\label{equation:regretfiniteasymptotic}
    \textnormal{Regret}_{p,T}(\textnormal{V-UCB}, \boldsymbol{\mathcal{D}}|\mathcal{H}) \leq \frac{p+1}{2}\left(1 + o(1)\right)\left(\bar{w}_p^2\left(\boldsymbol{{\sf w}},\boldsymbol{\mathcal{D}}\right)+\frac{G}{T}\right).
\end{equation}

\section{Instantiation of regret bounds for specific families of distributions}\label{section:applications}

In this section, we illustrate how to instantiate the general regret bounds of Theorems~\ref{theorem:upperboundinfinite} and~\ref{theorem:upperboundfinite} for common classes of distributional assumptions, such as sub-Gaussian, Gaussian, and exponential feedback. These instantiations help to interpret the fully general statements in Section~\ref{subsection:width} in terms of explicit problem parameters. 
We provide concrete examples of admissible widths and corresponding UCB-procedures from the literature, yielding explicit and interpretable regret bounds that depend directly on known distributional parameters. Such instantiations are critical for practical implementation and enable direct comparison with prior work. We also demonstrate how our analysis extends beyond classical assumptions, providing results even when the distributions exhibit heavier tails, as in the exponential case. All omitted proofs from this section are presented in Appendix~\ref{appendix:finalbounds}.

\subsection{Sub-Gaussian feedback}
  Sub-Gaussian distributions are known to satisfy a fast tail-decay behavior. In the case where the distributions $\mathcal{D}_g$ are sub-Gaussian, this gives ad hoc guarantees on the sample variance that is controlled by how quickly the tail decays. A consequence is that sub-Gaussian distributions satisfy Assumption \ref{assumption:subgaussian}.
  
    \begin{assumption}[Fast decay assumption]\label{assumption:subgaussian}
There exists a known positive real valued vector $\hat{\boldsymbol{c}} \in \mathbb{R}^G_+$ such that for each $g \in [G]$ and $\epsilon > 0$,
\begin{equation*}
    \mathbb{P}\left(\bigcap_{g \in [G], t \in [T]}|\hat{\sigma}_{g,t} - \sigma_g| \leq \hat{c}_g \sqrt{\frac{\log \frac{1}{\epsilon}}{n_{g,t}}}\right) \geq 1 - GT\epsilon.
\end{equation*}
\end{assumption}
 The $GT$ factor in front of $\epsilon$ should be thought of as a result of union bound. Intuitively, $c_g$ captures the trade-off between accuracy of the upper confidence bound and confidence in the estimate for distribution $\mathcal{D}_g$. Assumption \ref{assumption:subgaussian} is satisfied for a large class of distributions, including sub-Gaussian distributions (see Lemma \ref{lemma:subgaussian}). 
 
 \begin{restatable}[Sub-Gaussian feedback]{theorem}{subgaussian}\label{theorem:subgaussian}
Let $\mathcal{H}$ be the set of sub-Gaussian distributions with known parameters $\hat{\boldsymbol{c}} \geq \boldsymbol{\sigma}$. Consider the UCB-procedure defined by the estimator:
\[
\textsf{UCB}_{g,t}^{\textsf{sub-G}} := \hat{\sigma}_{g,t} + \hat{c}_g \sqrt{\frac{3 \log T}{n_{g,t}}},
\]
where $\hat{\sigma}_{g,t}$ is the sample standard deviation computed from the $n_{g,t}$ observations of group $g$ collected up to round $t$. Instantiating Variance-UCB iwith $\textsf{UCB}_{g,t}^{\textsf{sub-G}}$, yields the following the regret bound for the infinite norm:
\begin{equation*}
    \textnormal{Regret}_{\infty,T}(\textnormal{V-UCB}, \boldsymbol{\mathcal{D}}|\mathcal{H}) \leq 4\sqrt{3}(1+o(1))\frac{\|\hat{\boldsymbol{c}}\|_2}{\|\boldsymbol{\sigma}\|_2}\sqrt{\frac{G \log T}{T}},
\end{equation*}
and for any finite $p < +\infty$:
\begin{equation*}
    \textnormal{Regret}_{p,T}(\textnormal{V-UCB}, \boldsymbol{\mathcal{D}}|\mathcal{H}) \leq 85p(1 + o(1))\left(\sum_{g \in [G]} \frac{\hat{c}_g^2}{\sigma_g^2}\right)\frac{\log T}{T}.
\end{equation*}
\end{restatable}

The bound for the infinite norm case given in Theorem \ref{theorem:subgaussian} significantly improves the previously best known bound for $p = +\infty$ from the B-AS algorithm in \cite{carpentier2011upper}, which is
\begin{equation*}
     \textnormal{Regret}_{\infty,T}(\text{B-AS}, \boldsymbol{\mathcal{D}}|\mathcal{H}) \leq \frac{76400(1 + o(1))\|\hat{\boldsymbol{c}}\|_\infty G^2 \log^2 T}{\sigma_{\min}^2\sqrt{T}}.
\end{equation*}
We improve upon this bound in several ways --- including the dependence on $G, T, \|\boldsymbol{\sigma}\|$, $\hat{\boldsymbol{c}}$), and the numerical constant --- but our most significant improvement comes from providing a regret bound that does not scale with $(\sigma_{\min})^{-1}$. Thus we solve the open question posed by \cite{carpentier2011upper}. Corollary \ref{corollary:subgaussian} is stated in additive regret for ease of comparison to the prior work.

\begin{corollary}[Open question in \cite{carpentier2011upper}]\label{corollary:subgaussian}
        For the choice $\textsf{UCB}^{\textsf{sub-G}} = \hat{\sigma}_{g,t} + \hat{c}_g \sqrt{\frac{3 \log T}{n_{g,t}}}$ with sub-Gaussian feedback and variances bounded by $\hat{\boldsymbol{c}} \geq \boldsymbol{\sigma}$, 
        \begin{equation*}
            \mathbb{E}_{\boldsymbol{\pi}}[R_p(\boldsymbol{n};\boldsymbol{\sigma})] - R^*_p(\boldsymbol{\sigma}) \leq \frac{4\sqrt{3}(1 + o(1))\|\boldsymbol{\sigma}\|_2\|\hat{\boldsymbol{c}}\|_2\sqrt{G}}{T^{3/2}}.
        \end{equation*}
        In particular, the regret does not depend on $\sigma_{\min}^{-1}$.
\end{corollary}
\begin{proof}{\textit{Proof.}}
\begin{align*}
    \mathbb{E}_{\boldsymbol{\pi}}[R_p(\boldsymbol{n};\boldsymbol{\sigma})] - R^*_p(\boldsymbol{\sigma}) &= R^*_p(\boldsymbol{\sigma})\textnormal{Regret}_{\infty,T}(\text{V-UCB}, \boldsymbol{\mathcal{D}}|\mathcal{H}) \\ &\leq \frac{\|\boldsymbol{\sigma}\|^2_2}{T}\cdot 4\sqrt{3}(1+o(1))\frac{\|\hat{\boldsymbol{c}}\|_2}{\|\boldsymbol{\sigma}\|_2}\sqrt{\frac{G \log T}{T}} \\ &= \frac{4\sqrt{3}(1 + o(1))\|\boldsymbol{\sigma}\|_2\|\hat{\boldsymbol{c}}\|_2\sqrt{G}}{T^{3/2}},
\end{align*}
where the first step follows from the definition of normalized regret (see Equation \eqref{equation:definitionregret}), and the second step follows from $R^*_\infty(\boldsymbol{\sigma}) = \frac{\|\boldsymbol{\sigma}\|_2^2}{T}$ (see Lemma \ref{lemma:completeinformation}) and Theorem \ref{theorem:subgaussian}.
    $\hfill \square$
\end{proof}
\subsection{Gaussian feedback}
   When the distributions $\mathcal{D}_g$ are \textit{exactly} Gaussian, the analyst can leverage this extra information to design a sharper UCB-procedure. This is illustrated in Theorem \ref{theorem:gaussianfeedback}.
   
   \begin{restatable}[Gaussian feedback]{theorem}{gaussian}\label{theorem:gaussianfeedback}
         When $\mathcal{H}$ is the set of Gaussian distributions, the choice $        {\sf UCB}^{\sf G}_{g,t} = \hat{\sigma}_{g,t}\left(1 + \sqrt{\frac{3\log T}{n_{g,t}}} + \frac{3\log T}{n_{g,t}}\right)$, where $\hat{\sigma}_{g,t}^2$ is the sample variance, yields the following regret bound:
       \begin{equation*}
            \textnormal{Regret}_{\infty,T}(\textnormal{V-UCB}, \boldsymbol{\mathcal{D}}|\mathcal{H}) \leq 2\sqrt{3}(1 + o(1))\frac{\|\boldsymbol{\sigma}\|_1}{\|\boldsymbol{\sigma}\|_2}\sqrt{\frac{\log T}{T}}.
       \end{equation*}
       For any $p < +\infty$,
       \begin{equation*}
           \textnormal{Regret}_{p,T}(\textnormal{V-UCB}, \boldsymbol{\mathcal{D}}|\mathcal{H}) \leq 43p(1 + o(1))\frac{G \log T}{T}.
       \end{equation*}
   \end{restatable}

By noticing that $\|\boldsymbol{\sigma}\|_1 \leq \sqrt{G}\|\boldsymbol{\sigma}\|_2$, we arrive at the following worst-case regret bound over Gaussian distributions.

\begin{corollary}\label{corollary:gaussian}
     When $\mathcal{H}$ is the set of Gaussian distributions, using ${\sf UCB}^{\sf G}_{g,t} = \hat{\sigma}_{g,t}\left(1 + \sqrt{\frac{3\log T}{n_{g,t}}} + \frac{3\log T}{n_{g,t}}\right)$ in the Variance-UCB algorithm yields regret:  
     \[\textnormal{Regret}_{\infty,T}(\textnormal{V-UCB}, \boldsymbol{\mathcal{D}}|\mathcal{H}) \leq 2\sqrt{3}(1 + o(1))\sqrt{\frac{G\log T}{T}}.\]
\end{corollary}

We compare the bound in Corollary \ref{corollary:gaussian} for the infinite norm to the previous best known bound for $p=+\infty$ from the B-AS algorithm B-AS in \cite{carpentier2011upper}, which is:
\begin{equation*}
     \textnormal{Regret}_{\infty,T}(\text{B-AS}, \boldsymbol{\mathcal{D}}|\mathcal{H}) \leq 1.05 \times 10^5 \cdot \frac{\hat{\Sigma}}{\|\boldsymbol{\sigma}\|_2^2}\cdot \frac{G \log^2 T}{\sqrt{T}}.
\end{equation*}
The regret bounds in \cite{carpentier2011upper} require specifying  $\hat{\Sigma}$ as a known upper bound on $\|\boldsymbol{\sigma}\|_2^2$. Theorem \ref{theorem:gaussianfeedback} significantly improves upon this bound in several ways. Firstly, it does not require any knowledge on an upper bound on $\|\boldsymbol{\sigma}\|_2$, making it detail-free in $\boldsymbol{\sigma}$. Secondly, it reduces the dependence on $G$ (from $G$ to $\sqrt{G}$), thus tightening the regret bounds in terms of the number of groups $G$. Thirdly, the numerical constant in the leading term is reduced from approximately $10^5$ to $4\sqrt{2}$, representing an improvement of several orders of magnitude.

To highlight the practical impact of our new bounds, we translate them into
\emph{sample--size requirements}.
Starting from Corollary~\ref{corollary:gaussian}, the worst--case regret of
Variance--UCB under Gaussian feedback satisfies
\[
\text{Regret}_{\infty,T}\;\le\; 2\sqrt{3}\,
        \sqrt{\frac{G\log T}{T}}.
\]
Solving this expression for the value of $T$ such that $\text{Regret}_{\infty,T}\;\le\; \varepsilon$ yields 
\[
T_{\text{V-UCB}}(\varepsilon,G)=
\bigl\lceil
12\,G\,\varepsilon^{-2}\,
        \log\!\bigl(12\,G\,\varepsilon^{-2}\bigr)
\bigr\rceil .
\]
We perform the same inversion for the best previously known bound of
B-AS~\cite{carpentier2011upper}, and for the non-adaptive uniform policy, whose worst-case regret is a
fixed constant independent of~\(T\).
The resulting values are reported in Table~\ref{table:samplesg}, which lists \emph{the smallest total number of
samples \(T\)} that guarantees the worst-case relative regret
\(\text{Regret}_{\infty,T}\le\varepsilon\)
under Gaussian feedback, for target relative-regret level \(\varepsilon\in\{0.10,0.05,0.01\}\) and for three representative values of \(G\).
Uniform sampling never meets any non-trivial target because its regret
plateaus; B-AS does converge but requires extremely large \(T\) due to
loose constants and a suboptimal \(G\)-dependence;
Variance-UCB attains the same guarantees with sample sizes that are orders of magnitude smaller, thus making adaptive sampling feasible in practice.

\begin{table}[h]
    \centering
    {\scriptsize
    \renewcommand{\arraystretch}{0.5}
    \setlength{\tabcolsep}{3pt}
    \begin{tabular}{c c c c c c c c c c c c c c c}
        Algorithm & \multirow{2}{*}{Assumptions} & & \multicolumn{3}{c}{$G = 3$} & & \multicolumn{3}{c}{$G = 50$} & & \multicolumn{3}{c}{$G = 1000$} \\
        \cmidrule{4-6} \cmidrule{8-10} \cmidrule{12-14}
        & & & 10\% & 5\% & 1\% & & 10\% & 5\% & 1\% & & 10\% & 5\% & 1\% \\
        \cmidrule{1-14}
        Uniform & - && $\infty$& $\infty$& $\infty$&& $\infty$& $\infty$& $\infty$&& $\infty$& $\infty$&$\infty$\\
        \addlinespace
       B-AS & Knowledge of $\hat{\Sigma}$ && $9.9 \cdot 10^{12}$& $3.9\cdot10^{13}$& $9.9 \cdot 10^{14}$&& $2.7\cdot10^{15}$& $1.1 \cdot 10^{16}$& $2.7 \cdot10^{17}$&& $1.1 \cdot 10^{18}$& $4.4\cdot10^{18}$&$1.1 \cdot 10^{20}$ \\
        \addlinespace
        Variance-UCB & None && $9.6 \cdot 10^2$& $1.9 \cdot 10^3$& $9.6 \cdot 10^3$&& $1.6\cdot 10^4$& $3.2\cdot 10^4$& $1.6\cdot 10^5$&& $3.2\cdot 10^5$& $6.4\cdot 10^5$& $3.2 \cdot 10^6$ \\
        \addlinespace 
    \end{tabular}
    }
    \caption{{\normalfont Worst-case total number of samples $T$ required to achieve a relative regret guarantee of at most $10\%, 5\%$, and $1\%$ for various algorithms (uniform sampling without learning, previous best bound: B-AS in \cite{carpentier2011upper}, our result: V-UCB) for $p=\infty$ and $G \in \{3, 50, 100\}$. $\mathcal{H}$ is the set of Gaussian distributions}}
    \label{table:samplesg}
\end{table}

\subsection{Exponential feedback}
Existing analyses of established prior bounds (e.g., \cite{activelearning,carpentier2011upper}) rely on \emph{fast‑decay} assumptions -- bounded support or sub‑Gaussian tails with a known parameter.  
These analysis methods break down once the tail decays more slowly.  
Our approach, by contrast, depends only on constructing an admissible width that controls the gap
\(\boldsymbol{{\sf UCB}}-\boldsymbol{\sigma}\); it is therefore applicable even when the underlying distributions are not sub‑Gaussian. To demonstrate this flexibility, we next instantiate Theorems~\ref{theorem:upperboundinfinite} and \ref{theorem:upperboundfinite} on the family of exponential distributions.  

\begin{restatable}[Exponential feedback]{theorem}{exponential}\label{theorem:exponential} 
    When $\mathcal{H}$ is the set of exponential distributions, the choice ${\sf UCB}^{\sf exp}_{g,t} = \hat{\mu}_{g,t}\left(1 +  \sqrt{\frac{3\log T}{n_{g,t}}}\right)$, yields the following regret bound:
    \begin{equation*}
       \textnormal{Regret}_{\infty, T}(\textnormal{V-UCB}, \boldsymbol{\mathcal{D}}|\mathcal{H}) \leq 2\sqrt{3}(1 + o(1))\frac{\|\boldsymbol{\sigma}\|_1}{\|\boldsymbol{\sigma}\|_2}\sqrt{\frac{\log T}{T}}.
    \end{equation*}
For any $p < +\infty$,
\begin{equation*}
    \textnormal{Regret}_{p, T}(\textnormal{V-UCB}, \boldsymbol{\mathcal{D}}|\mathcal{H}) \leq 43p(1 + o(1))\frac{G \log T}{T}.
\end{equation*}
\end{restatable}

Theorem \ref{theorem:exponential} provides regret bounds in new settings that were not covered in prior work, since both \cite{carpentier2011upper} and \cite{activelearning} relied on the fast decay properties of bounded/sub-Gaussian distributions to derive their regret bound. 

The main difference between the Gaussian case in Theorem \ref{theorem:gaussianfeedback} and the Exponential case in Theorem \ref{theorem:exponential} lies in the choice of the UCB procedure, where the sample standard deviation $\hat{\sigma}$ (in Theorem \ref{theorem:gaussianfeedback}) is replaced with the sample mean $\hat{\mu}$ (in Theorem \ref{theorem:exponential}). This is because for exponential distributions, the standard deviation vector $\boldsymbol{\sigma}$ is exactly the mean vector $\boldsymbol{\mu}$. Intuitively, the exponential case is $\textnormal{simpler}$ to learn because the set of exponential distributions is well-structured. More generally, combining Theorems \ref{theorem:upperboundinfinite} and \ref{theorem:upperboundfinite} with a UCB-procedure that leverages the structure of the hypothesis class $\mathcal{H}$ will always give small regret bounds regardless of the size of the tails of the distributions in $\mathcal{H}$, as long as they are well-structured.

\section{Proofs of Theorems \ref{theorem:upperboundinfinite} and \ref{theorem:upperboundfinite}}\label{section:proof}
This section offers a sketch of the proofs of Theorems~\ref{theorem:upperboundinfinite} and~\ref{theorem:upperboundfinite}, deferring full technical details of the intermediate results to Appendix \ref{appendix:finalbounds}. In Section \ref{subsection:relativedistance}, we introduce the relative‐distance vector \(\boldsymbol\delta_T\) and invoke Propositions~\ref{theorem:regretscheme} and~\ref{theorem:regretschemefinite} to reduce deriving both the infinite‐norm and finite‑\(p\) regret bounds to high‐probability controls of \(\boldsymbol{\delta}_T\).  The core technical innovation (Lemmas~\ref{lemma:initialpoint}–\ref{prop:improvedrates} in Section \ref{section:bounding}) is a two‐stage bounding of \(\boldsymbol{\delta}_T\): first an initial coarse rate, then an iterative refinement via the potential‐function mapping \(F_T\), whose fixed point yields an optimally tight convergence rate.  Substituting these sharpened distance bounds into the regret scheme immediately delivers the claimed guarantees for \(p=\infty\) in Section \ref{subsection:finalinfinite} and \(p<\infty\) in Section \ref{subsection:finalfinite}.

\subsection{Relative distance}\label{subsection:relativedistance}
The starting point of the proofs of both Theorems \ref{theorem:upperboundinfinite} and \ref{theorem:upperboundfinite} is Lemma \ref{lemma:completeinformation}, which states that the vector $\boldsymbol{n}^*_T$ of the optimal number of samples from each group satsifies
\begin{equation*}
    \boldsymbol{n}^*_T = \frac{T}{\sum_{h \in [G]} \sigma_h^{\frac{2p}{p+1}}}\boldsymbol{\sigma}^{\frac{2p}{p+1}}.
\end{equation*}
In particular, this vector scales linearly in $T$ in all coordinates. The natural implication is that the optimal policy consists of randomizing over a \textit{static} allocation rate $\boldsymbol{\lambda}^*$, indicating the optimal sampling rate from each group, where 
\begin{equation}
    \lambda_g^* := \frac{\sigma_g^{\frac{2p}{p+1}}}{\sum_{h \in [G]}\sigma_h^{\frac{2p}{p+1}}}, \qquad \forall g \in [G].
\end{equation}
The vector $\boldsymbol{\lambda}^*$ will play a central role in the analysis, as Lemma \ref{lemma:completeinformation} can be concisely summarized as the proportionalities
\begin{equation*}
    \boldsymbol{n}^*_T \propto \boldsymbol{\sigma}^{\frac{2p}{p+1}} \propto \boldsymbol{\lambda}^*,
\end{equation*}
where $\propto$ means proportional to. For simplicity, we introduce $\Sigma_p(\boldsymbol{\sigma}) := \sum_{g \in [G]} \sigma_g^{\frac{2p}{p+1}}$, as the normalizing factor in these proportional relationships. Thus the optimal policy in the complete information setting can be fully described as:
\begin{equation}\label{eq:proportionalities}
    \boldsymbol{\lambda}^* = \frac{\boldsymbol{\sigma}^\frac{2p}{p+1}}{\Sigma_p(\boldsymbol{\sigma})}, \qquad \boldsymbol{n}^* = T \boldsymbol{\lambda}^*, \qquad R^*_p(\boldsymbol{\sigma}) = \frac{\Sigma_p^{1+\frac{1}{p}}(\boldsymbol{\sigma})}{T}.
\end{equation}

This suggests that a good policy should try to make $\boldsymbol{n}_T$ as close as possible to $\boldsymbol{n}^*_T$, or equivalently, sample randomly across all groups according to a sampling frequency $\boldsymbol{\lambda}_T$ that is close to the static optimal allocation $\boldsymbol{\lambda}^*$. This goal is fundamentally different from the reward maximization paradigm of the classic bandits framework, where a learner's goal is to identify the single best arm, and typical ``good'' policies (e.g., Thompson sampling, UCB) end up sampling from only one arm as $T\to \infty$. 

In the incomplete information setting, since $\boldsymbol{\sigma}$, and hence $\boldsymbol{\lambda}^*$, is unknown, any policy should incorporate learning, which is done by collecting samples. However, if the learning is performed without the objective in mind (e.g., regret minimization as in Equation \eqref{equation:definitionregret}), then the resulting number of collected samples $\boldsymbol{n}_T$ may be far from $\boldsymbol{n}^*_T$, yielding a poor regret. Therefore, there is an underlying exploration-exploitation challenge, where the analyst has to simultaneously \textit{learn} and \textit{sample close to} an unknown distribution $\boldsymbol{\lambda}^*$. To formalize this idea, we introduce the notion of \textit{relative distance}, expressing closeness between two distributions.

\begin{definition}[Relative distance]\label{definition:distance}
    Let $\boldsymbol{\lambda}^* \in \Delta_G$ be a distribution with $\boldsymbol{\lambda}^* > \boldsymbol{0}_G$. For any distribution $\boldsymbol{\lambda} \in \Delta_G$, define the relative distance of $\boldsymbol{\lambda}$ to $\boldsymbol{\lambda}^*$ as the vector:
    \begin{equation*}
        \boldsymbol{\delta}(\boldsymbol{\lambda}||\boldsymbol{\lambda}^*) := \boldsymbol{1} - \frac{\boldsymbol{\lambda}}{\boldsymbol{\lambda}^*}.
    \end{equation*}
\end{definition}
Notice that
\begin{equation*}
    \boldsymbol{\delta}(\boldsymbol{\lambda}||\boldsymbol{\lambda}^*) = \boldsymbol{0}_G \quad \text{if and only if} \quad \boldsymbol{\lambda} = \boldsymbol{\lambda}^*.
\end{equation*}
Define $\delta_{\max}(\boldsymbol{\lambda}||\boldsymbol{\lambda}^*) := \max_{g \in [G]} \delta_g(\boldsymbol{\lambda}||\boldsymbol{\lambda}^*)$. Also notice that
\begin{equation*}
    \delta_{\max}(\boldsymbol{\lambda}||\boldsymbol{\lambda}^*) = 0 \quad \text{if and only if} \quad  \boldsymbol{\lambda} = \boldsymbol{\lambda}^*.
\end{equation*}
In what follows, we formalize the importance of $\boldsymbol{\lambda}_T$ being close to $\boldsymbol{\lambda}^*$ for the policy to be good. To simplify notation, define $\boldsymbol{\delta}_T = \boldsymbol{\delta}(\boldsymbol{\lambda}_T||\boldsymbol{\lambda}^*)$, and omit the dependence on $\boldsymbol{\lambda}_T$ and $\boldsymbol{\lambda}^*$ when they are clear from context. As a consequence, the regret analysis of any policy can be performed by focusing on the distribution of $\boldsymbol{\delta}_T$, as we show next in Proposition \ref{theorem:regretscheme}.

\begin{restatable}[Regret analysis scheme for the infinite norm]{proposition}{regretscheme}\label{theorem:regretscheme}
   Consider any policy $\boldsymbol{\pi}$ and any instance $\boldsymbol{\mathcal{D}} \in \mathcal{H}^G$ with optimal sampling distribution $\boldsymbol{\lambda}^* \in \Delta_G$. Let $\boldsymbol{\delta}_T$ be the induced relative distance at time $T$, and let $\delta_{\max,T} = \max_{g \in [G]} \delta_{g,T}$. Then,
    \begin{equation*}
    \text{Regret}_{\infty, T}(\boldsymbol{\pi}, \boldsymbol{\mathcal{D}}|\mathcal{H}) \leq \inf_{c > 0} \left\{\mathbb{P}\left(\delta_{\max,T} > c \right) T + \frac{c}{(1 - c)^+} \right\}.
    \end{equation*}
\end{restatable}

To extend Proposition \ref{theorem:regretscheme} to finite $p$-norms, notice that the main idea behind Proposition \ref{theorem:regretscheme} stems from the simple observation that when $\delta_{\max} := \max_{g \in [G]} \delta_g \leq 1$,
\begin{equation*}
\frac{R_\infty(\boldsymbol{n};\boldsymbol{\sigma}) - R^*_\infty(\boldsymbol{\sigma})}{R^*_\infty(\boldsymbol{\sigma})} = \left\|\frac{1}{\boldsymbol{1} - \boldsymbol{\delta}_T}\right\|_\infty - 1 = \frac{\delta_{\max}}{1 - \delta_{\max}},
\end{equation*}
which is an increasing function in $\delta_{\max}$. It is this monotonicity property that reduces the analysis to only upper bounding $\delta_{\max}$ (with high probability). Such properties do not extend to $p < +\infty$, since a Taylor approximation around $\boldsymbol{0}_G$ yields
\begin{equation*}
    \frac{R_p(\boldsymbol{n};\boldsymbol{\sigma}) - R^*_p(\boldsymbol{\sigma})}{R^*_p(\boldsymbol{\sigma})} = \left\|\frac{\boldsymbol{(\lambda}^*)^{\frac{1}{p}}}{1 - \boldsymbol{\delta}_T}\right\|_p - 1 \approx \frac{p+1}{2}\sum_{g \in [G]} \lambda_g^* \delta_{g,T}^2,
\end{equation*}
which is not increasing. Thus, instead of requiring only an inequality of the type $\boldsymbol{\delta}_T \leq c$, we require a stronger double-bounding $\boldsymbol{\delta}_T \in [\boldsymbol{a}, \boldsymbol{b}]$, as stated in Proposition \ref{theorem:regretschemefinite}.
\begin{restatable}[Regret analysis scheme for finite $p$-norms]{proposition}{regretschemefinite}\label{theorem:regretschemefinite} Consider any policy $\boldsymbol{\pi}$ and any instance $\boldsymbol{\mathcal{D}}$ with optimal sampling distribution $\boldsymbol{\lambda}^* \in \Delta_G$. Let $\boldsymbol{\delta}_T$ be the induced relative distance at time $T$. Then,
\begin{equation*}
        \text{Regret}_{p, T}(\boldsymbol{\pi}, \boldsymbol{\mathcal{D}}|\mathcal{H}) \leq \inf_{\boldsymbol{a} \leq \boldsymbol{0}_G \leq \boldsymbol{b}}\left\{\mathbb{P}\left((\boldsymbol{a} \leq \boldsymbol{\delta}_T \leq \boldsymbol{b})^{\sf c}\right) T + \left(\frac{p+1}{2}\sum_{g \in [G]}\lambda_g^*(a_g^2 + b_g^2)\right)\left(1 + \epsilon(\boldsymbol{a}, \boldsymbol{b})\right)\right\},
\end{equation*}
where $\epsilon$ is a function that satisfies $\epsilon(\boldsymbol{a}, \boldsymbol{b}) \to_{(\boldsymbol{a}, \boldsymbol{b})\to \boldsymbol{0}}0$.
\normalsize
\end{restatable}
Propositions \ref{theorem:regretscheme} and \ref{theorem:regretschemefinite} represent the first stepping stone to deriving regret upper bounds. For a specific algorithm (in our case, Variance-UCB), the general strategy for the regret analysis should be to construct a specific $c > 0$ (resp. specific $\boldsymbol{a} \leq \boldsymbol{0}_G\leq \boldsymbol{b}$) that optimize the right hand side of Proposition \ref{theorem:regretscheme} (resp. Theorem
\ref{theorem:regretschemefinite}). Section \ref{section:bounding} next shows how to do this through achieving a high probability bound on the relative distance $\boldsymbol{\delta}_T$.

\subsection{Key argument: bounding the relative distance}\label{section:bounding}

To derive an upper bound on regret, following Propositions \ref{theorem:regretscheme} and \ref{theorem:regretschemefinite}, it suffices to establish high probability bounds on $\boldsymbol{\delta}_T$ (upper bounds for the infinite norm case, and both upper and lower bounds for the finite norm case). First, we identify a high‑probability event under which all UCB estimates stay within their admissible widths. Second, we derive an initial coarse bound on the maximum relative distance, and then refine it via a potential‑function argument that converges to the tight rate.  These bounds on~$\delta_{\max,T}$ immediately translate into high‑probability guarantees on~$\boldsymbol{\delta}_T$, which, combined with Propositions~\ref{theorem:regretscheme} and~\ref{theorem:regretschemefinite}, yield the regret bounds in Theorems~\ref{theorem:upperboundinfinite} and~\ref{theorem:upperboundfinite}. 

We consider the following event for a fixed admissible width $\boldsymbol{{\sf w}}$ (see Definition \ref{definition:width}):
\begin{equation}\label{eq.defineA}
    \mathcal{A} := \left\{\boldsymbol{0}_G \leq \boldsymbol{{\sf UCB}} -\boldsymbol{\sigma} \leq \boldsymbol{\sigma} \boldsymbol{{\sf w}}\right\} = \bigcap_{g \in [G],t\geq1} \{0 \leq {\sf UCB}_{g,t} - \sigma_g \leq \sigma_g {\sf w}_g\},
\end{equation}
and throughout this subsection we condition on~$\mathcal{A}$. Event~$\mathcal{A}$ ensures that the realized estimates never violate the admissible width, allowing us to treat all subsequent inequalities deterministically.

Using the structural properties of Variance-UCB, we start by proving an upper bound on $\delta_{\max,T}$. Following Definition \ref{definition:distance} and the proportionalities in Equation \eqref{eq:proportionalities}, $\delta_{\max,T}$ can be expressed as follows:
\[
\delta_{\max,T}
\;=\; \max_{g \in [G]} \left(1 - \frac{\lambda_{g,T}}{\lambda^*_g}\right)  = \max_{g \in [G]} \left(1 - \frac{T\lambda_{g,T}}{T\lambda_g^*}\right) = \max_{g\in[G]}\left(1 - \frac{n_{g,T}}{n^*_{g,T}}\right)
\] 
Thus $\delta_{\max,T}$ can be interpreted as the {\em relative convergence rate} of the empirical allocation $\boldsymbol{n}_T$ to the optimal allocation $\boldsymbol{n}^*_T$, we first
derive a {\em coarse} bound on this rate. Concretely, Lemma~\ref{lemma:initialpoint} shows that, conditioned on event $\mathcal{A}$,
\[
0\;\le\;\delta_{\max,T}\;\le\;f^0_T,
\] and we call $f^0_T$ the {\em initial rate}. This bound captures the
preliminary speed at which Variance‑UCB approaches the optimal frequencies,
and sets the stage for the finer, potential‑function‑based refinements that
follow.

\begin{restatable}[Initial rate]{lemma}{initialrate}\label{lemma:initialpoint}Conditioned on $\mathcal{A}$, $\delta_{\max, T}$ can be bounded as follows: 
\begin{equation}\label{eq:initial}
    0\leq \delta_{\max,T} \leq f^{0}_T := 1 - \frac{\frac{1}{2}}{\left(1 + \max_{h \in [G]}{\sf w}_h\left(n^*_{h,T/2}\right)\right)^{\frac{2p}{p+1}}}.
\end{equation}
\end{restatable}

Lemma \ref{lemma:initialpoint} gives an initial non-trivial bound on $\delta_{\max,T}$. Since this gives an upper bound on the rate of convergence of the sequence $(\delta_{\max,T})_{T \geq 1}$, we call it an \textit{initial convergence rate}. 
By recalling that $\delta_{\max, T} = \max_{g \in [G]} 1 - \frac{n_{g,T}}{n^*_{g,T}}$, Lemma \ref{lemma:initialpoint} can be rewritten as
\begin{equation*}
    \forall g \in [G], \qquad 1 - \frac{n_{g,T}}{n^*_{g,T}} \leq f^0_T,
\end{equation*}
or equivalently
\begin{equation*}
    \boldsymbol{n}_T \geq (1 - f_T^0)\boldsymbol{n}^*_T.
\end{equation*}

Notice that $f^{0}_T$ converges to $\frac{1}{2}$ as $T \to +\infty$, since $n^*_{h,T/2} \to +\infty$, and ${\sf w}_h(t) \to_{t \to +\infty} 0$ (see Definition \ref{definition:width}), so that
\begin{equation*}
    f^0_T = 1 - \frac{\frac{1}{2}}{\left(1 + \max_{h \in [G]}{\sf w}_h\left(n^*_{h,T/2}\right)\right)^{\frac{2p}{p+1}}} \to_{T \to +\infty} 1 - \frac{\frac{1}{2}}{(1 + 0)^{\frac{2p}{p+1}}} = \frac{1}{2}.
\end{equation*}
Hence, Lemma \ref{lemma:initialpoint} implies that
\begin{equation*}
    \boldsymbol{n}_T \geq \frac{1}{2}(1 + o(1))\boldsymbol{n}^*_T,
\end{equation*}
which is an initial rate on the growth of the sequence $\boldsymbol{n}_T$. 
In particular, since $\boldsymbol{n}^*_T$ grows to $+\infty$, so should $\boldsymbol{n}_T$, and Lemma \ref{lemma:initialpoint} implies that conditioned on $\mathcal{A}$, Variance-UCB eventually chooses each group arbitrarily many times. Setting $c = f^0_T$ in Proposition \ref{theorem:regretscheme} will yield an initial bound on regret, but the bound will be 
\begin{equation*}
\text{Regret}_{\infty, T}(\boldsymbol{\pi}, \boldsymbol{\mathcal{D}}|\mathcal{H}) \leq \mathbb{P}\left(\delta_{\max,T} > f^0_T \right) T + \frac{f^0_T}{(1 - f^0_T)^+}.
\end{equation*}
However, since $f^0_T \to \frac{1}{2}$, the bound above is bounded by below by $\Omega(1)$, which makes it vacuous, since a meaningful regret bound should at least go to $0$ as $T \to +\infty$. In other words, while Lemma \ref{lemma:initialpoint} yields an initial bound on $\delta_{\max, T}$, it is still unsatisfying.

Lemma \ref{lemma:initialpoint} provides an \emph{explicit} bound on $\delta_{\max, T}$, derived from an initial analysis of the structural properties of Variance-UCB. A more refined approach, however, yields a different \emph{implicit} bound on $\delta_{\max, T}$. Although this implicit bound does not directly translate into an explicit bound (and thus cannot independently yield a regret guarantee), it serves as a foundation for constructing a \emph{potential function}. This potential function characterizes candidate regions where $\delta_{\max, T}$ must reside, and when combined with an initial explicit bound, induces a dynamic system. Iterative analysis of this system progressively sharpens localization of $\delta_{\max, T}$, facilitating tighter regret analyses. We refer to this method as \emph{potential bounding}. We call this implicit bounding a \emph{potential bounding}, stated formally in Lemma \ref{lemma:potential} below.

\begin{restatable}[Potential bounding]{lemma}{potential}\label{lemma:potential}
    Conditioned on $\mathcal{A}$, $\delta_{\max,T}$ can be bounded as $$0 \leq \delta_{\max,T} \leq F_T(\delta_{\max,T}),$$ where the function $F_T$ is defined as:
    \begin{equation*}
    F_{T}: u \in [0, 1]  \mapsto 1 - \left(1 - \frac{G}{T}\right)\left[\sum_{h \in [G]}\lambda_h^*\left(1 + {\sf w}_{h}(\lambda^*_{h}T(1 - u) - 1)\right)^{\frac{2p}{p+1}}\right]^{-1}.
\end{equation*}
\end{restatable}

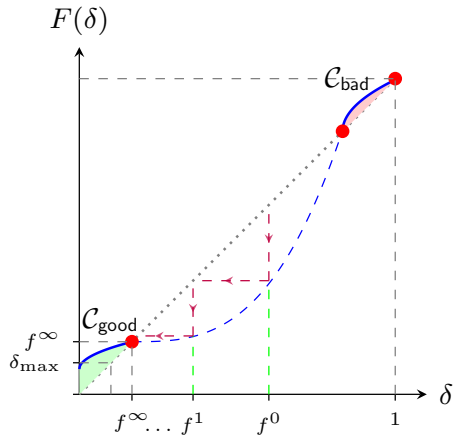
\begin{figure}[ht]
\centering
    \scalebox{1.2}{

    \begin{tikzpicture}[scale=3.5,>=stealth]
\tikzset{label style/.style={font=\tiny}}

  \draw[->] (0,0) -- (1.1,0) node[right] {\footnotesize$\delta$};
  \draw[->] (0,0) -- (0,1.1) node[above] {\footnotesize$F(\delta)$};

  \draw[dotted, thick, gray] (0,0) -- (1,1);

\draw[label style] (1,0) -- (1,-0.02) node[below] {$1$};
\draw[label style] (1/6,0) -- (1/6,-0.02) node[below] {$f^{\infty}$};
\draw[label style] (0.6, 0) -- (0.6, -0.02) node[below]{$f^{0}$};
\draw[label style] (0.36, 0) -- (0.36, -0.02) node[below]{$f^{1}$};
\draw[label style] (0.25, -0.07) node[below]{$\ldots$};
\draw[label style] (0, 0.1)--(-0.02, 0.1) node[left]{$\delta_{\max}$};
\draw[label style] (0, 1/6)--(-0.02, 1/6) node[left]{$f^{\infty}$};

  \draw (0,0) -- (-0.02,0) node[left] {};
  \def\fleft{\x, {0.212*sqrt(\x) + 0.08}}

  \def\fmiddle{\x, {(\x - 0.1667)/0.762)^3 + 0.1667}}

  \coordinate (A) at (5/6, 5/6);
  \coordinate (B) at (0.88, 0.93);
  \coordinate (C) at (0.93, 0.96);
  \coordinate (D) at (1, 1);
  \begin{scope}
    \path[fill=green!20]
      plot[domain=0:0.1667, samples=80] (\x, {0.212*sqrt(\x) + 0.08})
      -- plot[domain=0.1667:0, samples=80] (\x, \x)
      -- cycle;
  \end{scope}

  \begin{scope}
    \path[fill=red!20]
      plot[domain=0.8333:1, samples=80] (\x, {(0.40*sqrt(\x - 0.8333) + 0.8333})
      -- plot[domain=0.8333:1, samples=80] (\x, \x)
      -- cycle;
  \end{scope}

  \draw[blue, thick, domain=0:0.1667, samples=100, smooth]
       plot (\x, {0.212*sqrt(\x) + 0.08});

  \draw[blue, dashed, domain=0.1667:0.8333, samples=100, smooth]
       plot (\x, {((\x - 0.1667)/0.762)^3 + 0.1667});

  \draw[blue, thick, domain=0.8333:1, samples=100, smooth]
       plot (\x, {(0.40*sqrt(\x - 0.8333) + 0.8333});

  \draw[red, thick, fill=red]
       (1/6,1/6) circle[radius=0.5pt]
    (5/6,5/6) circle[radius=0.5pt]
    (1,1)   circle[radius=0.5pt];

    \draw[green, dashed] (0.60, 0) -- (0.60, 0.36);
    \draw[green, dashed] (0.36, 0) -- (0.36, 0.185);

    \draw[purple, dashed] (0.60, 0.36) -- (0.60, 0.60);
    \draw[purple, dashed] (0.60, 0.36) -- (0.36, 0.36);
    \draw[purple, dashed] (0.36, 0.185) -- (0.185, 0.185);
    \draw[purple, dashed] (0.36, 0.185) -- (0.36, 0.36);

        \draw[purple, ->, very thin, opacity=0.7] (0.60, 0.50) -- (0.60, 0.48); 
    \draw[purple, ->, very thin, opacity=0.7] (0.47, 0.36) -- (0.46, 0.36);
    \draw[purple, ->, very thin, opacity=0.7] (0.36, 0.27) -- (0.36, 0.26);
    \draw[purple, ->, very thin, opacity=0.7] (0.25, 0.185) -- (0.24, 0.185);

    \draw[gray, dashed] (0.1667, 0) -- (0.1667, 0.1667);
    \draw[gray, dashed] (0, 0.1667) -- (0.1667, 0.1667);
    \draw[gray, dashed] (0.1, 0) -- (0.1,0.1);
    \draw[gray, dashed] (0, 0.1) -- (0.1,0.1);

    \draw[gray, dashed] (1, 0) -- (1,1);
    \draw[gray, dashed] (0, 1) -- (1,1);

  \node at (0.1, 0.23) {\footnotesize$\mathcal{C}_{{\sf good}}$};
  \node at (0.85, 1.0) {\footnotesize$\mathcal{C}_{{\sf bad}}$};

\end{tikzpicture}
    }
    \caption{Main dynamics in bounding $\delta_{\max}$:\normalfont{ By using the structure of V-UCB, Lemma \ref{lemma:potential} shows that $\delta_{\max}$ is a postfixed point of $F$, meaning that it's either in the green area $\mathcal{C}_{\sf good}$ or the red area $\mathcal{C}_{\sf bad}$. To disqualify the red area, we use an initial upper bound $f_0$ obtained by classic tail-bounding arguments (Lemma \ref{lemma:initialpoint}). This improves the upper bound from $f^{\infty}$ to $\sup \mathcal{C}^{\sf good}$, which can be calculated as a limit of the dynamic system described by $F_T$ in Proposition \ref{prop:improvedrates}.}}
    \label{figure:analysis}
\end{figure}

We show in Appendix \ref{appendix:finalbounds} that $F_T$ is a non-decreasing smooth function that stabilizes the segment $[0,1]$ (see Figure \ref{figure:analysis}). Combining Lemmas \ref{lemma:initialpoint} and \ref{lemma:potential} gives the following bound on $\delta_{\max, T}$: 
\begin{equation*}
    \delta_{\max, T} \leq_{\text{(Lemma }\ref{lemma:potential})} F_T(\delta_{\max,T}) \leq_{(F_T \text{ non-decreasing + Lemma }\ref{lemma:initialpoint})} F_T(f^0_T).
\end{equation*}

By setting $f^1_T := F_T(f^0_T)$, we get $\delta_{\max, T} \leq F_T(f^0_T) = f^1_T$. Incidentally, using the new rate $f^1_T$ as a choice of $c$ in Proposition \ref{theorem:regretscheme} yields a regret bound that coincides with the results derived in \cite{activelearning, carpentier2011upper} (for the special case of sub-Gaussian distributions), where the authors  bound the tail of the number of samples from each group $\boldsymbol{n}$. In that sense, $f^1_T$ can also be found by the classic tail-bounding argument without going through the formalism of the relative distance $\delta_{\max,T}$. While this bound still achieves the optimal asymptotic rate in $T$, it has many flaws, such as a large numerical constant, and (most-importantly) a sub-optimal dependence in both $G$ and $\sigma_{\min}^{-1}$ (see comparisons in Section \ref{section:applications}).

Perhaps the most important technical contribution of our work is going beyond this rate by combining the relative-distance formalism with the potential function $F_T$: by iteratively re-injecting the most recent improved rate in the potential function $F_T$, we obtain a sequence of rates defined inductively as: 
$$f^{n+1}_T := F_T(f^n_T), \quad \forall n \geq 1,$$
and where each term $f^n_T$ satisfies $f^n_T \geq \delta_{\max, T}$. Using the curvature of the function $F_T$, we show in Lemma \ref{prop:improvedrates} that the sequence $\{f_T^n\}_{n \geq 1}$ is non-increasing and converges to a fixed point $f^\infty_T$ of $F_T$. That is,
$$f^0_T \geq f^1_T \geq \ldots \geq f^n_T \geq \ldots \geq \lim_{n \to +\infty} f^{n}_T = f^{\infty}_T = F_T(f^{\infty}_T).$$
By taking the limit in $n \to +\infty$, we obtain the \textit{optimized} convergence rate for $\delta_{\max,T}$: $$\delta_{\max,T} \leq f^{\infty}_T.$$
Conceptually, our technique can be viewed as re-injecting the tail-bounding technique argument infinitely many times, as visualized in Figure \ref{figure:analysis}. To properly formalize this idea, we introduce the set of \textit{good postfixed} points of $F_T$ :
\begin{equation*}
    \mathcal{C}^{\sf good}_T := \left\{x \in [0,1]: x \leq F_T(x),x \leq f^\infty_T\right\},
\end{equation*}
to rule out \textit{bad postfixed} points that are above the initial rate $f^0_T$ (visualized in red in Figure \ref{figure:analysis}).

\begin{restatable}[Improved convergence rate for relative distance]{lemma}{improvedrates}\label{prop:improvedrates}
    Define the following sequence
    \begin{equation*}
        \begin{cases}
            f_T^0 =  1 - \frac{\frac{1}{2}}{\left(1 + \max_{h \in [G]}{\sf w}_h\left(n^*_{h,T/2}\right)\right)^{\frac{2p}{p+1}}} \\
            f_T^{n+1} = F_T(f^n) &\forall n\geq 0.
        \end{cases}
    \end{equation*}
    The sequence $(f_T^n)_{n \geq 0}$ is non-increasing, and converges to a fixed point of $F_T$, denoted $f^\infty_T$. Moreover, conditioned on $\mathcal{A}$, it holds that $ \delta_{\max,T} \leq f^{\infty}_T = \sup\mathcal{C}_T^{\sf good} = \sup\left\{x \in [0,1]: x \leq F_T(x),x \leq f^\infty_T\right\}$.
\end{restatable}

Figure \ref{figure:analysis} illustrates the main dynamics yielding the bounding in Lemma \ref{prop:improvedrates}. The key benefit of Lemma \ref{prop:improvedrates} is that the new upper bound $\sup\mathcal{C}_{T}^{\sf good}$ is significantly smaller than the first rates $f^0_T, f^1_T$. 

In what follows, we bound $\sup\mathcal{C}_{T}^{\sf good}$ by the total uncertainty $\Bar{w}_p(\boldsymbol{\sigma},\boldsymbol{{\sf w}})$ and a lower bound on $\delta_{g,T}$. These are done by direct analysis of the fixed points of the function $F_T$.
\begin{restatable}[Bounding $\mathcal{C}_T^{\sf good}$]{lemma}{prefixed}\label{lemma:prefixed}
    $\sup \mathcal{C}_T^{\sf good} = (1 + o(1))\left(w_{1,p} + \frac{G}{T}\right)$.
\end{restatable}

To derive a lower bound on $\delta_{g,T}$, we use similar techniques, resulting in Lemma \ref{lemma:lowerbounddelta}.
\begin{restatable}[Lower bound on $\boldsymbol{\delta}_T$]{lemma}{lowerbounddelta}\label{lemma:lowerbounddelta}
    Conditioned on $\mathcal{A}$:
    \begin{equation*}
        \forall g \in [G], \qquad \delta_{g,t} \geq -\left[\frac{1}{n^*_{g,t}} + \left(\left(1 + {\sf w}_g(n^*_{g,t} - 1)\right)^{\frac{2p}{p+1}} - 1\right)\right].
    \end{equation*}
\end{restatable}

\subsection{Proof of Theorem \ref{theorem:upperboundinfinite}}\label{subsection:finalinfinite}

Here we give the full proof of Theorem \ref{theorem:upperboundinfinite}. The starting point is Proposition \ref{theorem:regretscheme}, which states that
\begin{equation*}
        \text{Regret}_{\infty, T}(\boldsymbol{\pi}, \boldsymbol{\mathcal{D}}|\mathcal{H}) \leq \inf_{c > 0} \left\{\mathbb{P}\left(\delta_{\max,T} > c \right) T + \frac{c}{(1 - c)^+} \right\}.
\end{equation*}
We choose $c = \sup \mathcal{C}_T^{\sf good}$. By Lemma \ref{prop:improvedrates}, conditioned on $\mathcal{A}$, the inequality $\delta_{\max, T} \leq c$ holds. This is equivalently restated as
\begin{equation*}
    \mathcal{A} \subset \{\delta_{\max, T} \leq c\}.
\end{equation*}
In particular,
\begin{equation}\label{ineq:highprobability}
    \mathbb{P}\left(\delta_{\max,T} > c \right) \leq \mathbb{P}(\mathcal{A}^{\sf c}) = \mathbb{P}\left(\left(\boldsymbol{0}_G \leq \boldsymbol{{\sf UCB}} -\boldsymbol{\sigma} \leq \boldsymbol{\sigma}{{\boldsymbol{\sf{w}}}}\right)^{\sf c}\right).
\end{equation}
Moreover,
\begin{equation*}
    c = \sup \mathcal{C}_T^{\sf good} = (1 + o(1))\left(w_{1,\infty} + \frac{G}{T}\right) = (1 + o(1))\left(\bar{w}_{\infty} + \frac{G}{T}\right),
\end{equation*}
where the first equality stems from the choice of $c$, the second equality stems from Lemma \ref{lemma:prefixed}, and the third equality stems from $\bar{w}_{\infty} = w_{1,\infty}$ (see Definition \ref{definition:uncertainty}). Finally, notice that both terms $\frac{G}{T}$ and $\bar{w}_{\infty}$ go to $0$ as $T \to +\infty$, hence $c \to_{T \to +\infty} 0$, and $(1 - c)^+ \to_{T \to +\infty} 1$, or equivalently $(1 - c)^+ = 1 - o(1)$.

Hence,
\begin{equation}\label{eq:cbounding}
    \frac{c}{(1 - c)^+} = \frac{1 + o(1)}{1 - o(1)}\cdot\left(\bar{w}_{\infty}(\boldsymbol{\sigma}, \boldsymbol{\mathcal{D}}) + \frac{G}{T}\right) = (1 + o(1))\left(\bar{w}_{\infty}(\boldsymbol{\sigma}, \boldsymbol{\mathcal{D}}) + \frac{G}{T}\right).
\end{equation}
Therefore,
\begin{align*}
    \text{Regret}_{\infty, T}(\boldsymbol{\pi}, \boldsymbol{\mathcal{D}}|\mathcal{H}) &\leq \mathbb{P}\left(\delta_{\max} > c \right) T + \frac{c}{(1 - c)^+} \\
    &\leq \mathbb{P}\left(\left(\boldsymbol{0}_G \leq \boldsymbol{{\sf UCB}} -\boldsymbol{\sigma} \leq \boldsymbol{\sigma}{{\boldsymbol{\sf{w}}}}\right)^{\sf c}\right)T + \frac{c}{(1 - c)^+} \\
    &= \mathbb{P}\left(\left(\boldsymbol{0}_G \leq \boldsymbol{{\sf UCB}} -\boldsymbol{\sigma} \leq \boldsymbol{\sigma}{{\boldsymbol{\sf{w}}}}\right)^{\sf c}\right)T + (1 + o(1))\left(\bar{w}_\infty(\boldsymbol{\sigma}, \boldsymbol{{\sf w}}) + \frac{G}{T}\right),
\end{align*}
where the first step stems from Proposition \ref{theorem:regretscheme}, the second step stems from Inequality \eqref{ineq:highprobability}, and the last step stems from Inequality \eqref{eq:cbounding}. This completes the proof of Theorem \ref{theorem:upperboundinfinite}.

\subsection{Proof of Theorem \ref{theorem:upperboundfinite}}\label{subsection:finalfinite}
We derive Theorem \ref{theorem:upperboundfinite} in a similar fashion, where the only difference is that $p<+\infty$ requires a double-sided bound on $\boldsymbol{\delta}_T$, instead of just an upper bound as in Theorem \ref{theorem:upperboundinfinite}. The starting point is Proposition \ref{theorem:regretschemefinite}, which states:
\begin{equation*}
        \text{Regret}_{p, T}(\boldsymbol{\pi}, \boldsymbol{\mathcal{D}}|\mathcal{H}) \leq \inf_{\boldsymbol{a} \leq \boldsymbol{0}_G \leq \boldsymbol{b}}\left\{\mathbb{P}\left((\boldsymbol{a} \leq \boldsymbol{\delta}_T \leq \boldsymbol{b})^{\sf c}\right) T + \frac{p+1}{2}\left(1 + \epsilon(\boldsymbol{a}, \boldsymbol{b})\right)\left(\sum_{g \in [G]}\lambda_g^*(a_g^2 + b_g^2)\right)\right\}.
\end{equation*}
We choose $\boldsymbol{b}$ (resp. $\boldsymbol{a}$)  to match the upper (resp. lower) bound obtained in Lemma \ref{prop:improvedrates} (resp. Lemma \ref{lemma:lowerbounddelta}). For each $g \in [G]$,
\begin{align*}
    a_g &:= -\left[\frac{1}{n^*_{g,t}} + \left(\left(1 + {\sf w}_h(n^*_{g,t} - 1)\right)^{\frac{2p}{p+1}} - 1\right)\right], \\
    b_g &:= (1 + o(1))\left(w_{1,p} + \frac{G}{T}\right).
\end{align*}
Then conditioned on $\mathcal{A}$, we have $\boldsymbol{a} \leq \boldsymbol{\delta} \leq \boldsymbol{b}$, which implies that
\begin{equation}\label{ineqconcentration2}
    \mathbb{P}\left((\boldsymbol{a} \leq \boldsymbol{\delta}_T \leq \boldsymbol{b})^{\sf c}\right) \leq \mathbb{P}(\mathcal{A}^{\sf c}) = \mathbb{P}\left(\left(\boldsymbol{0}_G \leq \boldsymbol{{\sf UCB}} -\boldsymbol{\sigma} \leq \boldsymbol{\sigma}{{\boldsymbol{\sf{w}}}}\right)^{\sf c}\right).
\end{equation}
Moreover, we show (in Lemma \ref{lemma:valueab}) that for $p < +\infty$,
\begin{equation}\label{equation:calculations}
   \sum_{g \in [G]}\lambda_g^* (a_g^2 + b_g^2) = (1 + o(1))\left(\bar{w}_p^2 + \frac{G}{T}\right), \qquad \text{and} \qquad \epsilon(\boldsymbol{a}, \boldsymbol{b}) = o(1).
\end{equation}
Therefore, 
\begin{align*}
    \text{Regret}_{p, T}(\boldsymbol{\pi}, \boldsymbol{\mathcal{D}}|\mathcal{H}) &\leq \mathbb{P}\left((\boldsymbol{a} \leq \boldsymbol{\delta}_T \leq \boldsymbol{b})^{\sf c}\right) T + \frac{p+1}{2}\left(1 + \epsilon(\boldsymbol{a}, \boldsymbol{b})\right)\left(\sum_{g \in [G]}\lambda_g^*(a_g^2 + b_g^2)\right) \\
    &\leq \mathbb{P}\left(\left(\boldsymbol{0}_G \leq \boldsymbol{{\sf UCB}} -\boldsymbol{\sigma} \leq \boldsymbol{\sigma}{{\boldsymbol{\sf{w}}}}\right)^{\sf c}\right)T + \frac{p+1}{2}(1 + o(1))\left(\bar{w}_p^2 + \frac{G}{T}\right),
\end{align*}
where the first step is from Proposition \ref{theorem:regretschemefinite}, and the second step stems from Equations \eqref{ineqconcentration2} and \eqref{equation:calculations}. This completes the proof of Theorem \ref{theorem:upperboundfinite}.

\section{Conclusion}\label{section:conclusion}
In this work, we introduced the Variance‐UCB algorithm and proved general regret bounds under both the infinite norm and finite $p$–norms. We instantiate these results for sub‐Gaussian, Gaussian, and exponential families of feedback. These instantiations significantly improve existing sample complexity guarantees and achieve new regret upper bounds, where none were previously known. Our results also achieve the first practical bounds for this problem in terms of sample complexity.

This paper introduces a unified framework for adaptive sampling under variance uncertainty. Central to our approach is the concept of \emph{admissible widths}, deterministic functions that capture confidence in sample‐based variance estimates, and the associated notion of \emph{decision error}, which quantifies how estimation uncertainty affects regret. By delegating uncertainty quantification to admissible widths, our framework supports arbitrary distributions, and focuses squarely on the intrinsic difficulty of learning variance parameters rather than on restrictive tail assumptions.

From an operational standpoint, our analysis provides a clear blueprint for sequential experimentation in A/B testing, clinical trials, and other data-dependent resource‐allocation settings. By treating the admissible width as a dynamic score of uncertainty, practitioners integrate active learning directly with estimation: treatment arms with higher estimated variance receive more samples, while those with lower variance are exploited. Because the framework imposes no sub‐Gaussian or bounded‐support requirements, it remains valid even under heavy‐tailed outcome distributions common in marketing lift studies, patient response times, and financial returns.

Future research can extend this toolkit to richer uncertainty regimes, incorporating covariate information, handling high‐dimensional estimators, or developing tools for alternate objectives.

\bibliographystyle{informs2014}

\bibliography{bib}
\newpage

\begin{APPENDICES}
\section{Proof of Theorems \ref{theorem:upperboundinfinite} and \ref{theorem:upperboundfinite}}\label{appendix:proofupperbound}
\subsection{Complete information setting}
\completeinformation*
\begin{proof}{\textit{Proof of Lemma \ref{lemma:completeinformation}.}}
 As a reminder, $R^*_p(\boldsymbol{\sigma})$ represents the optimal value in the complete information setting. The case of $p = +\infty$ is discussed in \cite{activelearning, carpentier2011upper}, and follows from the fact that $\max\left(\frac{\sigma_1^2}{n_1}, \ldots ,\frac{\sigma_G^2}{n_G}\right)$ is minimized if and only if its coordinates are equal, i.e., if and only if $\frac{\sigma_1^2}{n_1} = \ldots = \frac{\sigma_G^2}{n_G}$, which is equivalent to the desired expression.
 
We focus here on the case of $p < +\infty$. Recall the formal definition of $R^*_p(\boldsymbol{\sigma})$ from the optimization program stated in \eqref{linearprogram}:
    \begin{equation*}
R^*_{p}(\boldsymbol{\sigma}) = \quad \min_{\boldsymbol n \in \mathbb{R}_+^G} R_p(\boldsymbol n; \boldsymbol{\sigma}^2) \quad s.t. \quad \sum_{g \in [G]}n_g = T.
\end{equation*}
Since the function $x \mapsto x^p$ is increasing in $\mathbb{R}_+$, and the objective above satisfies $\left[R_p(\boldsymbol{n})\right]^p = \sum_{g \in [G]} \frac{\sigma_g^{2p}}{n_g^p}$, the program above has the same set of minimizers as the program
    \begin{equation*}
\min \sum_{g \in [G]} \frac{\sigma_g^{2p}}{n_g^p} \quad s.t. \quad \sum_{g \in [G]}n_g = T, \quad \boldsymbol{n} \geq \boldsymbol{0}_G.
\end{equation*}
 Any feasible point $\boldsymbol{n}$ with a zero coordinate yields an infinite objective in the program above, hence the coordinates of any minimizer to the optimization program must be strictly positive. Moreover, it must satisfy the KKT conditions:
    \begin{align*}
        \forall g \in [G],& \quad \frac{\partial}{\partial n_g}R_p(\boldsymbol{n})^p - \frac{\partial}{\partial n_g}\lambda\left(\sum_{h \in [G]} n_h - T\right) = 0, \\
        &\lambda \in \mathbb{R}.
    \end{align*}
    For each $g \in [G]$, the first line of the system above is equivalent to
    \begin{equation*}
        -p\frac{\sigma_g^{2p}}{n_g^{p+1}} - \lambda = 0.
    \end{equation*}
    Therefore the KKT conditions imply that:
    \begin{equation*}
        \frac{\sigma_1^{2p}}{n_1^{p+1}} = \cdots = \frac{\sigma_G^{2p}}{n_G^{p+1}}, \qquad \text{or equivalently,} \qquad \frac{\sigma_1^{\frac{2p}{p+1}}}{n_1} = \ldots = \frac{\sigma_G^{\frac{2p}{p+1}}}{n_G} = \frac{\sum_{h \in [G]} \sigma_h^{\frac{2p}{p+1}}}{\sum_{h \in [G]}n _h} = \frac{\sum_{h \in [G]} \sigma_h^{\frac{2p}{p+1}}}{T},
    \end{equation*}
    so that for each minimizer $\boldsymbol{n}$ must satisfy
    \begin{equation*}
       \forall g \in [G], \quad n_g = \frac{\sigma_g^{\frac{2p}{p+1}}T}{\sum_{h \in [G]} \sigma_h^{\frac{2p}{p+1}}}.
    \end{equation*}
    On the other hand, by continuity of $R_p$ over the compact feasible set $\{\boldsymbol{n} \geq \boldsymbol{0}_{G} | \sum_g n_g = T\}$, the set of minimizers is non-empty. Hence the optimization program defined in \eqref{linearprogram} indexed by $T$ has a unique minimizer, which we denote $\boldsymbol{n}^*_T$, and is the vector defined as: 
    \begin{equation*}
        \boldsymbol{n}^*_T = \frac{T}{\sum_{g \in [G]}\sigma_g^{\frac{2p}{p+1}}}\boldsymbol{\sigma}^{\frac{2p}{p+1}}.
    \end{equation*}
    Therefore, $R^*_p(\boldsymbol{\sigma}) = R_p(\boldsymbol{n}^*_T;\boldsymbol{\sigma})$. In particular, $\boldsymbol{n}^*$ is linear in $T$:
    \begin{equation*}
        \boldsymbol{n}^*_T = T \cdot \boldsymbol{n}^*_1 = T \cdot \boldsymbol{\lambda}^*_T.
    \end{equation*}
    This completes the proof of Lemma \ref{lemma:completeinformation}.
    $\hfill \square$
\end{proof}
\subsection{Proof of Propositions \ref{theorem:regretscheme} and \ref{theorem:regretschemefinite}}
\begin{lemma}\label{lemma:regret3}{\normalfont \textbf{[Regret characterization]}}
    For any policy $\boldsymbol{\pi}$ and instance $\boldsymbol{\mathcal{D}}$, we have
    \begin{equation*}
            \textnormal{Regret}_{p, T}(\boldsymbol{\pi}, \boldsymbol{\mathcal{D}}|\mathcal{H}) = \mathbb{E}_{\boldsymbol{\lambda}_T \sim (\boldsymbol{\pi}, \boldsymbol{\mathcal{D}})}\left[\left\|\frac{\boldsymbol{(\lambda}^*)^{\frac{1}{p}}}{1 - \boldsymbol{\delta}(\boldsymbol{\lambda}_T||\boldsymbol{\lambda}^*)}\right\|_p - 1\right].
    \end{equation*}
\end{lemma}
\begin{proof}{\textit{Proof of Lemma \ref{lemma:regret3}}}
Recall the definition of Regret (See Equation \eqref{equation:definitionregret})
\begin{equation*}
    \text{Regret}_{p,T}(\boldsymbol{\pi}, \boldsymbol{\mathcal{D}}) = \frac{\mathbb{E}_{\boldsymbol{n}_T \sim (\boldsymbol{\pi}, \boldsymbol{\mathcal{D}})}\left[R_p(\boldsymbol{n}_T;\boldsymbol{\sigma})\right] - R^*_p(\boldsymbol{\sigma})}{R^*_p(\boldsymbol{\sigma})}.
\end{equation*}
We shall write each of the terms $R_p(\boldsymbol{n}_T;\boldsymbol{\sigma})$ and $R^*_p(\boldsymbol{\sigma})$ in terms of $(\boldsymbol{\lambda}_T, \boldsymbol{\lambda}^*)$. On the one hand, following Lemma \ref{lemma:completeinformation}, we have
\begin{equation*}
    \left(\boldsymbol{\lambda}^*\right)^{1 + \frac{1}{p}} = \left(\frac{\boldsymbol{\sigma}^{\frac{2p}{p+1}}}{\Sigma_p(\boldsymbol{\sigma})}\right)^{\frac{p+1}{p}} = \frac{\boldsymbol{\sigma}^2}{\Sigma_p^{1 + \frac{1}{p}}(\boldsymbol{\sigma})}.
\end{equation*}
where we used Equation \eqref{eq:proportionalities}. By using the homogeneity of the norm, we have:
\begin{equation}\label{eq:rpvaluelambda}
    R_p(\boldsymbol{n}_T; \boldsymbol{\sigma})  = \left\|\frac{\boldsymbol{\sigma}^2}{\boldsymbol{n}_T}\right\|_p= \left\|\frac{\Sigma_p^{1 + \frac{1}{p}}(\boldsymbol{\sigma})\left(\boldsymbol{\lambda}^*\right)^{1 + \frac{1}{p}}}{T \boldsymbol{\lambda}_T}\right\|_p=\frac{\Sigma_p^{1 + \frac{1}{p}}(\boldsymbol{\sigma})}{T}\left\|\frac{\left(\boldsymbol{\lambda}^*\right)^{1 + \frac{1}{p}}}{\boldsymbol{\lambda}_T}\right\|_p.
\end{equation}
On the other hand, we have
\begin{equation}\label{eq:rpstarvalue}
    R_p^*(\boldsymbol{\sigma}) = R_p(\boldsymbol{n}^*_T;\boldsymbol{\sigma}) = \frac{\Sigma_p^{1 + \frac{1}{p}}(\boldsymbol{\sigma})}{T}\left\|\frac{\left(\boldsymbol{\lambda}^*\right)^{1 + \frac{1}{p}}}{\boldsymbol{\lambda}^*}\right\|_p = \frac{\Sigma_p^{1 + \frac{1}{p}}(\boldsymbol{\sigma})}{T}\left\|\left(\boldsymbol{\lambda}^*\right)^{\frac{1}{p}}\right\|_p = \frac{\Sigma_p^{1 + \frac{1}{p}}(\boldsymbol{\sigma})}{T},
\end{equation}
where the first equality follows from the optimality of $\boldsymbol{n}^*_T$ stated in Lemma \ref{lemma:completeinformation}, the second equality follows from Equation \eqref{eq:rpvaluelambda} applied to $\boldsymbol{n}^* = T \cdot \boldsymbol{\lambda}^*$, the third equation follows from simplifying the fraction, and the last equality stems from $\sum_{g \in [G]} \lambda_g^* = 1$.

Therefore, substituing each term in Equations \eqref{eq:rpvaluelambda} and \eqref{eq:rpstarvalue} in the regret's expression yields
\begin{align*}
    \text{Regret}_{p,T}(\boldsymbol{\pi}, \boldsymbol{\mathcal{D}}) &= \frac{\mathbb{E}_{\boldsymbol{n}_T \sim (\boldsymbol{\pi}, \boldsymbol{\mathcal{D}})}\left[R_p(\boldsymbol{n}_T;\boldsymbol{\sigma})\right] - R^*_p(\boldsymbol{\sigma})}{R^*_p(\boldsymbol{\sigma})}\\ &= \frac{\mathbb{E}_{\boldsymbol{\lambda}_T \sim (\boldsymbol{\pi}, \boldsymbol{\mathcal{D}})}\left[\frac{\Sigma_p^{1 + \frac{1}{p}}(\boldsymbol{\sigma})}{T}\left\|\frac{\left(\boldsymbol{\lambda}^*\right)^{1 + \frac{1}{p}}}{\boldsymbol{\lambda}_T}\right\|_p\right] - \frac{\Sigma_p^{1 + \frac{1}{p}}(\boldsymbol{\sigma})}{T}}{\frac{\Sigma_p^{1 + \frac{1}{p}}(\boldsymbol{\sigma})}{T}} \\
    \text{Regret}_{p,T}(\boldsymbol{\pi}, \boldsymbol{\mathcal{D}}) &= \mathbb{E}_{\boldsymbol{\lambda}_T \sim (\boldsymbol{\pi}, \boldsymbol{\mathcal{D}})}\left[\left\|\frac{\left(\boldsymbol{\lambda}^*\right)^{1 + \frac{1}{p}}}{\boldsymbol{\lambda}_T}\right\|_p\right] -1.
\end{align*}
And we derive the following intermediate regret expression:
\begin{equation}\label{regret:intermediate}
    \text{Regret}_{p, T}(\boldsymbol{\pi}, \boldsymbol{\mathcal{D}}) = \mathbb{E}_{\boldsymbol{\lambda}_T \sim (\boldsymbol{\pi}, \boldsymbol{\mathcal{D}})}\left[\left\|\frac{\boldsymbol{(\lambda}^*)^{1 + \frac{1}{p}}}{\boldsymbol{\lambda}_T}\right\|_p - 1\right].
\end{equation}

As a consequence of Lemma \ref{lemma:completeinformation}, the function 
\begin{equation*}
    \boldsymbol{\lambda} \in \Delta_G \mapsto \left\|\frac{\boldsymbol{(\lambda}^*)^{1 + \frac{1}{p}}}{\boldsymbol{\lambda}}\right\|_p - 1
\end{equation*}
is always non-negative, and is minimized uniquely at the point $\boldsymbol{\lambda} = \boldsymbol{\lambda}^*$ (where it equals $0$). Following the definition of the relative distance, we have 
\begin{equation*}
    \boldsymbol{\lambda} = \boldsymbol{\lambda}^*\left(1 - \boldsymbol{\delta}(\boldsymbol{\lambda}||\boldsymbol{\lambda}^*\right).
\end{equation*}
And we obtain:
\begin{align*}
        \text{Regret}_{p, T}(\boldsymbol{\pi}, \boldsymbol{\mathcal{D}}) &= \mathbb{E}_{\boldsymbol{\lambda}_T \sim (\boldsymbol{\pi}, \boldsymbol{\mathcal{D}})}\left[\left\|\frac{\boldsymbol{(\lambda}^*)^{1 + \frac{1}{p}}}{\boldsymbol{\lambda}_T}\right\|_p - 1\right] \\ &= \mathbb{E}_{\boldsymbol{\lambda}_T \sim (\boldsymbol{\pi}, \boldsymbol{\mathcal{D}})}\left[\left\|\frac{\boldsymbol{(\lambda}^*)^{1 + \frac{1}{p}}}{\boldsymbol{\lambda}^*\left(1 - \boldsymbol{\delta}(\boldsymbol{\lambda}_T||\boldsymbol{\lambda}^*)\right)}\right\|_p - 1\right] \\ &= \mathbb{E}_{\boldsymbol{\lambda}_T \sim (\boldsymbol{\pi}, \boldsymbol{\mathcal{D}})}\left[\left\|\frac{\boldsymbol{(\lambda}^*)^{\frac{1}{p}}}{1 - \boldsymbol{\delta}(\boldsymbol{\lambda}_T||\boldsymbol{\lambda}^*)}\right\|_p - 1\right],
\end{align*}
    where the first step stems from the expression of regret in Equation \eqref{regret:intermediate}, the second step stems from the definition of the relative distance $\boldsymbol{\delta}$ (Definition \ref{definition:distance}), and the third step stems from simplifying $\boldsymbol{\lambda}^*$. This concludes the proof of Lemma \ref{lemma:regret3}.
    $\hfill \square$
\end{proof}

\regretscheme*
\begin{proof}{\textit{Proof}}
    The inequality is straightforward for $c \geq 1$, as
    \begin{equation*}
        \frac{c}{(1 - c)^+} = +\infty.
    \end{equation*}
    We fix $c \in [0, 1)$. Following the definition of $\boldsymbol{\delta}$ (see Definition \ref{definition:distance}), we have for each pair $\boldsymbol{\lambda}^*, \boldsymbol{\lambda} \in \Delta_G$,
    \begin{equation*}
        \delta_g(\boldsymbol{\lambda} || \boldsymbol{\lambda}^*) = \lambda_g^*(1 - \lambda_g) \leq 1,
    \end{equation*}
    which implies that
    \begin{equation*}
        \delta_{\infty, t} = \max_{g \in [G]}\delta_{g,t} \leq 1.
    \end{equation*}
    Therefore,
    \small{
    \begin{align*}
        \text{Regret}_{\infty, T}(\boldsymbol{\pi}, \boldsymbol{\mathcal{D}}) &= \mathbb{E}_{\boldsymbol{\delta}_T \sim (\boldsymbol{\pi}, \boldsymbol{\mathcal{D}})}\left[\left\|\frac{1}{1 - \delta_{\infty, t}}\right\|_\infty - 1\right]\\ &=\mathbb{E}_{\boldsymbol{\delta}_T \sim (\boldsymbol{\pi}, \boldsymbol{\mathcal{D}})}\left[\frac{1}{1 - \delta_{\infty, t}} - 1\right] \\
        &=  \mathbb{P}\left(\delta_{\infty, t} > c\right)\mathbb{E}_{\boldsymbol{\delta}_T \sim (\boldsymbol{\pi}, \boldsymbol{\mathcal{D}})}\left[\frac{1}{1 - \delta_{\infty, t}} - 1 \bigg| \delta_{\infty, t} > c\right] + \mathbb{P}\left(\delta_{\infty, t} \leq c\right)\mathbb{E}_{\boldsymbol{\delta}_T \sim (\boldsymbol{\pi}, \boldsymbol{\mathcal{D}})}\left[\frac{1}{1 - \delta_{\infty, t}} - 1\bigg| \delta_{\infty, t} \leq c\right] \\
        &\leq \mathbb{P}\left(\delta_{\infty, t} > c\right)\times T + 1 \times \left(\frac{1}{1 - c} - 1\right) \\
        &= \mathbb{P}\left(\delta_{\max} > c \right) T + \frac{c}{1 - c},
    \end{align*}}
    \normalsize
    \noindent where the first step stems from Lemma \ref{lemma:regret3}, the second step follows from the inequality $\delta_{\infty, t}$, the third step stems from the law of total expectation, the last step stems from basic algebra, and the inequality in the fourth step stems from the following three inequalities:
    \small
    \begin{align*}
        &\mathbb{P}\left(\delta_{\infty, t} \leq c\right) \leq 1 \\
        &\mathbb{E}_{\boldsymbol{\delta}_T \sim (\boldsymbol{\pi}, \boldsymbol{\mathcal{D}})}\left[\frac{1}{1 - \delta_{\infty, t}} - 1 \bigg| \delta_{\infty, t} > c\right] \leq T\\
        &\mathbb{E}_{\boldsymbol{\delta}_T \sim (\boldsymbol{\pi},\boldsymbol{\mathcal{D}})}\left[\frac{1}{1 - \delta_{\infty, t}} - 1\bigg| \delta_{\infty, t} \leq c\right] \leq \frac{1}{1 - c} - 1,&\text{(The function }x \mapsto \frac{1}{1 - x} \text{ is increasing in }(-\infty;1])
    \end{align*}
    \normalsize
    \noindent This completes the proof of Proposition \ref{theorem:regretscheme}.
    $\hfill \square$

    \regretschemefinite*
    \begin{proof}{\textit{Proof.}}
    The starting point is the Taylor approximation around $\boldsymbol{0}_G$ of the function 
    \begin{equation*}
        \boldsymbol{\lambda} \in \Delta_G \mapsto \left\|\frac{(\boldsymbol{\lambda}^*)^{\frac{1}{p}}}{\boldsymbol{1} - \boldsymbol{\delta}(\boldsymbol{\lambda}||\boldsymbol{\lambda}^*)}\right\|_p.
    \end{equation*}
    derived in Proposition \ref{lemma:taylorfinitep}, which is
    \begin{equation*}
        \left\|\frac{\boldsymbol{(\lambda}^*)^{\frac{1}{p}}}{1 - \boldsymbol{\delta}}\right\|_p - 1 \leq \frac{p+1}{2}\sum_{g \in [G]} \lambda_g^* \delta_{g}^2 + \frac{p^2 \|\boldsymbol{\delta}\|^3_\infty}{(1 - \delta_{\max})^{2p + 3}}.
    \end{equation*}  
     Conditionally on $\boldsymbol{a} \leq \boldsymbol{\delta} \leq \boldsymbol{b}$, the right hand side in the inequality above can be bounded as
    \begin{equation*}
        \frac{p+1}{2}\sum_{g \in [G]} \lambda_g^* \delta_{g}^2 + \frac{p^2 \|\boldsymbol{\delta}\|^3_\infty}{(1 - \delta_{\max})^{2p + 3}}\leq \frac{p+1}{2}\sum_{g \in [G]} \lambda_g^*(a_g^2 + b_g^2) + \frac{p^2\max\left(\|\boldsymbol{a}\|^3_\infty,\|\boldsymbol{b}\|^3_\infty\right)}{((1 - b_{\max})^+)^{2p + 3}}.
    \end{equation*}
    \normalsize
    It remains to show the the rightmost term is negligible with respect to $\sum_{g \in [G]} \lambda_g^* (a_g^2 + b_g^2)$. Both the functions $\boldsymbol{x} \mapsto \sum_{g \in [G]} \lambda_g^* x_g^2$ and $\boldsymbol{x} \mapsto \|\boldsymbol{x}\|_\infty^2$ are the squares of norms (in the finite dimensional space $\mathbb{R}^G$), therefore by equivalence of norms, there exists a $c_{\boldsymbol{\lambda}} > 0$ (independent from the choice of $\boldsymbol{a}, \boldsymbol{b}$) such that
    \begin{equation*}
        \max\left(\|\boldsymbol{a}\|^2_\infty,\|\boldsymbol{b}\|^2_\infty\right) \leq c_{\boldsymbol{\lambda}}\sum_{g \in [G]}\lambda_g^* (a_g^2 + b_g^2),
    \end{equation*}
    so that
    \begin{equation*}
        \max\left(\|\boldsymbol{a}\|^3_\infty,\|\boldsymbol{b}\|^3_\infty\right) = \max(\|\boldsymbol{a}\|_\infty, \|\boldsymbol{b}\|_\infty) \cdot \max(\|\boldsymbol{a}\|^2_\infty, \|\boldsymbol{b}\|^2_\infty) \leq c_{\boldsymbol{\lambda}}\max(\|\boldsymbol{a}\|_\infty, \|\boldsymbol{b}\|_\infty)\sum_{g \in [G]}\lambda_g^*(a_g^2 + b_g^2).
    \end{equation*}
    Therefore, conditionally on $\boldsymbol{a} \leq \boldsymbol{0}_G \leq \boldsymbol{b}$, we have:
    \small
    \begin{align*}
        \frac{p+1}{2}\sum_{g \in [G]} \lambda_g^*(a_g^2 + b_g^2) + \frac{p^2\max\left(\|\boldsymbol{a}\|^3_\infty,\|\boldsymbol{b}\|^3_\infty\right)}{((1 - b_{\max})^+)^{2p + 3}} &= \left(\frac{p+1}{2}\sum_{g \in [G]}\lambda_g^*(a_g^2 + b_g^2)\right)\left(1 + \frac{\frac{p^2\max\left(\|\boldsymbol{a}\|^3_\infty,\|\boldsymbol{b}\|^3_\infty\right)}{((1 - b_{\max})^+)^{2p + 3}}}{\frac{p+1}{2}\sum_{g \in [G]}\lambda_g^*(a_g^2 + b_g^2)}\right)\\
        &\leq \left(\frac{p+1}{2}\sum_{g \in [G]}\lambda_g^*(a_g^2 + b_g^2)\right)\left(1 + \frac{2c_{\boldsymbol{\lambda}}p^2\max(\|\boldsymbol{a}\|_\infty, \|\boldsymbol{b}\|_\infty)}{(p+1)((1 - b_{\max})^+)^{2p+3}}\right)
    \end{align*}
    \normalsize
    so that by setting $\epsilon(\boldsymbol{a}, \boldsymbol{b}) := \frac{2c_{\boldsymbol{\lambda}}p^2\max(\|\boldsymbol{a}\|_\infty, \|\boldsymbol{b}\|_\infty)}{(p+1)((1 - b_{\max})^+)^{2p+3}}$, we have $\epsilon(\boldsymbol{a}, \boldsymbol{b}) \to_{(\boldsymbol{a},\boldsymbol{b}) \to \boldsymbol{0}} 0$ and we obtain conditionally on $\boldsymbol{a} \leq \boldsymbol{0}_G \leq \boldsymbol{b}$,
    \begin{equation}\label{ineq:intermediateregretscheme}
        \left\|\frac{\boldsymbol{(\lambda}^*)^{\frac{1}{p}}}{1 - \boldsymbol{\delta}}\right\|_p - 1  \leq \left(\frac{p+1}{2}\sum_{g \in [G]}\lambda_g^*(a_g^2 + b_g^2)\right)\left(1 + \epsilon(\boldsymbol{a}, \boldsymbol{b})\right).
    \end{equation}
    To complete the proof of Proposition \ref{theorem:regretschemefinite}, we apply -in a similar fashion to the proof of Proposition \ref{theorem:regretscheme}- the law of total expectation to yield
        \begin{align*}
            \text{Regret}_{p, T}(\boldsymbol{\pi}, \boldsymbol{\mathcal{D}}) &\leq \mathbb{P}\left((\boldsymbol{a} \leq \boldsymbol{0}_G \leq \boldsymbol{b})^{\sf c}\right)T + \mathbb{E}_{\boldsymbol{\lambda}_T \sim (\boldsymbol{\pi}, \boldsymbol{\mathcal{D}})}\left[\left\|\frac{\boldsymbol{(\lambda}^*)^{\frac{1}{p}}}{1 - \boldsymbol{\delta}(\boldsymbol{\lambda}_T||\boldsymbol{\lambda}^*)}\right\|_p - 1\bigg| \boldsymbol{a} \leq \boldsymbol{0}_G \leq \boldsymbol{b}\right] \\
            &\leq \mathbb{P}\left((\boldsymbol{a} \leq \boldsymbol{0}_G \leq \boldsymbol{b})^{\sf c}\right)T + \left(\frac{p+1}{2}\sum_{g \in [G]}\lambda_g^*(a_g^2 + b_g^2)\right)\left(1 + \epsilon(\boldsymbol{a}, \boldsymbol{b})\right),
        \end{align*}
        where the final step stems from Inequality \eqref{ineq:intermediateregretscheme}. This completes the proof.
    \end{proof}
    $\hfill \square$
\end{proof}

\subsection{Bounding the relative distance}
\initialrate*
\begin{proof}{\textit{Proof of Lemma \ref{lemma:initialpoint}}}
    Fix $g \in [G]$. Lemma \ref{lemma:structural} implies that
    \begin{equation}\label{ineqstruc1}
        \forall h \in [G], \qquad n_{g,t} \geq \frac{n_{h,t} - 1}{n^*_{h,t}(1+ {\sf w}_{h}(n_{h,t} - 1))^{\frac{2p}{p+1}}}\cdot n^*_{g,t}.
    \end{equation}
    Since $\sum_{h \in [G]} n_{h,t} = \sum_{h \in [G]} n^*_{h,t} = t$, there exists a group $h \in [G]$ that satisfies     
    \begin{equation*}
        n_{h,t} \geq n^*_{h,t}.
    \end{equation*}
    For such a group $h$, we necessarily have
    \begin{equation}\label{ineqstruc2}
        \frac{n_{h,t} - 1}{n^*_{h,t}(1+ {\sf w}_{h}(n_{h,t} - 1))^{\frac{2p}{p+1}}} \geq \frac{n^*_{h,t} - 1}{n^*_{h,t}(1+ {\sf w}_{h}(n^*_{h,t} - 1))^{\frac{2p}{p+1}}} \geq \frac{1/2}{\left(1 + {\sf w}_h\left(n^*_{h,t/2}\right)\right)^{\frac{2p}{p+1}}},
    \end{equation}
    where we used that ${\sf w}_h$ is non-increasing. We combine both inequalities \eqref{ineqstruc1} and \eqref{ineqstruc2}, while using that $\delta_{g,t} = 1 - \frac{n_{g,t}}{n^*_{g,t}}$ and taking the max over $g \in [G]$, and choosing $t = T$, to derive:
    \begin{equation*}
       \delta_{\max,T} = \max_{g \in [G]}\delta_{g,T} \leq 1 - \frac{1/2}{(1+ \max_{h \in [G]}{\sf w}_{h}(n^*_{h,T/2}))^{\frac{2p}{p+1}}} = f^0_T,
    \end{equation*}
    which completes the proof of Lemma \ref{lemma:initialpoint}.
    $\hfill \square$
\end{proof}
\potential*
\begin{proof}{\textit{Proof of Lemma \ref{lemma:potential}}}
    Taking Inequality \eqref{ineqstruc1} again, and summing it over $h \in [G]$ implies that for each group $g \in [G]$,
    \begin{equation*}
        n_{g,t} \sum_{h \in [G]}n^*_{h,t}(1 + {\sf w}_h(n_{h,t} - 1))^{\frac{2p}{p+1}} \geq \left(\sum_{h \in [G]}n_{h,t} - 1\right)n^*_{g,t},
    \end{equation*}
    or equivalently,
    \begin{equation*}
        \forall g \in [G], 1 - \delta_{g,t} \geq \frac{1 - \frac{G}{t}}{1 + \sum_{h \in [G]} \lambda_h^*\left[\left(1 + {\sf w}_h(n_{h,t} - 1)\right)^{\frac{2p}{p+1}} - 1\right]},
    \end{equation*}
    where we used that $\sum_{h \in [G]} n_{h,t} = \sum_{h \in [G]} n^*_{h,t} = t$, and that $\delta_{g,t} = 1 - \frac{n_{g,t}}{n^*_{g,t}}$ (see Definition \ref{definition:distance}). By taking the max on the left hand side, and by using that ${\sf w}_h$ is decreasing, we have:
    \begin{equation*}
        \delta_{\max, t} \leq 1 - \frac{1 - \frac{G}{t}}{1 + \sum_{h \in [G]} \lambda_h^*\left[\left(1 + {\sf w}_h(n^*_{h,t}(1 - \delta_{\max ,t}) - 1)\right)^{\frac{2p}{p+1}} - 1\right]} = F_T(\delta_{\max,t}),
    \end{equation*}
    which proves Lemma \ref{lemma:potential}.
    $\hfill \square$
\end{proof}
\improvedrates*
\begin{proof}{\textit{Proof.}}
We recall the following dynamic system
\begin{equation*}
    \begin{cases}
        f_T^0 = 1 - \frac{1/2}{(1+ \max_{h \in [G]}{\sf w}_{h}(n^*_{h,T/2}))^{\frac{2p}{p+1}}}\\
        f_T^{n+1} = F_T(f^n_T) &\forall n\geq 0.
    \end{cases}
\end{equation*}

Since $F_T$ is non-decreasing and continuous, the sequence $(f_T^n)_{n \geq 0}$ is monotonous, and takes its values in $[0,1]$, therefore it converges and its limit $f_T^\infty$ is well defined, is in $[0,1]$, and is a fixed point of $F_T$.

Consider the truncated set of postfixed points of $F_T$
\begin{equation*}
    \mathcal{C}^{\sf good}_{T} := \left\{x \in [0, 1] \bigg| \quad x \leq F_{T}(x), \quad \text{and}\quad x \leq f^{\infty}_T\right\}.
\end{equation*}
By definition of $\mathcal{C}^{\sf good}_T$, we have $f^\infty_T \geq \sup \mathcal{C}^{\sf good}_T$. On the other hand, $f^{\infty}_T$ is a fixed point of $F_T$, we have $f^T_\infty \leq F_T(f^T_\infty)$ and therefore $f^T_\infty \in \mathcal{C}^{\sf good}_T$. Therefore,
\begin{equation*}
    f^T_\infty = \sup \mathcal{C}^{\sf good}_{T}.
\end{equation*}
To complete the proof, it suffices to show that $\delta_{\max,T} \in \mathcal{C}^{\sf good}_T$. By Lemma \ref{lemma:potential}, we already have
\begin{equation*}
    \delta_{\max,T} \leq F_T(\delta_{\max,T}).
\end{equation*}
We show that $\delta_{\max,T} \leq f^{\infty,T}$ by induction on $n \geq 0$. The case $n = 0$ holds from Lemma \ref{lemma:initialpoint}. Assume that $\delta_{\max, T} \leq f^n_T$. Since $F_T$ is decreasing, we have:
\begin{equation*}
    F_T(\delta_{\max,T}) \leq F_T(f^n_T) = f^{n+1}_T,
\end{equation*}
so that by Lemma \ref{lemma:potential}, we have:
\begin{equation*}
     \delta_{\max,T} \leq F_T(\delta_{\max,T}) \leq f^{n+1}_T.
\end{equation*}
Therefore, by induction, we have:
\begin{equation*}
    \forall n \geq 0, \qquad \delta_{\max,T} \leq f^n_T.
\end{equation*}
Taking the limit in $n \to +\infty$ gives $\delta_{\max,T} \leq f^\infty_T$, which implies that $\delta_{\max,T} \in \mathcal{C}^{\sf good}_T$. This completes the proof of Proposition \ref{prop:improvedrates}.We recall the following dynamic system
\begin{equation*}
    \begin{cases}
        f_T^0 = 1 - \frac{1/2}{(1+ \max_{h \in [G]}{\sf w}_{h}(n^*_{h,T/2}))^{\frac{2p}{p+1}}}\\
        f_T^{n+1} = F_T(f^n_T) &\forall n\geq 0.
    \end{cases}
\end{equation*}

Since $F_T$ is non-decreasing and continuous, the sequence $(f_T^n)_{n \geq 0}$ is monotonous, and takes its values in $[0,1]$, therefore it converges and its limit $f_T^\infty$ is well defined, is in $[0,1]$, and is a fixed point of $F_T$.

Consider the truncated set of postfixed points of $F_T$
\begin{equation*}
    \mathcal{C}^{\sf good}_{T} := \left\{x \in [0, 1] \bigg| \quad x \leq F_{T}(x), \quad \text{and}\quad x \leq f^{\infty}_T\right\}.
\end{equation*}
By definition of $\mathcal{C}^{\sf good}_T$, we have $f^\infty_T \geq \sup \mathcal{C}^{\sf good}_T$. On the other hand, $f^{\infty}_T$ is a fixed point of $F_T$, we have $f^T_\infty \leq F_T(f^T_\infty)$ and therefore $f^T_\infty \in \mathcal{C}^{\sf good}_T$. Therefore,
\begin{equation*}
    f^T_\infty = \sup \mathcal{C}^{\sf good}_{T}.
\end{equation*}
To complete the proof, it suffices to show that $\delta_{\max,T} \in \mathcal{C}^{\sf good}_T$. By Lemma \ref{lemma:potential}, we already have
\begin{equation*}
    \delta_{\max,T} \leq F_T(\delta_{\max,T}).
\end{equation*}
We show that $\delta_{\max,T} \leq f^{\infty,T}$ by induction on $n \geq 0$. The case $n = 0$ holds from Lemma \ref{lemma:initialpoint}. Assume that $\delta_{\max, T} \leq f^n_T$. Since $F_T$ is decreasing, we have:
\begin{equation*}
    F_T(\delta_{\max,T}) \leq F_T(f^n_T) = f^{n+1}_T,
\end{equation*}
so that by Lemma \ref{lemma:potential}, we have:
\begin{equation*}
     \delta_{\max,T} \leq F_T(\delta_{\max,T}) \leq f^{n+1}_T.
\end{equation*}
Therefore, by induction, we have:
\begin{equation*}
    \forall n \geq 0, \qquad \delta_{\max,T} \leq f^n_T.
\end{equation*}
Taking the limit in $n \to +\infty$ gives $\delta_{\max,T} \leq f^\infty_T$, which implies that $\delta_{\max,T} \in \mathcal{C}^{\sf good}_T$. This completes the proof of Proposition \ref{prop:improvedrates}.
    $\hfill \square$
\end{proof}
\prefixed*
\begin{proof}{\textit{Proof.}}
    We introduce the slope
\begin{equation*}
    \epsilon_T := \frac{F_T(f^{\infty}_T) - F_T(0)}{f^{\infty}_T - 0}.
\end{equation*}
Since $f^\infty_T$ is a fixed point of $F_T$, we have by Proposition \ref{prop:improvedrates}:
\begin{equation*}
    \sup \mathcal{C}^{\sf good}_T = f_T^\infty = \frac{F_T(0)}{1 - \epsilon_T},
\end{equation*}
where the last equality stems from $F_T(0) = (1 + o(1))\left(w_{1,p} + \frac{G}{T}\right)$ (see Property $1$ of Lemma \ref{lemmaapp:propertiesF}).

It suffices to prove that $\epsilon_T = o(1)$. By the mean value theorem, there exists a $\bar{\delta} \in \left[0, \sup \mathcal{C}_T^{\sf good}\right]$ such that:
\begin{equation*}
    \epsilon_T = \frac{F_T(f^{\infty}_T) - F_T(0)}{f^{\infty}_T - 0}.
\end{equation*}
Since $\delta$ is uniformly bounded in $[0,1]$, by Property 3 of Lemma \ref{lemmaapp:propertiesF}, we have:
\begin{equation*}
     \epsilon_T = F_T'(\bar{\delta}) \to 0,
\end{equation*}
which shows that $\epsilon_T = o(1)$ and completes the proof of Lemma \ref{lemma:prefixed}.
$\hfill \square$
\end{proof}
\lowerbounddelta*
\begin{proof}{\textit{Proof.}}
    In the case where $n_{h,t} \leq n^*_{h,t}$, we must have
\begin{equation*}
    \delta_{h,t} = 1 - \frac{n_{h,t}}{n^*_{h,t}} \geq 0
\end{equation*}
and the desired inequality is immediate. We now assume that $n_{h,t} > n^*_{h,t}$, or equivalently, that $\delta_{h,t} < 0$. Conditionally on $\boldsymbol{0} \leq \boldsymbol{{\sf UCB}} - \boldsymbol{\sigma} \leq \boldsymbol{\sigma}\boldsymbol{{\sf w}}$, Inequality \eqref{ineqstruc2} holds we have for each pair of groups $g, h \in [G]$:
\begin{equation*}
     \frac{n_{g,t}}{n^*_{g,t}} \geq \frac{n_{h,t} - 1}{n^*_{h,t}(1+ {\sf w}_{h}(n_{h,t} - 1))^{\frac{2p}{p+1}}}.
\end{equation*}
Taking the minimum over $g \in [G]$, and using that $\min_{g \in [G]} \frac{n_{g,t}}{n^*_{g,t}} \leq 1$, the previous inequality implies that for each $h \in [G]$:
\begin{equation*}
    1 \geq \frac{n_{h,t} - 1}{n^*_{h,t}(1+ {\sf w}_{h}(n_{h,t} - 1))^{\frac{2p}{p+1}}},
\end{equation*}
which (by using $n_{h,t} = n^*_{h,t}(1 - \delta_{h,t})$), can be rewritten as
\begin{equation*}
    \forall h \in [G], \qquad \frac{(1 - \delta_{h,t}) - \frac{1}{n^*_{h,t}}}{(1+ {\sf w}_{h}(n^*_{h,t}(1 - \delta_{h,t}) - 1))^{\frac{2p}{p+1}}} \leq 1,
\end{equation*}
or equivalently,
\begin{equation*}
    \delta_{h,t} \geq -\left[\frac{1}{n^*_{h,t}} + \left(\left(1 + {\sf w}_h(n^*_{h,t}(1 - \delta_{h,t}) - 1)\right)^{\frac{2p}{p+1}} - 1\right)\right].
\end{equation*}
Since $\delta_{h,t} < 0$ (by assumption) and ${\sf w}_h$ is decreasing (by definition of admissible width), we must have:
\begin{equation*}
    {\sf w}_h(n^*_{h,t}(1 - \delta_{h,t}) - 1) \leq {\sf w}_h(n^*_{h,t} - 1),
\end{equation*}
hence
\begin{equation*}
    \delta_{h,t} \geq -\left[\frac{1}{n^*_{h,t}} + \left(\left(1 + {\sf w}_h(n^*_{h,t} - 1)\right)^{\frac{2p}{p+1}} - 1\right)\right],
\end{equation*}
which yields the desired inequality for the case where $\delta_{h,t} < 0$. This completes the proof of Proposition \ref{lemma:lowerbounddelta}.
$\hfill \square$
\end{proof}
\section{Final regret bounds}\label{appendix:finalbounds}
\subsection{Useful concentration bounds}
For completeness, we state some useful concentration bounds. The proof of Lemma \ref{lemma:subgaussian} is stated in  \cite{maurer2009empirical}, and the proofs of both Lemmas \ref{lemma:gaussian} and \ref{lemma:exponential} are both stated in Chapter 2 of the textbook \cite{10.1093/acprof:oso/9780199535255.001.0001}.

\begin{lemma}[Sample variance for sub-Gaussian distributions]\label{lemma:subgaussian}
    Let $\boldsymbol{\mathcal{D}}$ be $G \geq 1$ sub-Gaussian distributions with standard deviation vector $\boldsymbol{\sigma}$. Then there exists $\hat{\boldsymbol{c}} \geq \boldsymbol{\sigma}$ such that the pair $(\boldsymbol{\mathcal{D}}, \hat{\boldsymbol{c}})$ satisfies Assumption \ref{assumption:subgaussian}.
\end{lemma}

\begin{lemma}[Sample variance for Gaussian distributions]\label{lemma:gaussian}
    Let $\mathcal{D}$ be a Gaussian distribution with variance $\sigma^2 > 0$, and let $\hat{\sigma}^2$ denote the unbiased sample variance calculated from using $n$ i.i.d samples from $\mathcal{D}$, where $n \geq 2$ is possibly random but is almost surely bounded. We have for all $\epsilon \in (0, 1)$
    \begin{equation*}
        \mathbb{P}\left(\left|\frac{\hat{\sigma}^2}{\sigma^2}-1\right| \leq \sqrt{\frac{2\log \frac{1}{\epsilon}}{n}} + \frac{2\log \frac{1}{\epsilon}}{n}\right) \geq 1 - \epsilon
    \end{equation*}
\end{lemma}
\begin{lemma}[Sample mean for exponential distributions]\label{lemma:exponential}
    Let $\mathcal{D}$ be an exponential distribution with mean $\mu$ and standard deviation $\sigma$, and let $\hat{\mu}$ be the sample mean calculated from $n \geq 2$ i.i.d. samples from $\mathcal{D}$, where $n$ is almost surely bounded. Then $\mu = \sigma$ and for all $\epsilon \in (0,1)$.
    \begin{equation*}
        \mathbb{P}\left(\left|\frac{\hat{\mu}}{\sigma} - 1\right| \leq \sqrt{\frac{2\log \frac{1}{\epsilon}}{n}}\right)\geq 1 - \epsilon.
    \end{equation*}
\end{lemma}

\subsection{Regret bounds for Sub-Gaussian distributions}\label{appendix:finalsubgaussian}
\subgaussian*
\begin{proof}{\textit{Proof.}}
    The starting point are Theorems \ref{theorem:upperboundinfinite} and \ref{theorem:upperboundfinite}, where we have for any admissible width $\boldsymbol{\sf w}$ and $p \in [1, +\infty]$:
    \begin{equation*}
        \textnormal{Regret}_{p,T}(\textnormal{V-UCB}, \boldsymbol{\mathcal{D}}|\mathcal{H}) \leq \mathbb{P}\left[\left(\boldsymbol{0} \leq \boldsymbol{{\sf UCB}} - \boldsymbol{\sigma} \leq \boldsymbol{\sigma}\boldsymbol{{\sf w}}\right)^{\sf c}\right] T + (1 + o(1)) \times \begin{cases}
            \bar{w}_p(\boldsymbol{\sigma}, \boldsymbol{{\sf w}}) + \frac{G}{T} &\text{, if $p = +\infty$} \\
            \frac{p+1}{2}\bar{w}_p^2(\boldsymbol{\sigma}, \boldsymbol{{\sf w}})&\text{, if $p < +\infty$}
        \end{cases}
    \end{equation*}
    We choose ${\sf w}_g \equiv {\sf w}^{\sf sub-G}_{g}(n) := \frac{2\hat{c}_g}{\sigma_g}\sqrt{\frac{3\log T}{n}}$. In what follows, we bound both of
    \begin{equation*}
        \mathbb{P}\left[\left(\boldsymbol{0} \leq \boldsymbol{{\sf UCB}} - \boldsymbol{\sigma} \leq \boldsymbol{\sigma}\boldsymbol{{\sf w}}\right)^{\sf c}\right], \text{ and } \bar{w}_p(\boldsymbol{\sigma}, \boldsymbol{{\sf w}}).
    \end{equation*}
    \textbf{Bounding $\mathbb{P}\left[\left(\boldsymbol{0} \leq \boldsymbol{{\sf UCB}} - \boldsymbol{\sigma} \leq \boldsymbol{\sigma}\boldsymbol{{\sf w}}\right)^{\sf c}\right]$.} It is easy to see that $\boldsymbol{{\sf w}}^{\sf sub-G}$ is an admissible width (see Definition \ref{definition:width}). We set
    \begin{equation*}
        \mathcal{A} := \bigcap_{g, t}\left\{\left|\hat{\sigma}_{g,t} - \sigma_g\right| \leq \hat{c}_g\sqrt{\frac{\log T^3}{n_{g,t}}} = \hat{c}_g\sqrt{\frac{3\log T}{n_{g,t}}}\right\}. 
    \end{equation*}
    Following Assumption \ref{assumption:subgaussian} and Lemma \ref{lemma:subgaussian}, we have $\mathbb{P}(\mathcal{A}) \geq 1 - GT\cdot\frac{1}{T^3} = 1 - \frac{G}{T^2}$. Moreover, conditionally on $\mathcal{A}$, we have for each pair $g, t$
    \begin{align*}
    \textsf{UCB}^{\sf sub-G}_{g,t} - \sigma_g &= \hat{\sigma}_{g,t} + \hat{c}_g\sqrt{\frac{3\log T}{n_{g,t}}} - \sigma_g \\
    &= \hat{c}_g \sqrt{\frac{3\log T}{n_{g,t}}} - (\sigma_g - \hat{\sigma}_{g,t}) \\
    &\geq 0 \\
        \sigma_g{\sf w}^{\sf G}_{g}(n_{g,t}) - \left(\textsf{UCB}^{\sf sub-G}_{g,t} - \sigma_g\right) &= 2\hat{c}_g \sqrt{\frac{3\log T}{n_{g,t}}} - \left(\hat{\sigma}_{g,t} + \hat{c}_g\sqrt{\frac{3\log T}{n_{g,t}}} - \sigma_g\right) \\
        &= (\sigma_g - \hat{\sigma}_{g,t}) - \hat{c}_g \sqrt{\frac{3\log T}{n_{g,t}}} \\
        &\geq 0,
    \end{align*}
    so that $\mathcal{A} \subset  \left\{\boldsymbol{0} \leq \boldsymbol{{\sf UCB}} - \boldsymbol{\sigma} \leq \boldsymbol{\sigma}{\boldsymbol{\sf w}}\right\}$, which implies that
    \begin{equation}\label{eq:probabilitybounding}
        \mathbb{P}\left(\left(\boldsymbol{0} \leq \boldsymbol{{\sf UCB}} - \boldsymbol{\sigma} \leq \boldsymbol{\sigma}{\boldsymbol{\sf w}}\right)^{\sf c}\right)T \leq \mathbb{P}(\mathcal{A}^{\sf c})T \leq \left(1 - \left(1 - \frac{G}{T^2}\right)\right)T = \frac{G}{T}.
    \end{equation}
    \textbf{Calculating $\bar{w}_p(\boldsymbol{\sigma}, {\boldsymbol{\sf w}})$.} For simplicity, we drop the dependence in $(\boldsymbol{\sigma}, {\boldsymbol{\sf w}})$. As a reminder (see Definition \ref{definition:uncertainty}), we have:
    \begin{equation*}
        \bar{w}_p = \sqrt{w_{1,p}^2 + \mathbbm{1}(p < +\infty) w_{2,p}^2}.
    \end{equation*}
    On the one hand, we have:
    \begin{align*}
        w_{1,p} &= \sum_{g \in [G]} \frac{n^*_{g,T}}{T}\left[\left(1 + {\sf w}_g(n^*_{g,T} - 1)\right)^{\frac{2p}{p+1}}- 1\right] \\
        &= \sum_{g \in [G]} \frac{n^*_{g,T}}{T}\left[\left(1 + 2\frac{\hat{c}_g}{\sigma_g}\sqrt{\frac{3\log T}{n^*_{g,T} - 1}}\right)^{\frac{2p}{p+1}}- 1\right] \\
        &= 2\sqrt{3} \cdot \frac{2p}{p+1}(1 + o(1))\sum_{g \in [G]}\frac{\hat{c}_g}{\sigma_g}\sqrt{\frac{\log T}{n^*_{g,T}}}\cdot \frac{n^*_{g,T}}{T} \\
        &= \frac{p}{p+1}\cdot4\sqrt{3}(1 + o(1))\left( \sum_{g \in [G]}\frac{\hat{c}_g}{\sigma_g} \sqrt{\frac{\sigma_g^{\frac{2p}{p+1}}}{\Sigma_p(\boldsymbol{\sigma})}}\right)\sqrt{\frac{\log T}{T}} \\
        &= \frac{4\sqrt{3}p}{p+1}(1 + o(1))\left(\frac{\sum_{g \in [G]}\hat{c}_g\sigma_g^{\frac{-1}{p+1}}}{\sqrt{\Sigma_p(\boldsymbol{\sigma})}}\right)\sqrt{\frac{\log T}{T}}
    \end{align*}
    where the first step stems from the definition of $w_{1,p}$, the second step stems from the choise of ${\sf w}_g$, the third step stems from the first order approximation of $x \mapsto \left(1 + \frac{\kappa}{sqrt{x - 1}}\right)^{\frac{2p}{p+1}} - 1 \approx \frac{2p}{p+1}\frac{\kappa}{\sqrt{x}}$ in $x \to +\infty$, and the last step stems from Lemma \ref{lemma:completeinformation}, which states that $n^*_{g,T} = \frac{\sigma_g^{\frac{2p}{p+1}}}{\Sigma_p(\boldsymbol{\sigma})}T$. Similarly,
    \begin{align*}
        w_{2,p}^2 &= \sum_{g \in [G]} \frac{n^*_{g,T}}{T}\left[\left(1 + {\sf w}^2_g(n^*_{g,T} - 1)\right)^{\frac{2p}{p+1}}- 1\right] \\
        &= \sum_{g \in [G]} \frac{n^*_{g,T}}{T}\left[\left(1 + \frac{\hat{c}_g^2}{\sigma_g^2}\frac{3\log T}{n^*_{g,t}- 1}\right)^{\frac{2p}{p+1}}- 1\right] \\
        &= \frac{2p}{p+1}(1 + o(1))\sum_{g \in [G]} \frac{n^*_{g,T}}{T} \cdot \frac{\hat{c}_g^2}{\sigma_g^2}\cdot \frac{3\log T}{n^*_{g,T}} \\
        &= \frac{6p}{p+1}(1 + o(1))\left(\sum_{g \in [G]}\frac{\hat{c}_g^2}{\sigma_g^2}\right)\frac{\log T}{T}
    \end{align*}

    \textbf{Deriving the final bound.} We are now ready to conclude the proof. 
    \begin{enumerate}
        \item For $p = +\infty$, we have $\bar{w}_\infty = w_{1, \infty}$, and:
    \begin{align*}
            \textnormal{Regret}_{\infty,T}(\textnormal{V-UCB}, \boldsymbol{\mathcal{D}}|\mathcal{H}) &\leq \mathbb{P}\left(\left(\boldsymbol{0} \leq \boldsymbol{{\sf UCB}} - \boldsymbol{\sigma} \leq \boldsymbol{\sigma}{\boldsymbol{\sf w}}\right)^{\sf c}\right)T + (1 + o(1))\left(\bar{w}_\infty(\boldsymbol{\sigma}, \boldsymbol{{\sf w}}) + \frac{G}{T}\right) \\
            &\leq \frac{G}{T} + 4\sqrt{3}(1 + o(1))^2\left(\frac{\sum_{g \in [G]}\hat{c}_g\sigma_g^{\frac{-1}{+\infty+1}}}{\sqrt{\Sigma_\infty(\boldsymbol{\sigma})}}\right)\sqrt{\frac{\log T}{T}} + (1 + o(1))\frac{G}{T}\\
            &= 4\sqrt{3}(1 + o(1))^2\left(\frac{\|\hat{\boldsymbol{c}}\|_1}{\|\boldsymbol{\sigma}\|_2}\right)\sqrt{\frac{\log T}{T}}+ \left(2 + o(1)\right)\frac{G}{T} \\
            &= 4\sqrt{3}(1 + o(1))\frac{\|\hat{\boldsymbol{c}}\|_2}{\|\boldsymbol{\sigma}\|_2}\sqrt{\frac{G \log T}{T}},
    \end{align*}
    where the first step stems from Theorem \ref{theorem:upperboundinfinite}, the second step stems from Inequality \eqref{eq:probabilitybounding} and the expression of $w_{1, p}$ at $p = +\infty$, the third step stems from $x^0 = 1$ and $\sum_{g \in [G]} \hat{c}_g = \|\hat{\boldsymbol{c}}\|_1$, and the last step stems from $\|\hat{\boldsymbol{c}}\|_1 \leq \sqrt{G}\|\hat{\boldsymbol{c}}\|_2$ and $\frac{G}{T} = o\left(\sqrt{\frac{G \log T}{T}}\right)$.    
    \item For $p < +\infty$, we have:
    \begin{align*}
       \textnormal{Regret}_{p,T}(\textnormal{V-UCB}, \boldsymbol{\mathcal{D}}|\mathcal{H}) &\leq \mathbb{P}\left(\left(\boldsymbol{0} \leq \boldsymbol{{\sf UCB}} - \boldsymbol{\sigma} \leq \boldsymbol{\sigma}{\boldsymbol{\sf w}}\right)^{\sf c}\right)T + (1 + o(1))\frac{p+1}{2}\bar{w}^2_p(\boldsymbol{\sigma}, \boldsymbol{{\sf w}}) \\
       &\leq \frac{G}{T} + (1 + o(1))\frac{p+1}{2}\left(w_{1,p}^2 + w_{2,p}^2\right) \\
       &= \frac{G}{T} + \frac{p+1}{2}\frac{p}{p+1}(1 + o(1))^3\left(48\left(\frac{\sum_{g \in [G]}\hat{c}_g\sigma_g^{\frac{-1}{p+1}}}{\sqrt{\Sigma_p(\boldsymbol{\sigma})}}\right)^2 + 36\left(\sum_{g \in [G]}\frac{\hat{c}_g^2}{\sigma_g^2}\right)\right)\frac{\log T}{T} \\
       &= \frac{G}{T} + 12p(1 + o(1))\left(\frac{4\left(\sum_{g \in [G]}\hat{c}_g\sigma_g^{\frac{-1}{p+1}}\right)^2}{\Sigma_p(\boldsymbol{\sigma})} + 3\left(\sum_{g \in [G]}\frac{\hat{c}_g^2}{\sigma_g^2}\right)\right)\frac{\log T}{T},
    \end{align*}
    where the first step stems from Theorem \ref{theorem:upperboundfinite}, the second step stems from Inequality \eqref{ineq:highprobability} and the calculations of $w_{1,p}$ and $w_{2,p}$ as well as the expression of $\bar{w}_p$ stated in Definition \eqref{definition:uncertainty}, and the third step stems from regrouping the terms and using that $(1 + o(1))^3 = 1 + o(1)$. We simplify the term in $\hat{\boldsymbol{c}}, \boldsymbol{\sigma}$ using Cauchy's Schwartz inequality, we have:
    \begin{align*}
        \left(\sum_{g \in [G]}\hat{c}_g\sigma_g^{\frac{-1}{p+1}}\right)^2 = \left(\sum_{g \in [G]} \frac{\hat{c}_g}{\sigma_g} \cdot \sigma_g^{\frac{p}{p+1}}\right)^2 \leq \left(\sum_{g \in [G]}\frac{\hat{c}_g^2}{\sigma_g^2}\right)\Sigma_p(\boldsymbol{\sigma}),
    \end{align*}
    where we used that $\Sigma_p(\boldsymbol{\sigma}) = \sum_{g \in [G]} \sigma_g^{\frac{2p}{p+1}}$ (See Lemma \ref{lemma:completeinformation}). Therefore, the right hand side above can be simplified to
    \begin{equation*}
        \textnormal{Regret}_{p,T}(\textnormal{V-UCB}, \boldsymbol{\mathcal{D}}|\mathcal{H}) \leq 85p(1 + o(1))\left(\sum_{g \in [G]} \frac{\hat{c}_g^2}{\sigma_g^2}\right)\frac{\log T}{T}.
    \end{equation*}
    where we used $\frac{G}{T} \leq p(1+o(1))\frac{G\log T}{T}$.
    \end{enumerate}
    This completes the proof of Theorem \ref{theorem:subgaussian}.
    $\hfill \square$
\end{proof}
\subsection{Gaussian distributions}\label{appendix:finalgaussian}
\gaussian*
\begin{proof}{\textit{Proof.}}
    we choose ${\sf w}_g \equiv {\sf w}_g^{\sf G}(n) := \sqrt{\frac{3\log T}{n}} + \frac{3\log T}{n}$. It is easy to see that $\boldsymbol{{\sf w}}$ is an admissible width (see Definition \ref{definition:width}). Moreover, following Lemma \ref{lemma:gaussian}, and by union bound over $[G] \times [T]$, by setting $\epsilon = \frac{1}{T^3}$, we have with probability at least $1 - \frac{G}{T^2}$:
    \begin{align*}
        \textsf{UCB}^{\sf G}_{g,t} - \sigma_g = \sigma_g \underbrace{\left(\frac{\hat{\sigma}_{g,t}}{\sigma_g} -1 + \sqrt{\frac{3\log T}{n_{g,t}}} + \frac{3\log T}{n_{g,t}}\right)}_{\in \left[0, \textsf{w}^{\sf G}(n_{g,t})\right]} \in \left[0, {\sf w}_g^{\sf G} \sigma_g\right].
    \end{align*}
    Moreover, by following the same line of calculations as in the proof of Theorem \ref{theorem:gaussianfeedback} (by "replacing" $\hat{c}_g$ with $\sigma_g/2$ and using $\frac{\log T}{n} = o\left(\sqrt{\frac{\log T}{n}}\right)$), we have:
    \begin{align*}
        w_{1,p} &= \frac{2\sqrt{3}p}{p+1}(1 + o(1)) \cdot \frac{\sum_{g \in [G]} \sigma_g^{\frac{p}{p+1}}}{\sqrt{\sum_{g \in [G]} \sigma_g^{\frac{2p}{p+1}}}}\sqrt{\frac{\log T}{T}} \\
        w_{2,p}^2 &= \frac{3p}{p+1}(1 + o(1)) \frac{G \log T}{T}.
    \end{align*}
    Therefore, once again by following the same steps as in the proof for Theorem \ref{theorem:subgaussian}, we have:
    \begin{enumerate}
        \item For $p = +\infty$,
        \begin{equation*}
            \textnormal{Regret}_{\infty,T}(\textnormal{V-UCB}, \boldsymbol{\mathcal{D}}|\mathcal{H}) \leq 2\sqrt{3}(1 + o(1))\frac{\|\boldsymbol{\sigma}\|_1}{\|\boldsymbol{\sigma}\|_2} \sqrt{\frac{\log T}{T}}.
        \end{equation*}
        \item For $p < +\infty$,
        \begin{equation*}
            \textnormal{Regret}_{p,T}(\textnormal{V-UCB}, \boldsymbol{\mathcal{D}}|\mathcal{H}) \leq 43p(1 + o(1))\frac{G \log T}{T},
        \end{equation*}
    \end{enumerate}
    which completes the proof for Theorem \ref{theorem:gaussianfeedback}.
    $\hfill \square$
\end{proof}
\subsection{Exponential distributions}
\exponential*
\begin{proof}{\textit{Proof.}}
    The proof follows the exact same steps as the proof of Theorem \ref{theorem:gaussianfeedback}, except for this time, we set ${\sf w}^{\sf exp}(n) := \sqrt{\frac{2\log \frac{1}{\epsilon}}{n}}$ and we use the concentration bound derived in Lemma \ref{lemma:exponential}.
    $\hfill \square$
\end{proof}

\section{Auxiliary lemmas}
\subsection{Structural lemmas for the upper bound}
Through this section,  we assume that $\boldsymbol{0} \leq \boldsymbol{{\sf UCB}} - \boldsymbol{\sigma} \leq \boldsymbol{\sigma}\boldsymbol{{\sf w}}$. We introduce the (parametric) real-valued function:
\begin{equation*}
       F_T: u \in [0, 1]  \mapsto 1 - \left(1 - \frac{G}{t}\right)\left[\sum_{h \in [G]}\lambda_h^*\left(1 + {\sf w}_{h}(n^*_{h,t}(1 - u) - 1)\right)^{\frac{2p}{p+1}}\right]^{-1}.
\end{equation*}

\begin{lemma}[Analytical properties of $F_T$]\label{lemmaapp:propertiesF}
    The following two properties hold:
    \begin{enumerate}
        \item {\normalfont \textbf{[Regularity]}} The function $F_T$ is differentiable with a continuous derivative, non-decreasing, and satisfies $F_T([0, 1]) = [F_T(0), 1] \subset (0, 1]$. Moreover, 
        \begin{equation*}
            \frac{w_{1,p} + \frac{G}{T}}{1 + w_{1,p}} \leq F_T(0) = w_{1,p} + \frac{G}{t} \leq \frac{w_{1,p} + \frac{G}{T}}{(1 - w_{1,p})^+}.
        \end{equation*}
        In particular,
        \begin{equation*}
            \frac{F_T(0)}{w_{1,p} + \frac{G}{T}} \to 1.
        \end{equation*}
        \item {\normalfont \textbf{[Simple convergence to $0$]}} For each $c \in (0, 1)$, we have:
        \begin{equation*}
        F_T(c) \to 0.
        \end{equation*}
        
        \item {\normalfont \textbf{[Smoothness in $0^+$]}} For any doubly indexed, non-negative sequence $(x_{t})$ satisfying $x_{t} \to 0$, we also have
        \begin{equation*}
            F_T(x_{t}) \to 0 \qquad\text{and}\qquad F'_{t}(x_{t}) \to 0.
        \end{equation*}
    \end{enumerate}
\end{lemma}
\begin{proof}{\textit{Proof}.} 
We prove each of the two properties above
    \begin{enumerate}
        \item \textbf{[Regularity]}

        \textbf{Showing differentiability.} The differentiability is immediate as it follows from the differentiability of each ${\sf w}_h$ (See Definition \ref{definition:width}).
        
        \textbf{Showing monotonicity.} Recall that each ${\sf w}_h$ is non-increasing, hence each $u \mapsto {\sf w}_h(n^*_{h,t}(1 - u) - 1)$ is non-decreasing. Consequently, the weighted sum
        \begin{equation*}
            \sum_{h \in [G]}\lambda_h^*\left(1 + {\sf w}_{h}(n^*_{h,t}(1 - u) - 1)\right)^{\frac{2p}{p+1}}
        \end{equation*}
is also non-decreasing. On the other hand, the function $x \mapsto 1 - \frac{1 - \frac{G}{t}}{x}$ is also increasing. Hence, $F_T$ is non-decreasing as a composition of non-decreasing functions.

        \textbf{Showing that $F_T([0, 1]) = [F_T(0), 1] \subset (0, 1]$.} Since each ${\sf w}_h$ is continuous, so is $F$ as a composition of continuous functions. Since $F$ is non-decreasing, to show that $F_T([0, 1]) = [F_T(0), 1] \subset (0, 1]$, it suffices to show that $F_T(0) > 0$ and $F_T(1) = 1$. We have:
        \begin{align*}
            F_T(0) > 0 &\iff 1 - \left(1 - \frac{G}{t}\right)\left[\sum_{h \in [G]}\lambda_h^*\left(1 + {\sf w}_{h}(n^*_{h,t}(1 - u) - 1)\right)^{\frac{2p}{p+1}}\right]^{-1} > 0 \\
            &\iff \sum_{h \in [G]}\lambda_h^*\left(1 + {\sf w}_{h}(n^*_{h,t}(1 - u) - 1)\right)^{\frac{2p}{p+1}} > 1 - \frac{G}{t}.
        \end{align*}
        Since each ${\sf w}_h$ is non-negative, we have 
        \begin{equation*}
            \sum_{h \in [G]}\lambda_h^*\left(1 + {\sf w}_{h}(n^*_{h,t}(1 - u) - 1)\right)^{\frac{2p}{p+1}} \geq \sum_{g \in [G]} \lambda_g^* = 1 > 1 - \frac{G}{t},
        \end{equation*}
        where the strict inequality stems from $\frac{G}{t} < 1$. Hence $F_T(0) > 0$. On the other hand,
        \begin{align*}
            F_T(1) &= 1 - \left(1 - \frac{G}{t}\right)\left[\sum_{h \in [G]}\lambda_h^*\left(1 + {\sf w}_{h}(n^*_{h,t}(1 - 1) - 1)\right)^{\frac{2p}{p+1}}\right]^{-1} \\
            &= 1 - \left(1 - \frac{G}{t}\right)\left[\sum_{h \in [G]}\lambda_h^*\left(1 + {\sf w}_{h}(- 1)\right)^{\frac{2p}{p+1}}\right]^{-1} \\
            &= 1 - \left(1 - \frac{G}{t}\right)\left[\sum_{h \in [G]}\lambda_h^*\left(1 + (+\infty)\right)^{\frac{2p}{p+1}}\right]^{-1} \\
            &= 1 - 0 \\
            &= 1,
        \end{align*}
        where we used ${\sf w}_h$ being infinite on the negative numbers. This completes the proof.

        \noindent \textbf{Showing $\frac{w_{1,p} + \frac{G}{t}}{1 + w_{1,p}} \leq F_T(0) = w_{1,p} + \frac{G}{t}$.} We have
        \begin{align*}
            F_T(0) &= 1 - \frac{1 - \frac{G}{t}}{\sum_{h \in [G]}\lambda_h^*\left(1 + {\sf w}_{h}(n^*_{h,t} - 1)\right)^{\frac{2p}{p+1}}} =  1 - \frac{1 - \frac{G}{t}}{1 + w_{1,p}} =  \frac{w_{1,p} + \frac{G}{t}}{1 + w_{1,p}} \leq w_{1,p} + \frac{G}{t},
        \end{align*}
        where the first step stems from definition of $F_T$, the second step stems from the definition of $w_{1,p}$, and the last step stems from $w_{1,p} \geq 0$. 
        \item Let $c \in (0, 1)$. We have:
        \begin{equation*}
            \max_{h \in [G]}{\sf w}_h(c n^*_{h,t} - 1) = \max_{h \in [G]}{\sf w}_h(n^*_{h,c\cdot t} - 1) \to 0,
        \end{equation*}
        thus by the dominated convergence theorem, we have:
        \begin{equation*}
            F_T(c) = 1 - \left(1 - \frac{G}{t}\right)\left[\sum_{h \in [G]}\lambda_h^*\left(1 + {\sf w}_{h}(n^*_{h,t}(1 - c) - 1)\right)^{\frac{2p}{p+1}}\right]^{-1} \to 0.
        \end{equation*}
        \item Let $x_{t}$ be a non-negative sequence satisfying $x_{t} \to 0$. 
        
        \noindent \textbf{Limit of $F_T(x_{t})$.} Since $F_T \geq 0$, taking the limsup operator yields
        \begin{equation}\label{liminf}
             \limsup F_T(x_{t}) \geq  \liminf F_T(x_{t}) \geq 0.
        \end{equation}
        Following the definition of $F_T$, we have:
        \begin{align*}
            F_{\boldsymbol{\lambda}^*,t}(x_{t}) &= 1 - \left(1 - \frac{G}{t}\right)\left[\sum_{h \in [G]}\lambda_h^*\left(1 + {\sf w}_{h}(n^*_{h,t}(1 - x_{t}) - 1)\right)^{\frac{2p}{p+1}}\right]^{-1} 
        \end{align*}
        By taking the limsup, and using the fact that each ${\sf w}_h$ is decreasing, we have by the dominated convergence theorem:
        \footnotesize
        \begin{align*}
            \limsup F_{\boldsymbol{\lambda}^*,T}(x_{t}) &=  1 - \left(1 - \lim \frac{G}{T}\right)\left[1 + \limsup\sum_{h \in [G]}\lambda_h^*\left(\left(1 + {\sf w}_{h}( \left(n^*_{h,T}(1 - x_{t})\right) - 1)\right)^{\frac{2p}{p+1}} - 1\right)\right]^{-1}.
        \end{align*}
        Under the asymptotic regime we consider, we have:
        \begin{equation*}
            \lim \frac{G}{T} = 0.
        \end{equation*}
        Since $x_{t} \to 0$, we have for each $g \in [G]$:
        \begin{equation*}
            0 \leq \limsup {\sf w}_h\left(n^*_{h,T}(1 - x_{t}) - 1\right) \leq \limsup {\sf w}_h\left(\frac{n^*_{h,T}}{2}\right) \leq \max_{h \in [G]}{\sf w}_h\left(n^*_{h,T/2}\right) = 0.
        \end{equation*}
        Therefore, by the dominated convergence theorem,
        \begin{equation}\label{limsup}
             \limsup F_{\boldsymbol{\lambda}^*,T}(x_{t})  \leq 1 - \left(1 - 0\right)\left[1 + 0\right]^{-1} = 0.
        \end{equation}
        Combining inequalities $\eqref{liminf}$ and $\eqref{limsup}$ yields:
        \begin{equation*}
             \limsup F_T(x_{t}) =  \liminf F_T(x_{t}) = 0,
        \end{equation*}
        so that $\lim F_T(x_{t})$ is well defined and is equal to $0$.

        \noindent \textbf{Limit of $F'_{t}(x_{t})$.} The expression of $\lim F'_{t}(x_{t})$ is
        \small
        \begin{align*}
            F'_{\boldsymbol{\lambda}^*,T}(x_{t}) &= -\frac{2p}{p+1}\left(1 - \frac{G}{T}\right) \cdot(-1)\cdot\left[\sum_{h \in [G]}\lambda_h^*\left(1 + {\sf w}_{h}(n^*_{h,T}(1 - x_{t}) - 1)\right)^{\frac{2p}{p+1}}\right]^{-2} \\ &\cdot \left[\sum_{g \in [G]} \lambda_g^*\cdot(-n^*_{h,T})\cdot{\sf w}_h'(n^*_{h,T}(1 - x_{t}) - 1)\left(1 + {\sf w}_{h}(n^*_{h,T}(1 - x_{t}) - 1)\right)^{\frac{p-1}{p+1}}\right] \\
            &= \frac{2p}{p+1}\left(1 - \frac{G}{T}\right)^{-1}\left(1 - F_T(x_{t})\right)^2 \\
            &\cdot \left[\sum_{g \in [G]} \lambda_g^*\cdot\left[-n^*_{h,T}(1 - x_{t})\cdot{\sf w}_h'(n^*_{h,T}(1 - x_{t}) - 1)\right]\frac{\left(1 + {\sf w}_{h}(n^*_{h,T}(1 - x_{t}) - 1)\right)^{\frac{p-1}{p+1}}}{1 - x_{t}}\right],
        \end{align*}
        so that taking the $\limsup$ implies
        \small
        \begin{align*}
            &\limsup \left|F'_{\boldsymbol{\lambda}^*,T}(x_{t})\right| \leq \frac{2p}{p+1}\left(1 - \limsup \frac{G}{T}\right)^{-1}\left(1 - \liminf F_T(x_{t})\right)^2 \\
            &\times \limsup\left|\sum_{g \in [G]} \lambda_g^*\cdot\left[-n^*_{h,T}(1 - x_{t})\cdot{\sf w}_h'(n^*_{h,T}(1 - x_{t}) - 1)\right]\frac{\left(1 + {\sf w}_{h}(n^*_{h,T}(1 - x_{t}) - 1)\right)^{\frac{p-1}{p+1}}}{1 - x_{t}}\right| \\
            &= \frac{2p}{p+1}\times \limsup \sum_{g \in [G]} \lambda_g^*\cdot\left[n^*_{h,T}(1 - x_{t})\cdot \left|{\sf w}_h'\right|(n^*_{h,T}(1 - x_{t}) - 1)\right]\frac{\left(1 + {\sf w}_{h}(n^*_{h,T}(1 - x_{t}) - 1)\right)^{\frac{p-1}{p+1}}}{1 - x_{t}}
        \end{align*}
        where we used that $\limsup \frac{G}{T} = \liminf F_{\boldsymbol{\lambda}^*,T}(x_{t}) = 0$. It remains to calculate
        \small
        \begin{equation*}
            \limsup \sum_{g \in [G]} \lambda_g^*\cdot\left[n^*_{h,T}(1 - x_{t})\cdot \left|{\sf w}_h'\right|(n^*_{h,T}(1 - x_{t}) - 1)\right]\frac{\left(1 + {\sf w}_{h}(n^*_{h,T}(1 - x_{t}) - 1)\right)^{\frac{p-1}{p+1}}}{1 - x_{t}}.
        \end{equation*}
        Since ${\sf w}_h$ is convex and decreasing, we have for each $t \geq 0$:
        \begin{equation*}
            |{\sf w}'_h(t)| = -{\sf w}'_h(t) \leq -\frac{{\sf w}_h(2t) - {\sf w}(t)}{2t - t} = \frac{{\sf w}(t) - {\sf w}_h(2t)}{t} \leq \frac{{\sf w}_h(t)}{t},
        \end{equation*}
        where the first step stems from ${\sf w}_h$ being decreasing, the second step stems from the convexity of of ${\sf w}_h$, and the last step stems from ${\sf w}_h \geq 0$. Applying the inequality above implies that:
        \begin{equation*}
            n^*_{h,T}(1 - x_{t})\cdot \left|{\sf w}_h'\right|(n^*_{h,T}(1 - x_{t}) - 1) \leq \frac{ n^*_{h,T}(1 - x_{t})}{\left[n^*_{h,T}(1 - x_{t}) - 1\right]^+}\cdot {\sf w}_h\left(n^*_{h,T}(1 - x_{t}) - 1\right),
        \end{equation*}
        which goes to $0$ (for each $h \in [G]$) under the convergence regime we consider. Moreover, we also have,
        \begin{equation*}
            \frac{\left(1 + {\sf w}_{h}(n^*_{h,T}(1 - x_{t}) - 1)\right)^{\frac{p-1}{p+1}}}{1 - x_{t}} \to 0.
        \end{equation*}
        Therefore by the dominated convergence theorem, we have:
        $\limsup |F'_{\boldsymbol{\lambda}^*,T}(x_{t})| \to 0$, which implies that $\lim F'_{\boldsymbol{\lambda}^*,T}(x_{t})$ exists and is equal to $0$.
       \end{enumerate}
        $\hfill \square$
\end{proof}

\begin{lemma}[Structural inequality of Variance-UCB]\label{lemma:structural}
    For all groups $g, h \in [G]$, we have
    \begin{equation*}
        \frac{n^*_{g,t}}{n_{g,t}} \leq \frac{n^*_{h,t}(1+ {\sf w}_{h}(n_{h,t} - 1))^{\frac{2p}{p+1}}}{n_{h, t} - 1}.
    \end{equation*}
\end{lemma}
\begin{proof}{\textit{Proof.}}
       Let $s+1$ be the last time a group $h$ has been sampled. Then, at time $s$, group $h$ must have a maximal
       \begin{equation*}
           \frac{{\sf UCB}^{\frac{2p}{p+1}}_{g,s}}{n_{g,s}}.
       \end{equation*}
       Equivalently, for all groups $g$ we must have:
    \begin{equation}\label{ineq0}
        \frac{{\sf UCB}^{\frac{2p}{p+1}}_{g,s}}{n_{g,s}} \leq  \frac{{\sf UCB}^{\frac{2p}{p+1}}_{h,s}}{n_{h,s}}.
    \end{equation}
    Recall that the inequalities  $\boldsymbol{0} \leq \boldsymbol{{\sf UCB}} - \boldsymbol{\sigma} \leq \boldsymbol{\sigma}\boldsymbol{{\sf w}}$ are assumed to be true. On the one hand, by maximality of $s$, group $h$ has not been chosen between time $s+1$ and time $t$, thus for $t \geq s+1$, the equality $n_{h, s} = n_{h, t} - 1$ must be true, and:
    \begin{equation}\label{ineq1}
        \frac{{\sf UCB}^{\frac{2p}{p+1}}_{h,s}}{n_{h,s}} \leq \frac{(\sigma_h + \sigma_h{\sf w}_{h}(n_{h,s}))^{\frac{2p}{p+1}}}{n_{h, s}} = \frac{\sigma_h^{\frac{2p}{p+1}}(1 + {\sf w}_{h}(n_{h,t} - 1))^{\frac{2p}{p+1}}}{n_{h, t} - 1}.
    \end{equation}
    On the other hand, 
    \begin{equation}\label{ineq2}
         \frac{{\sf UCB}^{\frac{2p}{p+1}}_{g,s}}{n_{g,s}} \geq \frac{\sigma_g^{\frac{2p}{p+1}}}{n_{g,t}}.
    \end{equation}
    Combining Inequalities \eqref{ineq1} and \eqref{ineq2} in Inequality \eqref{ineq0} yields
    \begin{equation*}
        \forall g, h \in [G], \quad  \frac{\sigma_g^{\frac{2p}{p+1}}}{n_{g,t}} \leq \frac{\sigma_h^{\frac{2p}{p+1}}(1+ {\sf w}_{h}(n_{h,t} - 1))^{\frac{2p}{p+1}}}{n_{h, t} - 1},
    \end{equation*}
    by multiplying both terms by $\frac{t}{\Sigma_p}$, we obtain 
    \begin{equation*}
        \frac{n^*_{g,t}}{n_{g,t}} \leq \frac{n^*_{h,t}(1+ {\sf w}_{h}(n_{h,t} - 1))^{\frac{2p}{p+1}}}{n_{h, t} - 1},
    \end{equation*}
    where we used $n^*_{g,t} = \frac{\sigma_g^{\frac{2p}{p+1}}}{\Sigma_p} t$ (see Lemma \ref{lemma:completeinformation}). This completes the proof of Lemma \ref{lemma:structural}.
    $\hfill \square$
\end{proof}

\begin{lemma}\label{lemma:valueab}
    For each $g \in [G]$, we set
    \begin{align*}
    a_g &:= -\left[\frac{1}{n^*_{g,T}} + \left(\left(1 + {\sf w}_h(n^*_{g,T} - 1)\right)^{\frac{2p}{p+1}} - 1\right)\right], \\
    b_g &:= \left(w_{1,p} + \frac{G}{T}\right).
\end{align*}
We have:
\begin{equation*}
        \sum_{g \in [G]}\lambda_g^* a_g^2 = w_{2,p}^2 + \frac{G}{T}\cdot o(1), \qquad \sum_{g \in [G]}\lambda_g^* b_g^2= w_{1,p}^2 + \frac{G}{T} \cdot o (1).
\end{equation*}
Moreover, we have for $p < +\infty$
\begin{equation*}
    \sum_{g \in [G]}\lambda_g^* (a_g^2 + b_g^2) = \bar{w}_p^2 + \frac{G}{T}\cdot o(1) = (1 + o(1))\left(\bar{w}_p^2 + \frac{G}{T}\right).
\end{equation*}
\end{lemma}
\begin{proof}{\textit{Proof.}}
    \begin{enumerate}
        \item We have:
        \begin{align*}
             \sum_{g \in [G]}\lambda_g^* b_g^2 &= \left(w_{1,p} + \frac{G}{T}\right)^2 \\
             &= w_{1,p}^2 + \frac{G}{T}\left(w_{1,p} + \frac{G}{T}\right) \\
             &= w_{1,p}^2 + \frac{G}{T} \cdot o(1),
        \end{align*}
        where the first step follows from the observation that $b_g$ does not depend on the choice of $g$ and that $\sum_{g \in [G]} \lambda_g^* = 1$, and the third step follows from $w_{1,p}, \frac{G}{T} \to 0$.
        \item Similarly, we have:
        \begin{align*}
            \sum_{g \in [G]}\lambda_g^* a_g^2 &= \sum_{g \in [G]} \lambda_g^* \left[\frac{1}{n^*_{g,T}} + \left(\left(1 + {\sf w}_h(n^*_{g,T} - 1)\right)^{\frac{2p}{p+1}} - 1\right)\right]^2 \\
            &= \sum_{g \in [G]}\lambda_g^*\left(\left(1 + {\sf w}_h(n^*_{g,T} - 1)\right)^{\frac{2p}{p+1}} - 1\right)^2 + \sum_{g \in [G]} \frac{\lambda_g^*}{(n_{g,T}^*)^2} + \sum_{g \in [G]} \frac{2\lambda_g^*}{n_{g,T}^*}\left(\left(1 + {\sf w}_h(n^*_{g,T} - 1)\right)^{\frac{2p}{p+1}} - 1\right) \\
            &= w_{2,p}^2 + \frac{1}{T}\left(\sum_{g \in [G]}\frac{1}{n^*_{g,T}}\right) + \frac{2}{T}\sum_{g \in [G]}\left(\left(1 + {\sf w}_h(n^*_{g,T} - 1)\right)^{\frac{2p}{p+1}} - 1\right) \\
            &= w_{2,p}^2 + \frac{G}{T}\cdot o(1).
        \end{align*}
        where the first step follows from the definition of $a_g$, the second step follows from expanding the square, and the third step follows from the definition of $w_{2,p}$ (see Definition \ref{definition:uncertainty}). To understand the fourth step, notice that for each fixed $g$, we have:
        \begin{equation*}
            \frac{1}{n^*_{g,T}} \to 0, \qquad \text{and} \qquad \left(\left(1 + {\sf w}_h(n^*_{g,T} - 1)\right)^{\frac{2p}{p+1}} - 1\right) \to 0,
        \end{equation*}
        hence
        \begin{equation*}
            \sum_{g \in [G]} \frac{1}{n^*_{g,T}} = G \cdot o(1), \qquad \sum_{g \in [G]} \left(1 + {\sf w}_h(n^*_{g,T} - 1)\right)^{\frac{2p}{p+1}} - 1 = G \cdot o(1)
        \end{equation*}
        \item The equality $    \sum_{g \in [G]}\lambda_g^* (a_g^2 + b_g^2) = \bar{w}_p^2 + \frac{G}{T}\cdot o(1).
$ is straightforward from the first two, by noticing that for finite $p$, we have $\bar{w}_p^2 = w_{1,p}^2 + w_{2,p}^2$ (see Definition \ref{definition:uncertainty}).
    \end{enumerate}
\end{proof}
\subsection{Taylor inequality for $R_p$ with finite $p$}\label{appendix:proofdp}
    The goal of this section is to derive a Taylor approximation of the objective in the regret's expression stated in Proposition \ref{lemma:regret3}:
    \begin{equation*}
        \boldsymbol{\lambda} \in \Delta_G \mapsto \left\|\frac{(\boldsymbol{\lambda}^*)^{\frac{1}{p}}}{\boldsymbol{1} - \boldsymbol{\delta}}\right\|_p.
    \end{equation*}
    Mainly, we want to prove the following bounding:
    \begin{proposition}\label{lemma:taylorfinitep}
        For any $\boldsymbol{\lambda},\boldsymbol{\lambda}^* \in \Delta_G$, we have:
        \begin{equation*}
            \left|\left\|\frac{(\boldsymbol{\lambda}^*)^{\frac{1}{p}}}{\boldsymbol{1} - \boldsymbol{\delta}}\right\|_p - 1 - \frac{p+1}{2}\sum_{g \in [G]}\lambda_g^*\delta_g^2\right|  \leq \frac{p^2 \|\boldsymbol{\delta}\|_\infty^3}{(1 - \delta_{\max})^{3p + 3}}
        \end{equation*}
    \end{proposition}
   Before we prove Proposition \ref{lemma:taylorfinitep}, we need to prove some intermediate results. These are established as follows:
   \begin{enumerate}
       \item Writing down the Taylor decomposition into two terms, a dominant term and a rest (see Equation \eqref{equation:taylorintegral}).
       \item Writing down the needed calculations (see Lemma \ref{lemma:derivatives}.
       \item Establishing a closed formula for the dominant term (see Lemma \ref{lemma:dominant}).
       \item Bounding the rest (see Lemma \ref{lemma:boundingrest}).
   \end{enumerate}
   
   For any pair $(\boldsymbol{\lambda}, \boldsymbol{\lambda}^*) \in \Delta_G$, we have
    \begin{equation*}
        \sum_{g \in [G]}\lambda_g^*\delta_g(\boldsymbol{\lambda}||\boldsymbol{\lambda}^*) = \sum_{g \in [G]}\lambda_g^*\left(1 - \frac{\lambda_g}{\lambda_g^*}\right) = \sum_{g \in [G]}\lambda_g^* - \lambda_g = 1 - 1,
    \end{equation*}
    so that $\boldsymbol{\lambda}^* \perp \boldsymbol{\delta}(\cdot || \boldsymbol{\lambda}^*)$. Moreover, 
    \begin{equation*}
    \forall g \in [G], \qquad \delta_g(\boldsymbol{\lambda}||\boldsymbol{\lambda}^*) = 1 - \frac{\lambda_g}{\lambda_g^*} \leq 1.
    \end{equation*}
    Conversely, it is easy to prove that any vector satisfying
    \begin{equation*}
        \boldsymbol{\delta} \perp \boldsymbol{\lambda}^*, \qquad \boldsymbol{\delta} \leq \boldsymbol{1},
    \end{equation*}
    induces a distribution $\boldsymbol{\lambda} \in \Delta_G$ for which $\boldsymbol{\delta} = \boldsymbol{\delta}(\boldsymbol{\lambda}||\boldsymbol{\lambda}^*)$(for that, it suffices to check that $\boldsymbol{\lambda}(\boldsymbol{\delta}) = \boldsymbol{\lambda}^*(1 - \boldsymbol{\delta})$ works). Hence, for simplicity, we focus on calculating a Taylor approximation of the function:
    \begin{equation*}
        r: \boldsymbol{\delta} \mapsto \left\|\frac{(\boldsymbol{\lambda}^*)^{\frac{1}{p}}}{\boldsymbol{1} - \boldsymbol{\delta}}\right\|_p.
    \end{equation*}
    We introduce the relevant partial derivatives:
    \begin{align*}
        &\forall g, h, k \in [G], \qquad \partial_g r = \frac{\partial r}{\partial \delta_g}, \qquad \partial_{g,h}r = \partial_g \left[\partial_h r\right], \qquad \partial_{h,g,k}r_p = \partial_k\left[\partial_{h,g} r_p\right] \\
        &\forall g \in [G], \qquad \partial_g^0 r = \text{Id}, \qquad \partial_g^2 r = \partial_{g,g} r, \qquad \partial_g^3 r = \partial_{g,g,g}r,
    \end{align*}
    where Id is the identity function. By Schwartz's integrability theorem, the partial derivatives can be swapped and the indices above are interchangeable. Moreover, for each $G-$uplet $\boldsymbol{\alpha} \in \mathbb{N}^G$, we introduce the following multi-index notation
    \begin{align*}
        |\boldsymbol{\alpha}| &= \sum_{g \in [G]} \alpha_g \\
        \boldsymbol{\alpha}! &= \alpha_1! \ldots \alpha_G ! \\
        \boldsymbol{x}^{\boldsymbol{\alpha}} &= x_1^{\alpha_1} \ldots x_G^{\alpha_G} \\
        \partial^{\boldsymbol{\alpha}} r &= \partial_1^{\alpha_1} \ldots \partial_G^{\alpha_G} r,
    \end{align*}
    The multivariate Taylor's formula (with rest) applied on the function $r$ on the variable $\boldsymbol{\delta}$ around $\boldsymbol{0}_G$ up to order $2$ (with rest of order $3$)  can be written as
    \begin{equation}\label{equation:taylorintegral}
        r(\boldsymbol{\delta}) = \underbrace{\sum_{\boldsymbol{\alpha}:|\boldsymbol{\alpha}| \leq 2} \frac{\partial^{\boldsymbol{\alpha}}r(\boldsymbol{0}_G)}{\boldsymbol{\alpha}!}\boldsymbol{\delta}^{\boldsymbol{\alpha}}}_{\text{Dominant term}} + \underbrace{\sum_{|\boldsymbol{\alpha}| = 3} \frac{|\boldsymbol{\alpha}|}{\boldsymbol{\alpha}!}\int_0^1(1 - t)^{|\boldsymbol{\alpha}| - 1} \partial^{\boldsymbol{\alpha}}\left[r(t \boldsymbol{\delta})\right]( \boldsymbol{\delta})^{\boldsymbol{\alpha}}dt}_{\text{Rest}}.
    \end{equation}
    The first term in Equation \eqref{equation:taylorintegral}
    \begin{equation*}
\sum_{\boldsymbol{\alpha}:|\boldsymbol{\alpha}| \leq 2} \frac{\partial^{\boldsymbol{\alpha}}r(\boldsymbol{0}_G)}{\boldsymbol{\alpha}!}\boldsymbol{\delta}^{\boldsymbol{\alpha}}
    \end{equation*}
    is called the dominant term, and the second term
    \begin{equation*}
        \sum_{|\boldsymbol{\alpha}| = 3} \frac{|\boldsymbol{\alpha}|}{\boldsymbol{\alpha}!}\int_0^1(1 - t)^{|\boldsymbol{\alpha}| - 1} \partial^{\boldsymbol{\alpha}}\left[r(t \boldsymbol{\delta})\right]( \boldsymbol{\delta})^{\boldsymbol{\alpha}}dt
    \end{equation*}
    is called the rest. The next part of the proof will consist of (1) calculating exactly the dominant term (in terms of $\boldsymbol{\lambda}^*$ and $\boldsymbol{\delta}$), and (2) bounding the rest. Before that, we will derive formulas for the derivatives of $r$. To avoid tedious calculations, we introduce the function
    \begin{equation*}
        R: \boldsymbol{\delta} \mapsto \frac{1}{p} r^p(\boldsymbol{\delta}) = \frac{1}{p}\sum_{g \in [G]}\frac{\lambda_g^*}{(1 - \delta_g)^{p}}
    \end{equation*}
    \begin{lemma}[Derivatives expressions]\label{lemma:derivatives}
        For each $\boldsymbol{\delta}$, we have:
        \begin{enumerate}
            \item $\partial_g R(\boldsymbol{\delta}) = \frac{\lambda_g^*}{(1 - \delta_g)^{1+p}}$
            \item $\partial_{g,h} R(\boldsymbol{\delta}) = \mathbbm{1}_{g = h}(p+1)\frac{\lambda_g^*}{(1 - \delta_g)^{p+2}}$
            \item $\partial_{g,h,k}R(\boldsymbol{\delta}) = \mathbbm{1}_{g = h = k}(p+1)(p+2)\frac{\lambda_g^*}{(1 - \delta_g)^{p+3}}$
        \end{enumerate}
        Moreover, the following functional equalities hold
        \begin{enumerate}
            \item $\partial_g R = (\partial_g r) r^{p-1}$
            \item $\partial_{g,h} R = (\partial_{g,h} r) r^{p-1} + (p-1)(\partial_g r)(\partial_h r)r^{p-2} = (\partial_{g,h} r) r^{p-1} + p^{-1}(p-1)[\partial_g R][\partial_h R]R^{-1}$
            \item $\partial_{g,h,k}R = (\partial_{g, h, k} r) r^{p-1} + (p-1)\left[\sum_{\text{cyc}}(\partial_{g,h} r)(\partial_k r) \right]r^{p-2} + (p-1)(p-2)(\partial_g r)(\partial_h r)(\partial_k r) r^{p-3}$
        \end{enumerate}
        Finally, the following expression holds:
        \scriptsize
        \begin{equation*}
            \partial_{g, h, k} r = (p+1)(p+2)r^{1-p}\mathbbm{1}(g = h = k) \frac{\lambda_g^*}{(1 - \delta_g)^{p+3}} + (p - 1)(2p - 1) r^{1 - 3p}\frac{\lambda_g^* \lambda_h^* \lambda^*_k}{[(1 - \delta_g)(1 - \delta_h)(1 - \delta_t)]^{p+1}} -(p^2-1)r^{1-2p}\sum_{\text{cyc}}\frac{\mathbbm{1}(g = h)\lambda_g^*\lambda_k^*}{(1 - \delta_g)^{p+2}(1 - \delta_k)^{p+1}}
        \end{equation*}
        \normalsize
        \end{lemma}
        \begin{proof}{\textit{Proof.}}
        First, we derive the three partial derivatives formulas.
            \begin{enumerate}
                \item Let $g \in [G]$. We have 
                \begin{align*}
                    \partial_g R &= \frac{1}{p} \sum_{h \in [G]}\lambda_h^*\cdot \partial_g\left[\boldsymbol{\delta} \mapsto \frac{1}{(1 - \delta_h)^{p}}\right] = \frac{1}{p} \cdot p \mathbbm{1}_{g = h} \cdot \left[\lambda_h^* \cdot \left((1-\delta_h)^{-p-1}\right)\right] = \frac{\lambda_g^*}{(1 - \delta_g)^{1 + p}},
                \end{align*}
                where the first step stems from the linearity of the derivation operator, the second step stems from the derivation of the function $x \mapsto (1 - x)^{-p}$, and the third step stems from simplifying the expression.
            \item For $g, h \in [G]$, we have:
            \begin{equation*}
                \partial_{g,h}R = -\lambda_g^*\partial_{h}\left[\boldsymbol{\delta} \mapsto (1 - \delta_g)^{-p-1}\right] = -\lambda_g^* \mathbbm{1}_{g = h} (p+1)\cdot (-1) \cdot (1 - \delta_h)^{-p -2} = \mathbbm{1}_{g = h}(p+1) \frac{\lambda_g^*}{(1 - \delta_g)^{p+2}},
            \end{equation*}
            where the first step stems from the linearity of the derivation operator and the expression of the derivative of $R$ derived earlier, the second step stems from the derivation of the function $x \mapsto (1 - x)^{-p-1}$, and the last step stems from simplifying the expression. To derive the second part, we replace each of $\partial_g r$ and $\partial_h r$ by its value from the first functional equality, and use that $r^p = p R$.
            \item Similarly, for $g, h, k \in [G]$, we have:
            \begin{equation*}
                \partial_{g, h, k}R = \mathbbm{1}_{g = h}(p+1)\lambda_g^* \partial_k\left[\boldsymbol{\delta} \mapsto \frac{1}{(1 - \delta_g)^{p+2}}\right] = \mathbbm{1}_{g = h}\mathbbm{1}_{g = k} \lambda_g^* (1 - \delta_g)^{-p -3} = \mathbbm{1}_{g = h = k} \frac{\lambda_g^*}{(1 - \delta_g)^{p+3}},
            \end{equation*}
            where the first step stems from derivating $\partial_{g,h}R$ and the linearity of derivation, the second step stems from derivating $x \mapsto (1 - x)^{-p-2}$, and the third step stems from simplifying the expression.
            \end{enumerate}
            Next, we derive the functional equalities.
            \begin{enumerate}
                \item By choice of the function $R$, we have
                \begin{equation*}
                    r^p = p \cdot R,
                \end{equation*}
                so that by derivating both sides we obtain
                \begin{equation*}
                    p (\partial_g r)r^{p-1} = p \partial_g R \iff \partial_g R = (\partial_g r) r^{p-1},
                \end{equation*}
                which proves the first equality.
                \item By applying the derivation product rule on the equality above, we obtain
                \begin{equation*}
                    \partial_{g, h} R = \partial_{h}[(\partial_g r) r^{p-1}] = (\partial_{h,g}r)r^{p-1} + (p-1)(\partial_g r)[\partial_h (r^{p-1})] = (\partial_{h,g}r)r^{p-1} + (p-1)(\partial_g r)(\partial_h r)r^{p-2}. 
                \end{equation*}
                \item In a similar fashion, the third expression can be obtained by applying the product derivation rule to the expression above.
            \end{enumerate}
           Finally, we derive the expression of $\partial_{g,h,k}r$ by combining both of the closed forms of the derivatives of $R$ as well as their functional equalities. The starting point is the functional equality for $\partial_{g,h,k}R$, which can be rewritten as
                \begin{align*}
                    \partial_{g,h,k}r &= [\partial_{g,h,k}R]r^{1-p} \\ &- (p-1)(p-2)\underbrace{[(\partial_gr)r^{p-1}][(\partial_h r)r^{p-1}][(\partial_k)r^{p-1}]}_{\text{Apply 2.1 for each term}}(r^{1-p})^{3}r^{p-3} r^{1-p}  \\ &- (p-1)\left[\sum_{cyc} \underbrace{[(\partial_{g,h}r)r^{p-1}]}_{\text{Apply 2.2}}\underbrace{[(\partial_kr)r^{p-1}]}_{\text{Apply 2.1}}\right]r^{p-2}(r^{1-p})^2r^{1-p} \\
                    &=[\partial_{g,h,k}R]r^{1-p} \\
                    &-(p-1)(p-2)[\partial_g R][\partial_h R][\partial_k R] r^{3(1 - p) + (p-3) + (1 - p)} \\
                    &- (p-1)\left[\sum_{cyc} \left[\partial_{g,h}R -\left( (p-1)[\partial_{g}R][\partial_{h}R] \underbrace{p^{-1}R^{-1}}_{\text{Apply $r^p = p R$}}\right)\right]\cdot\left[\partial_k R\right]\right]r^{p-2 + 2(1-p) + 1-p}\\
                    &= [\partial_{g,h,k} R]r^{1-p} \\
                    &-(p-1)(p-2)[\partial_g R][\partial_h R][\partial_k R]r^{1-3p} \\
                    &-(p-1)\sum_{cyc}[\partial_{g,h}R][\partial_k R]r^{1-2p} \\
                    &+(p-1)^2\sum_{cyc}[\partial_g R][\partial_h R][\partial_k R]r^{1-3p} \\
                    &= \underbrace{[\partial_{g,h,k} R]}_{\text{Apply 1.3}}r^{1-p} + [ 3(p-1)^2 - (p-1)(p-2) ]\underbrace{[\partial_g R][\partial_h R][\partial_k R]}_{\text{Apply 1.1 for each term}}r^{1-3p} - (p-1)r^{1-2p}\sum_{cyc}\underbrace{[\partial_{g,h}R][\partial_k R]}_{\text{Apply 1.2 and 1.1}} \\
                    &= (p+1)(p+2)\mathbbm{1}_{g = h = k}r^{1-p}\frac{\lambda_g^*}{(1 - \delta_g)^{p+3}}  + (p-1)(2p - 1)r^{1- 3p}\frac{\lambda_g^* \lambda_h^* \lambda_k^*}{[(1 - \delta_g)(1 - \delta_h)(1 - \delta_k)]^{1 + p}} \\
                    &-(p-1)(p+1)r^{1-2p}\sum_{cyc} \frac{\mathbbm{1}_{g = h}\lambda_g^* \lambda_k^*}{(1 - \delta_g)^{p+2}(1 - \delta_k)^{p+1}},
                \end{align*}
                which completes the proof.
            $\hfill \square$
        \end{proof}
        \begin{lemma}[Calculating the dominant term]\label{lemma:dominant} The first term (i.e., dominant term) in Equation \eqref{equation:taylorintegral} satisfies
        \begin{equation*}
            \sum_{\boldsymbol{\alpha}:|\boldsymbol{\alpha}| \leq 2} \frac{\partial^{\boldsymbol{\alpha}}r(\boldsymbol{0}_G)}{\boldsymbol{\alpha}!}\boldsymbol{\delta}^{\boldsymbol{\alpha}} =  1 + \frac{p+1}{2}\sum_{g} \lambda_g^* \delta_g^2
        \end{equation*}
        \end{lemma}
        \begin{proof}{\textit{Proof.}}
        The dominant term above can be expressed simpler as:
        \begin{equation*}
            \sum_{\boldsymbol{\alpha}:|\boldsymbol{\alpha}| \leq 2} \frac{\partial^{\boldsymbol{\alpha}}r(\boldsymbol{0}_G)}{\boldsymbol{\alpha}!}\boldsymbol{\delta}^{\boldsymbol{\alpha}} = r(\boldsymbol{0}_G) + \langle \nabla r(\boldsymbol{0}_G), \boldsymbol{\delta}\rangle + \frac{1}{2}\langle H(\boldsymbol{0}_G)\boldsymbol{\delta}, \boldsymbol{\delta}\rangle,
        \end{equation*}
            where $\nabla r$, $H$ are respectively the gradient and hessian of the function $r$. We evaluate each of the three terms separately:
            \begin{enumerate}
                \item \textbf{Evaluating $r(\boldsymbol{0}_G)$.} We have
                \begin{equation*}
                    r(\boldsymbol{0}_G) = \left\|\frac{(\lambda_g^*)^{\frac{1}{p}}}{\boldsymbol{1} - \boldsymbol{0}_G}\right\|_p = \left(\sum_{g \in [G]} \left(\lambda_g^*\right)^{\frac{1}{p}\cdot p}\right)^{\frac{1}{p}} = 1,
                \end{equation*}
                where the first equality stems from the definition of $r$, the second equality stems from the definition of the $p$ norm, and the third equality stems from $\sum_{g \in [G]} \lambda_g^* = 1$.
                \item \textbf{Evaluating $\langle \nabla r(\boldsymbol{0}_G), \boldsymbol{\delta}\rangle$.} By combining both expressions for $\partial_g R$ provided in Lemma \ref{lemma:derivatives}, we have
                \begin{equation*}
                    \partial_g R(\boldsymbol{0}_G) = \frac{\lambda_g^*}{(1 - 0)^{1+p}} = [\partial_g r(\boldsymbol{0}_G)]r^{p-1}(\boldsymbol{0}_G),
                \end{equation*}
                or equivalently,
                \begin{equation*}
                    \partial_g r(\boldsymbol{0}_G) = r^{1 - p}(\boldsymbol{0}_G)\left[\frac{\lambda_g^*}{(1 - 0)^{1+p}}\right] = \lambda_g^*,
                \end{equation*}
                where we used that $r(\boldsymbol{0}_G) = 1$. Hence,
                \begin{equation*}
                    \langle \nabla r(\boldsymbol{0}_G), \boldsymbol{\delta}\rangle = \sum_{g \in [G]}\lambda_g^* \delta_g = 0,
                \end{equation*}
                where we used $\boldsymbol{\delta} \perp \boldsymbol{\lambda}^*$.
                \item \textbf{Evaluating $\frac{1}{2}\langle H(\boldsymbol{0}_G)\boldsymbol{\delta}, \boldsymbol{\delta}\rangle$.} Similarly, we combine both expressions for $\partial_{g,h} R$ provided in Lemma \ref{lemma:derivatives}:
                \begin{equation*}
                    \partial_{g,h} R(\boldsymbol{0}_G) = \mathbbm{1}(g = h)(p+1)\frac{\lambda_g^*}{(1 - 0)^{p+2}} = [\partial_{g,h} r(\boldsymbol{0}_G)] r^{p-1}(\boldsymbol{0}_G) + (p-1)(\partial_g r(\boldsymbol{0}_G))(\partial_h r(\boldsymbol{0}_G)) r^{p-2}(\boldsymbol{0}_G),
                \end{equation*}
                or equivalently, by isolating the term $\partial_{g,h}r(\boldsymbol{0}_G)$,
                \begin{align*}
                    \partial_{g,h} r(\boldsymbol{0}_G) &= r^{1-p}(\boldsymbol{0}_G)\left[\mathbbm{1}_{g = h}(p+1)\lambda_g^* - (p-1)[\partial_g r \cdot \partial_h r \cdot r^{p-2}](\boldsymbol{0}_G)\right] \\
                    &= 1 \times \left[\mathbbm{1}_{g = h}(p+1) \lambda_g^* - (p-1)(\lambda_g^*)\cdot(\lambda_h^*)\cdot 1\right] \\
                    &= (p+1)\mathbbm{1}_{g = h} \lambda_g^* - (p-1)\lambda_g^* \lambda_h^*
                \end{align*}
                where we used $r_p(\boldsymbol{0}_G) = 1$, and $\partial_g r(\boldsymbol{0}_G) =\lambda_g^*$ for all groups. Therefore,
                \begin{align*}
                    \frac{1}{2}\langle H(\boldsymbol{0}_G)\boldsymbol{\delta}, \boldsymbol{\delta}\rangle &= \frac{1}{2}\sum_{g, h}\partial_{g,h} r(\boldsymbol{0}_G) \delta_g \delta_h \\
                    &= \frac{1}{2}\sum_{g, h}\left[(p+1)\mathbbm{1}_{g = h} \lambda_g^* - (p-1)\lambda_g^* \lambda_h^*\right] \delta_g \delta_h \\
                    &= \frac{p+1}{2}\sum_{g,h} \mathbbm{1}_{g = h}\lambda_g^* \delta_g^2 - \frac{p-1}{2}\sum_{g,h}(\lambda_g^*\delta_g)\cdot(\lambda_h^*\delta_h) \\
                    &= \frac{p+1}{2}\sum_{g} \lambda_g^* \delta_g^2 - \frac{p-1}{2}\left(\sum_{g \in [G]}\lambda_g^*\delta_g\right)^2 \\
                    &= \frac{p+1}{2}\sum_{g} \lambda_g^* \delta_g^2,
                \end{align*}
                where the first step stems from the definition of the scalar product, the second step stems from the expression of the hessian at coordinate $(g,h)$ at point $\boldsymbol{0}_G$ derived earlier, the third step stems from separating the difference, the fourth step stems from using $\mathbbm{1}_{g = h}$ and factorizing the sum, and the last step stems from $\boldsymbol{\lambda}^* \perp \boldsymbol{\delta}$.
            \end{enumerate}
            Combining the three previous expressions gives that the dominant term is exactly
            \begin{equation*}
                1 +  \frac{p+1}{2}\sum_{g} \lambda_g^* \delta_g^2,
            \end{equation*}
            which completes the proof of Lemma \ref{lemma:dominant}.
            $\hfill \square$
        \end{proof}
    \begin{lemma}[Bounding the rest]\label{lemma:boundingrest} The rest (i.e., second term) in Equation \eqref{equation:taylorintegral} satisfies the following double bounding
    \begin{align*}
        \left|\sum_{\boldsymbol{\alpha}: |\boldsymbol{\alpha}| = 3} \frac{|\boldsymbol{\alpha}|}{\boldsymbol{\alpha}!}\int_0^1(1 - t)^{|\boldsymbol{\alpha}| - 1} \partial^{\boldsymbol{\alpha}}\left[r(t \boldsymbol{\delta})\right]dt \right| \leq \frac{p^2 \|\boldsymbol{\delta}\|_\infty^3}{(1 - \delta_{\max})^{3p + 3}}
    \end{align*}
        
    \end{lemma}
    \begin{proof}{\textit{Proof.}}
    As a reminder, the expression of the rest is
    \begin{equation*}
         \sum_{\boldsymbol{\alpha}: |\boldsymbol{\alpha}| = 3} \frac{|\boldsymbol{\alpha}|}{\boldsymbol{\alpha}!}\int_0^1(1 - t)^{|\boldsymbol{\alpha}| - 1} \partial^{\boldsymbol{\alpha}}\left[r(t \boldsymbol{\delta})\right]dt,
    \end{equation*}
    which is equal to (by noticing that $|\alpha|! = 3!$)
    \begin{equation*}
        \frac{1}{3!}\sum_{\boldsymbol{\alpha}: |\boldsymbol{\alpha}| = 3} \frac{|\boldsymbol{\alpha}|!}{\boldsymbol{\alpha}!}\int_0^1 |\boldsymbol{\alpha}|(1 - t)^{|\boldsymbol{\alpha}| - 1} \partial^{\boldsymbol{\alpha}}[r(t \boldsymbol{\alpha})]dt.
    \end{equation*}
    Each $\boldsymbol{\alpha} \in \mathbb{N}^G$ satisfying $|\boldsymbol{\alpha}| = 3$ can be uniquely encoded into a unique multiset $\{g, h, k\}$ (a set allowing for repetition) in $[G]^3$. Moreover, each multiset $\{g, h, k\}$ can be encoded into a $\frac{|\alpha|!}{\boldsymbol{\alpha}!}$ permutations of $(g, h, k)$. Therefore,
        \begin{equation*}
            \frac{1}{3!}\sum_{\boldsymbol{\alpha}: |\boldsymbol{\alpha}| = 3} \frac{|\boldsymbol{\alpha}|!}{\boldsymbol{\alpha}!}\int_0^1 |\boldsymbol{\alpha}|(1 - t)^{|\boldsymbol{\alpha}| - 1} \partial^{\boldsymbol{\alpha}}[r(t \boldsymbol{\alpha})]dt = \frac{1}{3!}\sum_{g, h, k \in [G]}\int_0^1 3(1 - t)^2 \partial_{g,h,k} r(t \boldsymbol{\delta}) \delta_g \delta_h \delta_k dt
        \end{equation*}
        By linearity of the expectation, the rest is equal to
        \begin{equation}\label{equation:restsimple}
            \frac{1}{6}\int_0^1 3(1-t)^2\left[\sum_{g,h,k}\partial_{g,h,k} r(t \boldsymbol{\delta})\delta_g\delta_h\delta_k\right]dt
        \end{equation}
        
       We now focus on bounding the sum $\sum_{g,h,k}\partial_{g,h,k} r(t \boldsymbol{\delta})\delta_g\delta_h\delta_k$. Following the formula for $\partial_{g,h,k}r$ derived in Lemma \ref{lemma:derivatives}:
        \begin{align*}
            \partial_{g,h,k}r &= (p+1)(p+2)r^{1-p}\mathbbm{1}(g = h = t) \frac{\lambda_g^*}{(1 - \delta_g)^{p+3}} \\
        &+(p-1)(2p - 1) r^{1 - 3p}\frac{\lambda_g^* \lambda_h^* \lambda^*_t}{[(1 - \delta_g)(1 - \delta_h)(1 - \delta_t)]^{p+1}} \\
        &-(p^2-1)r^{1-2p}\sum_{\text{cyc}}\frac{\mathbbm{1}(g = h)\lambda_g^*\lambda_t^*}{(1 - \delta_g)^{p+2}(1 - \delta_t)^{p+1}}
        \end{align*}
        We have:
        \begin{align*}
            \sum_{g, h, k}\partial_{g,h,k} r(t \boldsymbol{\delta})\delta_g \delta_h \delta_k &= (p+1)(p+2)r^{1-p}(t\boldsymbol{\delta})\sum_{g,h,k}\mathbbm{1}(g = h = k) \frac{\lambda_g^*\delta_g^3}{(1 - t\delta_g)^{p+3}} \\
        &+(2p - 1)(p-1) r^{1 - 3p}(t\boldsymbol{\delta})\sum_{g,h,k}\frac{(\lambda_g^*\delta_g) (\lambda_h^*\delta_h)(\lambda^*_k\delta_k)}{[(1 - t\delta_g)(1 - t\delta_h)(1 - t\delta_k)]^{p+1}} \\
        &-(p^2-1)r^{1-2p}(t\boldsymbol{\delta})\sum_{g,h,k}\sum_{\text{cyc}}\frac{\mathbbm{1}(g = h)(\lambda_g^*\delta_g^2)(\lambda_k^* \delta_k)}{(1 - t\delta_g)^{p+2}(1 - t\delta_k)^{p+1}}
        \end{align*}
        We introduce each of the terms
        \begin{align*}
            R_1 &= r^{1-p}(t\boldsymbol{\delta})\sum_{g,h,k}\mathbbm{1}(g = h = k) \frac{\lambda_g^*\delta_g^3}{(1 - t\delta_g)^{p+3}} \\
            R_2 &= r^{1-3p}(t\boldsymbol{\delta})\sum_{g,h,k}\frac{(\lambda_g^*\delta_g) (\lambda_h^*\delta_h)(\lambda^*_k\delta_k)}{[(1 - t\delta_g)(1 - t\delta_h)(1 - t\delta_k)]^{p+1}}\\
            R_3 &= r^{1-2p}(t\boldsymbol{\delta})\sum_{g,h,k}\sum_{\text{cyc}}\frac{\mathbbm{1}(g = h)(\lambda_g^*\delta_g^2)(\lambda_k^*\delta_k)}{(1 - t\delta_g)^{p+2}(1 - t\delta_k)^{p+1}}
        \end{align*}
        so that
        \begin{equation}\label{equation:sumsterms}
            \sum_{g, h, k}\partial_{g,h,k} r(t \boldsymbol{\delta})\delta_g \delta_h \delta_k = (p+1)(p+2) R_1 + (2p - 1)(p-1)R_2 - (p^2 - 1) R_3.
        \end{equation}
        We bound each of the terms $R_1, R_2$, and $R_3$. Since $\boldsymbol{0}_G$ minimizes $r$, we have $r(t\boldsymbol{\delta}) \geq r(\boldsymbol{0}_G) = 1$. Moreover, since $p \geq 1$, we have $1 - 3p \leq 1 - 2p \leq 1 - p \leq 0$ and
        \begin{equation*}
            0 \leq r^{1-3p}(t\boldsymbol{\delta}) \leq r^{1-2p}(t\boldsymbol{\delta}) \leq r^{1-p}(t\boldsymbol{\delta}) \leq 1.
        \end{equation*}
        
        \noindent \textbf{Bounding $R_1$:} We have
        \begin{equation*}
            R_1 = r^{1-p}(t\boldsymbol{\delta})\sum_{g, h ,k \in [G]}\frac{\lambda_g^*\delta_g^3}{(1 - t \delta_g)^{p+3}} \leq  \mathbbm{1}(g = h = k)  \leq 1 \cdot \sum_{g \in [G]} \frac{\lambda_g^*\delta_g^3}{(1 - t \delta_g)^{p+3}} = \sum_{g \in [G]} \frac{\lambda_g^*\delta_g^3}{(1 - t \delta_g)^{p+3}}.
        \end{equation*}
        On the one hand, the function $x \mapsto \frac{x^3}{(1 - tx)^{p+3}}$ is convex for $x \leq 1$, hence by Jensen's inequality,
        \begin{equation*}
            R_1 = r^{1-p}(t\boldsymbol{\delta})\sum_{g \in [G]} \frac{\lambda_g^*\delta_g^3}{(1 - t \delta_g)^{p+3}} \geq r^{1-p}(t\boldsymbol{\delta}) \cdot\frac{\left(\sum_{g \in [G]} \lambda_g^* \delta_g\right)^3}{\left(1 - t\left(\sum_{g \in [G]} \lambda_g^*\delta_g\right)\right)^{p+3}} = 0,
        \end{equation*}
        where we used $\boldsymbol{\delta} \perp \boldsymbol{\lambda}^*$. On the other hand, the function $x \mapsto \frac{x^3}{(1 - tx)^{p+3}}$ is increasing in $[-\infty, 1)$, hence:
        \begin{equation*}
            R_1 \leq \sum_{g \in [G]} \frac{\lambda_g^*\delta_g^3}{(1 - t \delta_g)^{p+3}} \leq \frac{\delta_{\max}^3}{(1 - t \delta_{\max})^{p+3}} \leq \frac{\delta_{\max}^3}{(1 - \delta_{\max})^{p+3}},
        \end{equation*}
        where the second inequality stems from $(1 - t \delta_{\max}) \geq (1 - \delta_{\max})$. Hence
        \begin{equation}\label{equation:bornr1}
            0 \leq R_1 \leq \frac{\delta_{\max}^3}{(1 - \delta_{\max})^{p+3}}.
        \end{equation}
        
        \noindent \textbf{Simplifying $R_2$:} We have
        \begin{equation*}
            R_2 = r^{1-3p}(t\boldsymbol{\delta})\sum_{g, h, k}\frac{(\lambda_g^*\delta_g) (\lambda_h^*\delta_h)(\lambda^*_k\delta_k)}{[(1 - t\delta_g)(1 - t\delta_h)(1 - t\delta_k)]^{p+1}} = r^{1-3p}(t\boldsymbol{\delta})\left(\sum_{g} \frac{\lambda_g^* \delta_g}{(1 - t \delta_g)^{p+1}}\right)^3.
        \end{equation*}
        On the one hand, the functions $x \mapsto \frac{x}{(1 - tx)^{p+1}}$ and $x \mapsto x^3$ are both increasing, thus
        \begin{equation*}
            R_2 = r^{1-3p}(t\boldsymbol{\delta})\left(\sum_{g} \frac{\lambda_g^* \delta_g}{(1 - t \delta_g)^{p+1}}\right)^3 \leq \left(\sum_{g} \frac{\lambda_g^* \delta_{\max}}{(1 - t \delta_{\max})^{p+1}}\right)^3 \leq \frac{\delta_{\max}^3}{(1 - t\delta_{\max})^{3p+3}}.
        \end{equation*}
        On the other hand, we have
        \begin{equation*}
            \sum_{g \in [G]} \frac{\lambda_g^*\delta_g}{(1 - t\delta_g)^{p+1}} \geq \sum_{g: \delta_g < 0} \frac{\lambda_g^*\delta_g}{(1 - t\delta_g)^{p+1}} \geq \sum_{g: \delta_g < 0} \lambda_g^*\delta_g = -\left(\sum_{g: \delta_g > 0}\lambda_g^* \delta_g\right) \geq -\delta_{\max},
        \end{equation*}
        where the first step stems from eliminating the non-negative terms in the sum, the second step stems from $(1 - t\delta_g)^{p+1} \geq 1$ for $\delta_g < 0$, the third step stems from $\boldsymbol{\delta} \perp \boldsymbol{\lambda}^*$, and the last step stems from $\delta_g \leq \delta_{\max}$ for all $g \in [G]$. Hence
        \begin{equation*}
            R_2 = r^{1-3p}(t\boldsymbol{\delta})\left(\sum_{g \in [G]} \frac{\lambda_g^*\delta_g}{(1 - t\delta_g)^{p+1}}\right)^3 \geq - r^{1-p}(t\boldsymbol{\delta})\delta_{\max}^3 \geq - \delta_{\max}^3.
        \end{equation*}
        Therefore,
        \begin{equation}\label{equation:bornr2}
            -\delta_{\max}^3 \leq R_2 \leq \frac{\delta_{\max}^3}{(1 - t\delta_{\max})^{3p + 3}}.
        \end{equation}
        \noindent \textbf{Simplifying $R_3$:} We have
        \begin{align*}
            R_3 &= r^{1-2p}(t\boldsymbol{\delta})\sum_{g,h,k}\sum_{\text{cyc}}\frac{\mathbbm{1}(g = h)(\lambda_g^*\delta_g^2)(\lambda_k^* \delta_k)}{(1 - t\delta_g)^{p+2}(1 - t\delta_k)^{p+1}} \\ &= r^{1-2p}(t\boldsymbol{\delta})\sum_{\text{cyc}}\sum_{g,h,k}\frac{\mathbbm{1}(g = h)(\lambda_g^*\delta_g^2)(\lambda_k^* \delta_k)}{(1 - t\delta_g)^{p+2}(1 - t\delta_k)^{p+1}} \\
            &= r^{1-2p}(t\boldsymbol{\delta})\sum_{\text{cyc}}\sum_{g,k}\frac{(\lambda_g^*\delta_g^2)(\lambda_k^* \delta_k)}{(1 - t\delta_g)^{p+2}(1 - t\delta_k)^{p+1}} \\
            &= r^{1-2p}(t\boldsymbol{\delta})\sum_{\text{cyc}}\left(\sum_{g \in [G]} \frac{\lambda_g^*\delta_g^2}{(1-t\delta_g)^{p+2}}\right)\left(\sum_{k \in [G]} \frac{\lambda_k^*\delta_k}{(1 - t\delta_k)^{p+1}}\right) \\
            &= 3r^{1-2p}(t\boldsymbol{\delta})\left(\sum_{g \in [G]} \frac{\lambda_g^*\delta_g^2}{(1-t\delta_g)^{p+2}}\right)\left(\sum_{g \in [G]} \frac{\lambda_g^*\delta_g}{(1 - t\delta_g)^{p+1}}\right)
        \end{align*}
        where the first step stems from the definition of $R_3$, the second step stems from permuting the sums, the third step stems from using $\mathbbm{1}(g = h)$, the fourth step stems from factorizing the sum, and the final equality stems from the fact that the quantity inside the cyclic sum does not depend on the choice of the index. 
     On the one hand, we have from the bounding inequalities for $R_2$, we have
        \begin{equation*}
            \frac{\delta_{\max}}{(1 - \delta_{\max})^{p+1}} \geq \sum_{g \in [G]} \frac{\lambda_g^*\delta_g}{(1 - t\delta_g)^{p+1}} \geq -\delta_{\max}.
        \end{equation*}
        On the other hand, since $1 - t\delta_g \geq 1 - \delta_{\max} \geq 0$ for all $g \in [G]$, we must have:
        \begin{equation*}
           0 \leq \sum_{g \in [G]} \frac{\lambda_g^*\delta_g^2}{(1-t\delta_g)^{p+2}} \leq \frac{1}{(1 - t\delta_{\max})^{p+2}}\sum_{g \in [G]}\lambda_g^* \delta_g^2 \leq \frac{\|\boldsymbol{\delta}\|_\infty^2}{(1 - t\delta_{\max})^{p+2}}.
        \end{equation*}
        Therefore, by multiplying both inequalities above, we obtain
        \begin{equation}\label{equation:bornr3}
            -3\|\boldsymbol{\delta}\|^2_\infty\delta_{\max}\leq R_3 \leq \frac{3\|\boldsymbol{\delta}\|^2_\infty\delta_{\max}}{(1 - t\delta_{\max})^{2p+3}}.
        \end{equation}

        \noindent \textbf{Conclusion.} We use Inequalities \eqref{equation:bornr1}, \eqref{equation:bornr2}, \eqref{equation:bornr3}, with the expression of $\sum_{g, h, k}\partial_{g,h,k} r(t \boldsymbol{\delta})\delta_g \delta_h \delta_k$ stated in Equation \eqref{equation:sumsterms} to derive a bounding on $\sum_{g, h, k}\partial_{g,h,k} r(t \boldsymbol{\delta})\delta_g \delta_h \delta_k$, from which a bounding on the rest will follow via its expression derived in \eqref{equation:restsimple}. On the one hand,
        \begin{align*}
            \sum_{g, h, k}\partial_{g,h,k} r(t \boldsymbol{\delta})\delta_g \delta_h \delta_k &= (p+1)(p+2) R_1 + (2p - 1)(p-1)R_2 - (p^2 - 1) R_3 \\
            &\geq 0 -(2p - 1)(p-1)\delta_{\max}^3 - \frac{3(p^2 - 1)\|\boldsymbol{\delta}\|^2_\infty\delta_{\max}}{(1 - \delta_{\max})^{2p+3}} \\
            &\geq -\frac{(p-1)(3p+2)\|\boldsymbol{\delta}\|_\infty^3}{(1 - \delta_{\max})^{2p + 3}},
        \end{align*}
        where the last inequality stems from $\delta_{\max} \leq \|\boldsymbol{\delta}\|_\infty$, and $0 \leq (1 - \delta_{\max})^{2p+3} \leq 1$. Thus, the rest can be lower bounded as follows
        \begin{equation*}
            \frac{1}{6}\int_0^1 3(1 - t)^2 \left[\sum_{g, h, k}\partial_{g,h,k} r(t \boldsymbol{\delta})\delta_g \delta_h \delta_k\right]dt \geq -\frac{(p-1)(3p+2)\|\boldsymbol{\delta}\|_\infty^3}{6(1 - \delta_{\max})^{2p + 3}} \geq -\frac{p^2\|\boldsymbol{\delta}\|_\infty^3}{(1 - \delta_{\max})^{2p + 3}}
        \end{equation*}
        where we used $\frac{1}{6}\int_0^1 3(1 - t)^2 = \frac{1}{6}$, and $\frac{(p-1)(3p+2)}{6} \leq p^2$ for $p \geq 1$. On the other hand,
        \begin{align*}
            \sum_{g, h, k}\partial_{g,h,k} r(t \boldsymbol{\delta})\delta_g \delta_h \delta_k &= (p+1)(p+2) R_1 + (2p - 1)(p-1)R_2 - (p^2 - 1) R_3 \\
            &\leq \frac{(p+1)(p + 2)\delta_{\max}^3}{(1 - t\delta_{\max})^{p+3}} + \frac{(2p - 1)(p-1)\delta_{\max}^3}{(1 - t\delta_{\max})^{3p + 3}} + 3(p^2 - 1)\|\boldsymbol{\delta}\|_\infty^2 \delta_{\max} \\
            &\leq \frac{6p^2 \|\boldsymbol{\delta}\|_\infty^3}{(1 - \delta_{\max})^{3p + 3}},
        \end{align*}
        where the last inequality stems from $t \leq 1$, $1- \delta_{\max} \leq 1$, $\delta_{\max} \leq \|\boldsymbol{\delta}\|_\infty$, and $(p+1)(p+2) + 3(p^2 - 1) + (2p - 1)(p-1) = 6p^2$,
        and the rest can be upper bounded as follows
        \begin{equation*}
            \frac{1}{6}\int_0^1 3(1 - t)^2 \left[\sum_{g, h, k}\partial_{g,h,k} r(t \boldsymbol{\delta})\delta_g \delta_h \delta_k\right]dt \leq \frac{p^2 \|\boldsymbol{\delta}\|_\infty^3}{(1 - \delta_{\max})^{3p + 3}},
        \end{equation*}
        where once again we used $\frac{1}{6}\int_0^1 3(1 - t)^2 = \frac{1}{6}$. Therefore,
        \begin{equation*}
            \left|\frac{1}{6}\int_0^1 3(1 - t)^2 \left[\sum_{g, h, k}\partial_{g,h,k} r(t \boldsymbol{\delta})\delta_g \delta_h \delta_k\right]dt\right| \leq \frac{p^2 \|\boldsymbol{\delta}\|_\infty^3}{(1 - \delta_{\max})^{3p + 3}}
        \end{equation*}
        
        This completes the proof of Lemma \ref{lemma:boundingrest}.
        $\hfill \square$
    \end{proof}

    We are now ready to prove Proposition \ref{lemma:taylorfinitep}.
    \begin{proof}\textit{Proof of Proposition \ref{lemma:taylorfinitep}.}
    Let $\boldsymbol{\lambda}, \boldsymbol{\lambda}^* \in \Delta_G$. From Equation \eqref{equation:taylorintegral}, we have
    \begin{equation*}
         r(\boldsymbol{\delta}) - \underbrace{\sum_{\boldsymbol{\alpha}:|\boldsymbol{\alpha}| \leq 2} \frac{\partial^{\boldsymbol{\alpha}}r(\boldsymbol{0}_G)}{\boldsymbol{\alpha}!}\boldsymbol{\delta}^{\boldsymbol{\alpha}}}_{\text{Dominant term}} = \underbrace{\sum_{|\boldsymbol{\alpha}| = 3} \frac{|\boldsymbol{\alpha}|}{\boldsymbol{\alpha}!}\int_0^1(1 - t)^{|\boldsymbol{\alpha}| - 1} \partial^{\boldsymbol{\alpha}}\left[r(t \boldsymbol{\delta})\right]( \boldsymbol{\delta})^{\boldsymbol{\alpha}}dt}_{\text{Rest}}.
    \end{equation*}

    From the definition of $r$, we have
    \begin{equation*}
        r(\boldsymbol{\delta}) = \left\|\frac{(\boldsymbol{\lambda}^*)^{\frac{1}{p}}}{\boldsymbol{1} - \boldsymbol{\delta}}\right\|_p.
    \end{equation*}
    Following Lemma \ref{lemma:dominant}, the dominant term is equal to
    \begin{equation*}
        1 + \frac{p+1}{2}\sum_{g \in [G]}\lambda_g^* \delta_g^2.
    \end{equation*}
    From Lemma \ref{lemma:boundingrest}, the rest is bounded by
    \begin{equation*}
        \frac{6p^2 \|\boldsymbol{\delta}\|_\infty^3}{(1 - \delta_{\max})^{3p + 3}}.
    \end{equation*}
    Therefore,
    \begin{equation*}
        \left|\left\|\frac{(\boldsymbol{\lambda}^*)^{\frac{1}{p}}}{\boldsymbol{1} - \boldsymbol{\delta}}\right\|_p  - 1 - \frac{p+1}{2}\sum_{g \in [G]}\lambda_g^* \delta_g^2\right| \leq \frac{6p^2 \|\boldsymbol{\delta}\|_\infty^3}{(1 - \delta_{\max})^{3p + 3}},
    \end{equation*}
    which completes the proof of Proposition \ref{lemma:taylorfinitep}.
    $\hfill \square$
    \end{proof}

\end{APPENDICES}

\end{document}